# Airalogy: AI-empowered universal data digitization for research automation

Zijie Yang[1,2,3,6,7], Qiji Zhou[1,2,6,7], Fang Guo[1,2,6,7], Sijie Zhang[1,2,6,8], Yexun Xi[1,2,6,8], Jinglei Nie[1,2,6,8], Yudian Zhu[1,2,6,8], Liping Huang[1,4,6,8], Chou Wu[1,3,6,8], Yonghe Xia[1,3,6,8], Xiaoyu Ma[1,3,6,8], Yingming Pu[1,4,6,8], Panzhong Lu[1,2,6,8], Junshu Pan[1,2,6,8], Mingtao Chen[1,2,6,8], Tiannan Guo[1,5,6,9], Yanmei Dou[1,3,6,9], Hongyu Chen[1,4,6,9], Anping Zeng[1,2,6,9], Jiaxing Huang[1,2,6,9], Tian Xu[1,3,6,9], Yue Zhang[1,2,6,9]

[1] Research Center for Industries of the Future, Westlake University, Hangzhou, Zhejiang, China
[2] School of Engineering, Westlake University, Hangzhou, Zhejiang, China
[3] School of Life Sciences, Westlake University, Hangzhou, Zhejiang, China
[4] School of Science, Westlake University, Hangzhou, Zhejiang, China
[5] School of Medicine, Westlake University, Hangzhou, Zhejiang, China
[6] Airalogy, Westlake University, Hangzhou, Zhejiang, China
[7] These authors contributed equally
[8] These authors contributed equally
[9] Corresponding authors
{yangzijie, zhouqiji, guofang, zhangsijie, xiyexun, niejinglei, zhuyudian, huangliping, wuchou, xiayonghe, maxiaoyu, puyingming, lupanzhong, panjunshu, chenmingtao, guotiannan, douyanmei, chenhongyu, zenganping, huangjiaxing, xutian, zhangyue}@westlake.edu.cn

**Abstract**

Research data are the foundation of Artificial Intelligence (AI)-driven science, yet current AI applications remain limited to a few fields with readily available, well-structured, digitized datasets. Achieving comprehensive AI empowerment across multiple disciplines is still out of reach. Present-day research data collection is often fragmented, lacking unified standards, inefficiently managed, and difficult to share. Creating a single platform for standardized data digitization needs to overcome the inherent challenge of balancing between universality (supporting the diverse, ever-evolving needs of various disciplines) and standardization (enforcing consistent formats to fully enable AI). No existing platform accommodates both facets. Building a truly multidisciplinary platform requires integrating scientific domain knowledge with sophisticated computing skills. Researchers often lack the computational expertise to design customized and standardized data recording methods, whereas platform developers rarely grasp the intricate needs of multiple scientific domains. These gaps impede research data standardization and hamper AI-driven progress. In this study, we address these challenges by developing Airalogy (https://airalogy.com), the world's first AI- and community-driven platform that balances universality and standardization for digitizing research data across multiple disciplines. Airalogy represents entire research workflows using customizable, standardized data records and offers an advanced AI research copilot for intelligent Q&A, automated data entry, analysis, and research automation. Already deployed in laboratories across all four schools of Westlake University, Airalogy has the potential to accelerate and automate scientific innovation in universities, industry, and the global research community—ultimately benefiting humanity as a whole.

# 1 Introduction

In recent years, Artificial Intelligence (AI) has demonstrated remarkable potential to advance research and innovation across diverse scientific disciplines[1-12], facilitate research discoveries, and even independently lead research as "AI scientists"[13-18]. Strategic use of AI can accelerate scientific progress and enhance the technological competitiveness of universities, enterprises, and national research endeavors. Despite the above promises, only about 8% of research papers employed AI-related methods, indicating substantial untapped potential for AI-empowered science[19]. This underutilization is largely due to a critical barrier: the lack of high-quality data[19-22]. Current AI-driven studies tend to focus on scientific fields with well-established, publicly available datasets, such as protein structures[1,2], chemical compounds[3,4], or medical images[9,10]. In many other scientific domains, however, assembling sufficient high-quality data—including experimental protocols and records—is challenging, thereby limiting the reach of AI-empowered research.

This issue can be addressed by recording diverse real-world data produced daily in laboratories, where most experimental protocols and data originate. For this purpose, platforms have been introduced to encourage electronic data recording in laboratories[23,24]. However, existing platforms cover only a fraction of experiment types, which falls far short of the needs of broad innovation. Our survey across the four schools of Westlake University (Schools of Life Sciences, Engineering, Science, and Medicine) revealed that only a small fraction of labs employ these tools in their day-to-day research. The core challenge for effective, general research data digitization is the balancing between universality and standardization (**Fig. 1a**). By universality, we refer to the need for a recording platform to be applicable across diverse disciplines and laboratories, whereas standardization ensures that recorded research data follow a uniform structure conducive to efficient use and analysis.

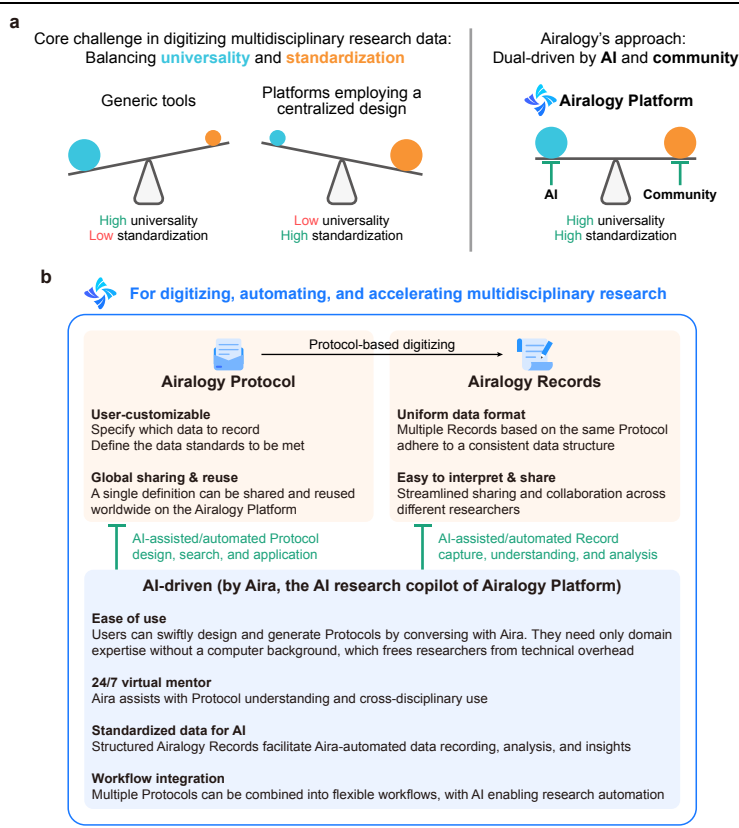

**Fig. 1: Overview of the Airalogy Platform.**
**a,** The core challenge of digitizing multidisciplinary research data and the corresponding solution offered by Airalogy. **b,** Standardized electronic data recording via community-customizable and AI-driven Airalogy Protocols, producing structured, standardized Airalogy Records. The Platform provides an AI research copilot named "Aira", which streamlines the design, application, and analysis of both Protocols and Records.

Concerning universality, many Westlake laboratories use text editors as generic tools to document experimental data, without considering standardization. In a drug screening experiment, for instance, different researchers might produce entries such as: "Record 1 from Researcher 1: Tested Compound A in HeLa cells at 5 μM, inhibition rate 20%", "Record 2 from Researcher 1: Tested Compound A in HeLa cells at 10 μM, inhibition rate 30%", and "Record 3 from Researcher 2: C: HeLa, D: Compound A, I: 45". The shortcomings of these methods are threefold. First, variability in individual recording styles complicates the automated extraction of key parameters (e.g., cell line, drug name, concentration, inhibition rate) even when they are all documented. Second, the absence of a standardized format hampers data sharing; for example, "I" in Record 3 could be mistakenly read as "initial cell density", obscuring the original intent. Third, inexperienced researchers may omit essential details, leading to incomplete records (e.g., Record 3 lacks the drug concentration). Thus, while the generic tool offers excellent universality, it does not deliver the standardization essential for effective data extraction and inter-laboratory sharing.

Concerning standardization, platforms such as Benchling[25] provide a set of pre-defined protocol templates for common experiments like drug treatment assays. However, even a "standard" experiment can have multiple variations—for example, different methods for measuring cell growth inhibition may require recording different parameters. A single platform that relies on pre-defined protocols cannot feasibly meet every specialized need. Moreover, scientific research spans a wide array of fields, and laboratory practices vary significantly among disciplines and change rapidly with scientific advances. For instance, our participating lab pioneered a method for synthesizing gold nanopropellers in 2024[26], whose experimental procedures are newly developed; thus, no existing standard protocol suits it. Consequently, creating pre-defined protocols for every niche experiment is neither financially nor logistically feasible.

In response, we propose a community-driven strategy for balancing universality and standardization in general research data digitization. The main idea is to allow researchers to define protocols for their own experiments, specifying which data to record and the standards by which those data should be recorded. Once created, these protocols can be shared and reused worldwide, allowing users to capture their research data according to its specifications, generating multiple records in a uniform format that is easily interpretable and readily shareable across diverse research fields (**Fig. 1b**).

The contrast between our approach and existing research recording philosophies parallels the difference between Wikipedia[27] and a centrally edited encyclopedia such as the *Encyclopedia Britannica*[28]: while a centralized encyclopedia can offer many articles, it cannot cover every niche topic—particularly those evolving rapidly. In contrast, Wikipedia's community-driven model allows anyone to contribute entries, meeting a wide range of needs. Moreover, the shared nature of a single platform ensures that the contributed content is globally accessible, benefiting the entire community.

The community-driven approach, however, also presents a new challenge: researchers may excel in their scientific domains but lack the computer expertise to design customized protocols. Fortunately, the maturation of AI technologies, particularly Large Language Models (LLMs)[29-35], presents an opportunity to overcome this obstacle. With the capability to understand experimental design, LLMs allows users with domain expertise but no coding experience to quickly define protocols or understand protocols created by others. LLMs can also serve as a virtual mentor available 24/7, delivering real-time guidance on protocol usage and providing automated methods to efficiently convert laboratory data in arbitrary forms into standardized electronic research records. In addition, equipped with research data, LLMs can perform intelligent, automated analyses of research records, further accelerating and automating innovation across various scientific domains.

To verify this strategy, we developed Airalogy (https://airalogy.com), a research platform driven jointly by the community and AI. Airalogy serves as a universal solution for the standardized digitization of research data across multiple scientific disciplines, and also provides "Aira", an AI copilot for accelerating and automating research (**Fig. 1b**). Within the platform, user-defined protocols can be used to guide experiments, standardize records, facilitate global collaboration, and support AI research automation.

Here, we present six real-world case studies drawn from laboratories in diverse disciplines across all four schools of Westlake University, which illustrate how Airalogy's universal and standardized approach to research data digitization yields broader impacts, including the global sharing of cutting-edge research and customized protocols (**Case Study 1; Fig. 6a**), AI-driven tutoring (**Case Study 2; Fig. 6b**), AI-empowered research (**Case Study 3; Fig. 6c**), dynamic evolution of protocols (**Case Study 4; Fig. 6d**), automated research data management and analysis (**Case Study 5; Fig. 6e**), and the fostering of global research cooperation and division of labor (**Case Study 6; Fig. 6f**). We believe that such a community-driven research philosophy—exemplified by Airalogy—will accelerate and automate AI-empowered scientific progress, spur innovation, and enhance the competitiveness of universities, research institutes, enterprises, and national research endeavors, ultimately benefiting humanity as a whole.

# 2 Results

## 2.1 Overview of the Airalogy Platform

Airalogy ([https://airalogy.com](https://airalogy.com), **Fig. 1a**) is the world's first AI- and community-driven platform for the standardized digitization of universal research data. Its design is modeled on the organizational structure of real-world laboratories and the practical execution of research activities. It adopts a three-tier hierarchy: Lab → Project → Protocol (**Supplementary Fig. S37a**). A Lab represents an actual or virtual laboratory, where members can be invited and organized into various groups. Project facilitates independent management of different projects in a lab; every Project can configure its own access permissions (e.g., fully open or restricted to specific groups or members; **Supplementary Fig. S37c**) to meet diverse confidentiality and access-control requirements. Within a given Project, users can use multiple Airalogy Protocols to support the data digitization and analysis.

An Airalogy Protocol can be viewed as a user-customizable, standardized protocol for electronic research data recording and analysis. Researchers generate standardized Records based on these Protocols (**Figs. 1b, 2**). Once designed, a Protocol can be reused across the Platform, promoting global sharing of both digitized data and analytical methods. For instance, a Project aimed at screening new drugs to inhibit tumor cell growth might include one experimental Protocol for conducting drug-inhibition assays (**Fig. 2**) and one analytical Protocol for processing the resulting data (**Fig. 3**). In addition, each Protocol on the Airalogy Platform features a Discussions section (**Supplementary Fig. S37b**), allowing users to pose questions and provide answers, thus accumulating expert knowledge and fostering collaboration.

As described in the **Introduction**, the Airalogy Platform incorporates an AI research copilot named "Aira" (**Fig. 1b**), which enhances the Platform through semi-automated Protocol design (**Fig. 4a**), intelligent Protocol search and recommendation (**Fig. 4b**), round-the-clock experiment guidance (**Fig. 4c**), automated Record entry (**Fig. 4d**), and automated data analysis (**Fig. 4e**). With Aira, users can define a research workflow and then receive step-by-step AI guidance tailored to their specific research goals, thereby accelerating and automating research (**Fig. 5**).

## 2.2 Community-driven Protocol design, sharing, and collaboration for data digitization

### 2.2.1 Digitizing lab data using customized Airalogy Protocols

Airalogy Protocols consist of customizable data fields (known as Airalogy Fields), enabling standardized, digitized records (Airalogy Records). The community-driven nature of Airalogy is evident in the flexibility of its Protocols: in addition to incorporating textual or multimedia content, researchers can also specify the types, default values, and constraints for data fields, and also cross-field assignment relationships (**Supplementary Table S4**). As a result, each Protocol functions both as an experimental guideline and as a tool for recording/analyzing data.

We illustrate the functionality of an Airalogy Protocol with a typical scenario. Alice, an experienced researcher, has designed an Airalogy Protocol for screening tumor-inhibiting drugs using the CCK-8 assay (hereafter referred to as "Alice's Protocol", the source code for defining this Protocol is available on GitHub at https://github.com/airalogy/examples). Meanwhile, Bob, a less experienced member of Alice's lab, can directly use Alice's Protocol—without needing to design one himself—to conduct experiments and record the resulting research data, thereby generating corresponding Airalogy Records ("Bob's Records") (**Fig. 2a**).

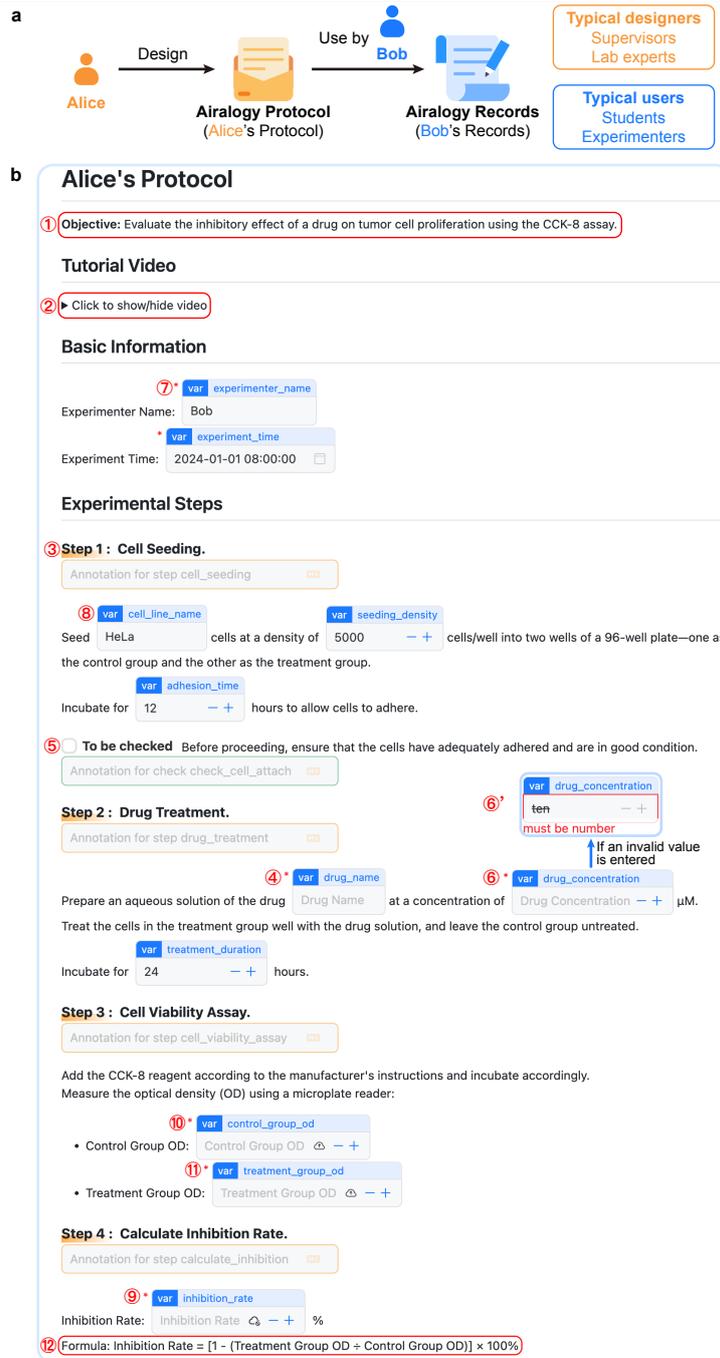

**Fig. 2: Standardized digital recording of research data using an Airalogy Protocol.**

**a,** Bob uses an Airalogy Protocol created by Alice to conduct an experiment and digitize the resulting data, producing corresponding Airalogy Records. On the Airalogy Platform, the individuals who design a Protocol and those who use it can differ significantly in expertise—for instance, a Protocol designer might be a seasoned laboratory supervisor, lab expert, or principal investigator, whereas users could be less experienced (e.g., students or junior experimenters). **b,** Schematic illustration of an Airalogy Protocol, with Alice's Protocol as an example. The Protocol can include textual content (①) and multimedia material (e.g., video, ②; for space considerations, the video content is folded here). Users can freely add Step-type (③), Variable-type (④, ⑥–⑧), and Checkpoint-type (⑤) Airalogy Fields, allowing key research data to be captured in a structured manner. Variable-type Fields support custom value constraints, and the Airalogy Platform automatically renders input boxes based on each Field's constraints—e.g., a text box for characters (④), a numeric box for numbers (⑥). If users enter values that violate predefined constraints (e.g., entering text instead of numbers), the Interface reports an error (⑥'), ensuring data adhere to the Protocol's standards. Variable-type Fields can also have default values (⑦, ⑧), reducing repetitive entries for commonly used parameters. Additionally, certain Fields may be automatically computed from others (e.g., a custom Field ⑨ can be calculated based on Fields ⑩ and ⑪, using a specified formula ⑫), thereby facilitating automated data entry and improving both efficiency and accuracy.

When Bob initiates a new Record based on Alice's Protocol, Airalogy creates a graph user interface (GUI) (**Fig. 2b; Supplementary Fig. S28**). In this Protocol, Alice included both textual descriptions (**Fig. 2b** ①) and multimedia content (e.g., video, **Fig. 2b** ②) to provide intuitive guidance for users. She defined various types of Airalogy Fields (such as Step, Variable, and Checkpoint) to capture different aspects of experimental information. For example, a Step-type Field with the ID "cell_seeding" (**Fig. 2b** ③) marks the first step in the experiment, a Variable-type Field "drug_name" (**Fig. 2b** ④) records the cell line in use, and a Checkpoint-type Field "check_cell_attachment" (**Fig. 2b** ⑤) prompts users to confirm critical experimental conditions.

Variable-type Fields can be assigned specific value constraints. The Airalogy Platform automatically renders the input boxes for these Fields in distinct styles based on their constraints, thereby facilitating efficient data entry (**Supplementary Fig. S31**). For example, the Field "drug_name" (**Fig. 2b** ④) is defined as a text, while "drug_concentration" (**Fig. 2b** ⑥) is defined as a number restricted to values above zero. If Bob enters data that violate these constraints (e.g., non-numeric text or negative values), the Interface displays an error and prompts for correction (**Fig. 2b** ⑥'), ensuring standardized data recording. To streamline frequently repeated inputs, Alice can define default values for Fields, which the Platform automatically populates whenever a new record is initiated—for instance, auto-filling the current user's name in "experimenter_name" (**Fig. 2b** ⑦) or setting the default cell line to "HeLa" (**Fig. 2b** ⑧). For highly variable Fields (e.g., "drug_name", **Fig. 2b** ④), no default is provided, requiring Bob to manually enter the value.

Because many experimental data Fields are interdependent, Airalogy allows users to specify assignment relationships among Fields so that certain values can be automatically computed based on others. For instance, the value of the "inhibition_rate" Field (**Fig. 2b** ⑨) can be calculated from the values of the "control_group_od" (**Fig. 2b** ⑩) and "treatment_group_od" Fields (**Fig. 2b** ⑪) using a specified formula (**Fig. 2b** ⑫). While this example is simple, an Airalogy Protocol can describe far richer dependency chains. It supports multi-level assignments, where the output of one calculation can be immediately used as the input for the next (**Supplementary Fig. S30**), as well as advanced logic and even AI/LLM-driven computations (**Supplementary Code S9**). In this way, a Protocol can not only be designed as a guide for experimental procedures and data recording, but also be used to facilitate automated digitization and data processing through programing and AI (**Case Study 3**).

Step-type Fields are automatically numbered (e.g., "Step 1", "Step 2", **Fig. 2b** ③) and include annotation boxes for users to optionally document unexpected observations or deviations encountered during each step (**Fig. 2b ③, Supplementary Fig. S32**). Checkpoint-type Fields similarly capture pass/fail outcomes and allow users to optionally annotate reasons or special considerations (**Fig. 2b ⑤, Supplementary Fig. S33**). This structured method of capturing step- and checkpoint-specific details facilitates post-experiment analysis of the factors contributing to experimental success or failure by directly correlating unusual events with these steps or critical points, thereby enabling rapid identification of anomalies and further optimization of the experimental Protocol (**Case Study 4**).

After Bob finishes data entry, submitting the Record (**Supplementary Fig. S28**), Airalogy assigns it a unique ID and timestamp (**Supplementary Code S17**), which protects its integrity. Should Bob need to revise a submitted Record, the Airalogy Platform creates a new copy with a separate unique ID, preserving the untampered history of the original data. This approach ensures that Airalogy Platform can serve as a robust evidence repository with time-stamped, immutable history of verifiable discovery. Moreover, version control applies not only to Records but also to the Protocol itself; each updated Protocol version is tracked (**Supplementary Code S12**), so that users or the community can maintain the latest experimental guidelines while retaining a full version history.

In addition to recording research data, each Airalogy Protocol is associated with a Discussions section (**Supplementary Figs. S37b, 38, 39**), which allows authorized Project members or public users to discuss issues related to the Protocol and historical Records. For example, although Alice's Protocol aims to standardize the experimental procedure, unexpected situations may arise that were not accounted for during protocol design. For instance, Alice set HeLa cells as the default tumor cell line at a seeding density of 5,000 cells/well (**Fig. 2b ⑧**). However, Bob may wish to screen drugs against another cell line, such as HepG2 for liver cancer, which likely requires a different seeding density. Since the Protocol itself does not address this specific scenario, Bob can turn to the Discussions section and post a question: "How many cells/well should I use for HepG2?" (**Supplementary Fig. S39**). Other users, such as Charlie, might offer concise answers (e.g., "8,000"), while Dave could provide a more detailed explanation, including his reasoning and further recommendations for various cell lines. Readers can vote on these answers, allowing higher-rated answers (often those deemed more comprehensive, such as the Dave's answer) to surface to the top.

## 2.2.2 Chaining multiple Protocols together into a Pipeline

Because Airalogy Records follow a predictable and standardized structure defined by Airalogy Protocols (**Supplementary Method S2.3**), data can be seamlessly transferred between Protocols. This means that data collected from one Protocol can easily serve as the input for a subsequent Protocol. By leveraging this feature, multiple Protocols—each potentially defined by different experts focusing on distinct aspects of research—can be chained together into an Airalogy Protocol Pipeline, thereby facilitating collaboration and enabling the completion of more complex research tasks.

Consider a typical scenario where different individuals in a laboratory possess varying expertise, making effective collaboration challenging. However, on the Airalogy Platform, experts with different skill sets—Alice, Frank, and Bob—can work together seamlessly (**Fig. 3a**). Alice, an experienced experimentalist, has designed a comprehensive experimental Protocol (**Fig. 2b**) but lacks data analysis expertise. In contrast, Frank, a seasoned data analyst with strong coding skills yet limited experimental experience, has developed a robust analytical Protocol (**Fig. 3b**). Bob, the primary experimental operator but a novice in Protocol design, can simply reuse Alice's Protocol to conduct five experiments on the tumor-inhibiting effects of Compound A at different concentrations in HeLa cells, thereby generating five corresponding Records (**Fig. 3c**). Bob then wishes to analyze these results to determine the half-maximal inhibitory concentration (IC50) of Compound A. Previously, such analysis would have required Bob to write new code from scratch. With the Airalogy Platform, however, he can effortlessly reuse Frank's analytical Protocol. To this end, Bob only needs to select the five Records generated from Alice's Protocol (**Fig. 3b ①**). Subsequently, Frank's Protocol automatically analyzes the selected Records, calculates the IC50 value (**Fig. 3b ②**), and generates the desired IC50 curve (**Fig. 3b ③**). Optionally, Bob can also then submit the analysis results, thereby creating a new Airalogy Record based on Frank's Protocol for future review or extended analysis. Moreover, data generated by one Protocol can be further analyzed using multiple downstream Protocols, each featuring distinct visualization or reporting schemes, which greatly enhances the efficiency and convenience of experimental operations, data analysis, and scientific reporting (**Supplementary Case Study S6**).

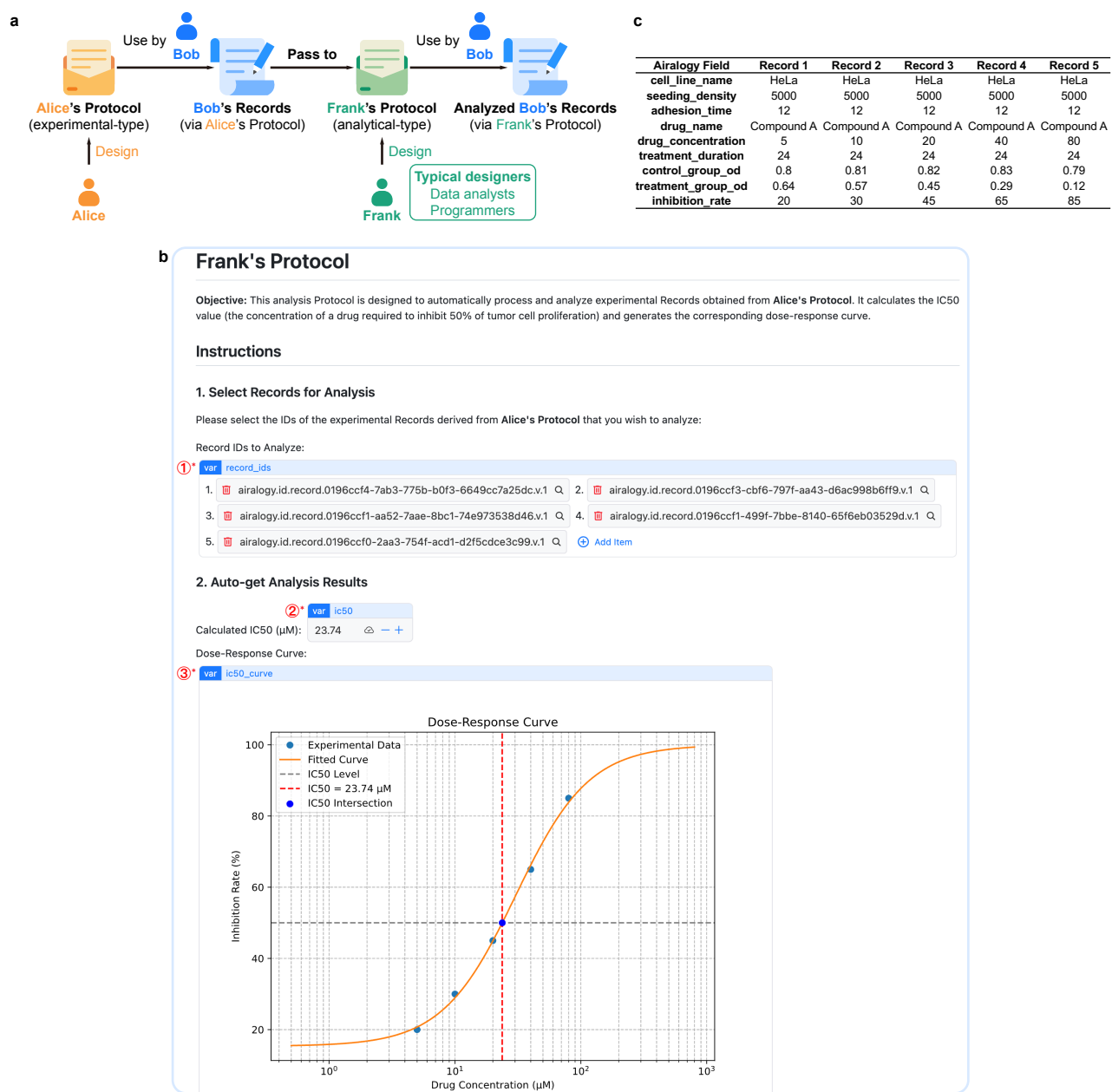

**Fig. 3: Chaining multiple Airalogy Protocols into a Pipeline for collaborative digitization.**

**a,** Different Airalogy Protocols, created by various users on the Airalogy Platform, can be linked into a Pipeline via Airalogy Records. Records generated by one Protocol may be designed to act as input for another Protocol. Here, Bob utilizes this mechanism to collaborate on Protocols developed by Alice (an experimental expert) and Frank (a data analysis expert). This approach not only supports standardized and electronic recording of experiments through the Platform but also enables automated data analysis of the resulting experimental outputs. **b,** Schematic illustration of an analysis-focused Airalogy Protocol, using Frank's Protocol as an example. When Bob applies Frank's Protocol, he only needs to select the relevant Records (generated from Alice's Protocol) in the "record_ids" Field (①). Frank's Protocol then automatically computes the IC50 value (②) from those Records and plots the IC50 curve (③). **c,** Partial data corresponding to specific Airalogy Fields in the five Records chosen by Bob in ①. In these five Records, Bob held other conditions constant while incrementally increasing the concentration of Compound A; as the drug concentration rose, the tumor cell inhibition rate increased accordingly.

### 2.2.3 Global sharing of Airalogy Protocols

On the Airalogy Platform, users may choose to make their Protocols publicly accessible, enabling others to search for and directly reuse them at the Airalogy Hub (**Supplementary Figs. S40, S41**). For example, Alice's and Frank's Protocols can be directly reused by researchers around the globe who are doing relevant studies to screen tumor-inhibiting drugs. The Airalogy Hub is essentially a repository for all publicly shared Protocols on the Platform, thereby serving as the first stop for novice scientists, who can reuse community-provided, pre-defined Protocols without having to start from scratch—thus enhancing both efficiency and quality. As more users share their research Protocols, Airalogy is expected to support all experimental record types currently available on platforms such as Benchling, while also extending to increasingly customized experiment types across diverse disciplines. AI search and recommendation features help user to find those that suit their needs (see **§2.3.2**). This design aims to foster a thriving global multidisciplinary research community ecosystem.

## 2.3 AI-driven research data digitization

As a multidisciplinary platform, Airalogy serves users with diverse backgrounds and varying levels of expertise in both science and information technology. To ensure that every user can leverage their domain knowledge without requiring advanced programing skills, we developed an AI copilot named "Aira", based on Large Language Models (LLMs). Serving as a reliable mentor and available 24/7, Aira is deeply integrated throughout the Airalogy Platform (**Supplementary Fig. S43**; the detailed implementation of Aira's AI methods is discussed in **Supplementary Methods**), offers a range of unique features, such as AI-automated/assisted Protocol design (**§2.3.1; Fig. 4a**), Protocol search (**§2.3.2; Fig. 4b**), research guidance (**§2.3.3; Fig. 4c**), research recording (**§2.3.4; Fig. 4d**), and data analysis (**§2.3.5; Fig. 4e**).

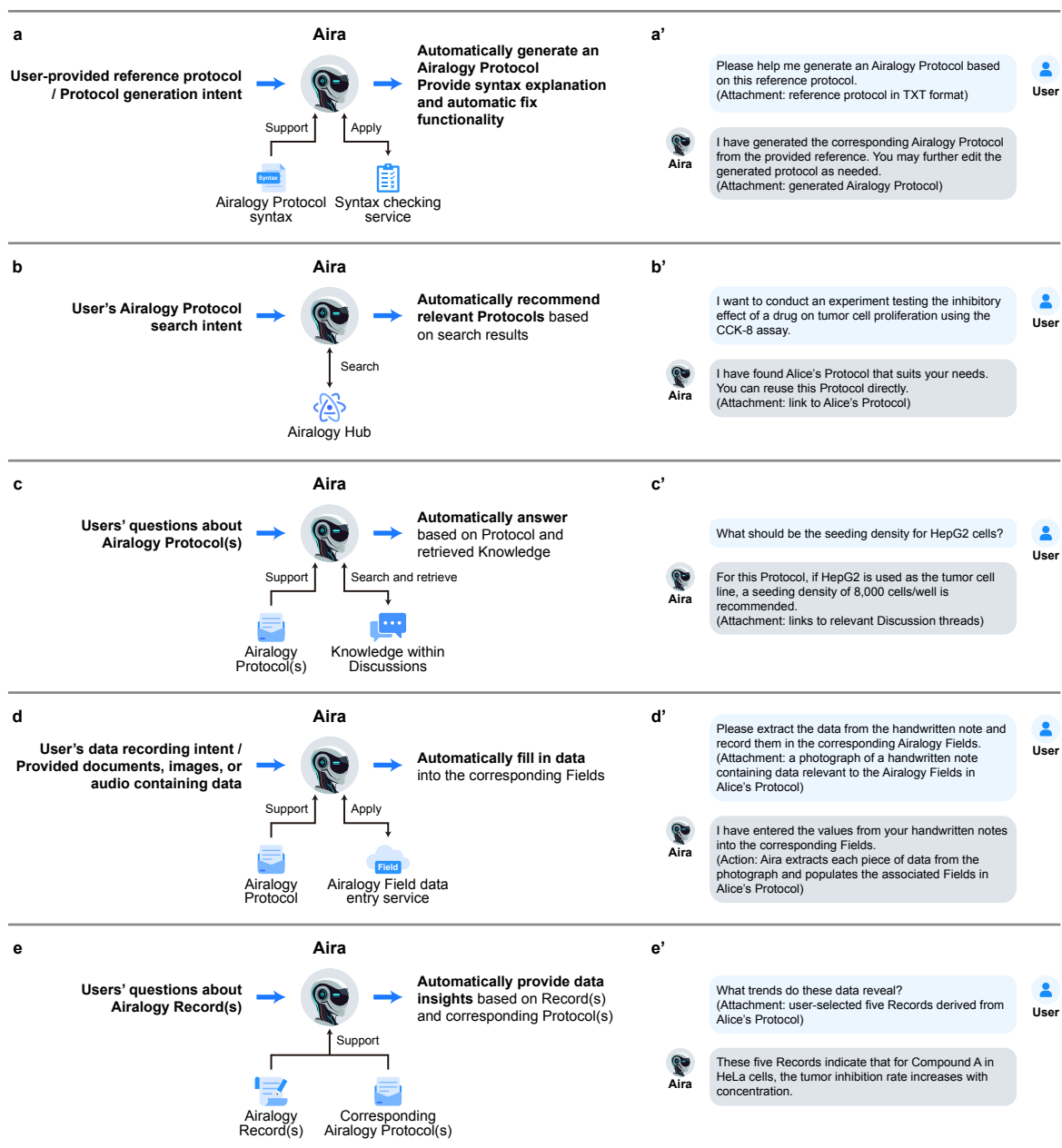

**Fig. 4: Aira, an AI research copilot, providing comprehensive intelligent support for the recording and utilization of Airalogy Protocols and Records.**

Aira's capabilities include: **a,** automated Protocol generation; **b,** intelligent search and recommendation of Protocols; **c,** Protocol-specific Q&A and guidance; **d,** automated standardization of multimodal research data entry; **e,** conversational intelligent analysis of Records. Panels **a'–e'** illustrate examples corresponding to each of these features, respectively. Detailed methodologies underlying Aira's functionalities are provided in **Supplementary Methods**.

## 2.3.1 AI-assisted Protocol design

To ensure that users from diverse disciplines can easily design Airalogy Protocols using only their domain expertise—regardless of their familiarity with computer programing—Aira provides an automated Protocol-generation feature. Leveraging an LLM-based agent system, Aira can generate all of the underlying code required to define a complete Protocol from either (1) a user-supplied reference document (e.g., a methods section copied from a published paper) or (2) a brief, plain-language description of the intended experiment (**Fig. 4a; Supplementary Fig. S44**). For instance, when Alice wishes to build the Protocol shown in **Fig. 2b**, she may start from scratch or simply paste a reference procedure in text; Aira responds by automatically generating the complete code for what becomes "Alice's Protocol" (**Fig. 4a'**). In addition, the Airalogy Platform features a user-friendly online editor for Airalogy Protocols (**Supplementary Fig. S42**), enabling expert users to conveniently refine the Aira-generated Protocol and further tailor it to their requirements. This editor also incorporates an AI-based dialog function: if users encounter difficulties with the customization syntax, they can consult Aira for guidance or request editing/modification suggestions.

### 2.3.2 AI-driven Protocol search

The Airalogy Hub provides access to pre-defined Airalogy Protocols shared by other researchers, allowing users to leverage existing Protocols instead of designing new ones from scratch. To further streamline the process of finding relevant Protocols, Aira offers a conversational search and recommendation feature (**Fig. 4b; Supplementary Fig. S45**). For example, when a novice user asks, "I want to conduct an experiment testing the inhibitory effect of a drug on tumor cell proliferation using the CCK-8 assay", Aira automatically interprets the query and retrieves Alice's Protocol (**Fig. 4b'**). The user can then quickly locate and reuse the Protocol with a single click, avoiding the need to manually browse through the entire Hub.

### 2.3.3 AI-guided research

To ensure high-quality digital research records, users must fully understand both the scientific rationale and the procedural steps of the Protocols they follow. Otherwise, experimental quality might be compromised, resulting in suboptimal data. On the Airalogy Platform, the creators of an Airalogy Protocol and its users can differ in expertise (e.g., a Protocol may be designed by an expert, while used by a novice who is not fully familiar with its details). To fill the gap and expedite recording, Aira provides intelligent, on-demand guidance for Protocol users. When a user accesses a Protocol, Aira automatically references the Protocol's content and, if applicable, searches the associated Discussions section for community-sourced expert insights (**Fig. 4c; Supplementary Fig. S46**). For example, if Bob, while using Alice's Protocol, asks, "What should be the seeding density for HepG2 cells?", Aira automatically retrieves relevant Discussion threads (e.g., Dave's recommendation for HepG2 seeding density within this Protocol; **Supplementary Fig. S39**) and immediately replies, "For this Protocol, if HepG2 is used as the tumor cell line, a seeding density of 8,000 cells/well is recommended." (**Fig. 4c'**), sparing Bob the wait for a human mentor's response. Leveraging the broad, cross-disciplinary knowledge of LLMs in tandem with community-contributed expertise, Aira delivers real-time, intelligent guidance to users of Airalogy Protocols across diverse scientific fields.

### 2.3.4 AI-automated research recording

Each Airalogy Protocol provides a GUI (**Fig. 2b**) that allows users to intuitively and rapidly enter data into designated Airalogy Fields, producing standardized Records. However, in many real-world scenarios, certain research data are initially recorded in non-digital or non-standard formats. Researchers may be unable to use a computer during certain experimental conditions (requiring them to note information on paper first) or they may already have large amounts of unstructured electronic records (e.g., Word documents) that need to be converted into standardized Airalogy Records. To address these needs, Aira supports the automated conversion of multimodal data into standardized Records through conversation. Users can provide information in text, speech, images (e.g., photographs of handwritten notes or instrument displays), or documents (e.g., historical Word-based records) within a chat session. Aira leverages an LLM agent system to deeply understand data based on the active Airalogy Protocol and automatically fills in the corresponding Fields to generate a standardized Record (**Fig. 4d; Supplementary Figs. S47, S48**). For example, when conducting a cell experiment under Alice's Protocol, Bob may have to record data on paper if electronic devices are not permitted inside a biosafety cabinet. Upon leaving the cabinet, he can initiate a new Record under Alice's Protocol and upload a photograph of the handwritten notes to Aira, which then identifies and enters the data into the appropriate Fields (**Fig. 4d'**).

### 2.3.5 AI-assisted data analysis

Beyond digitizing research data across multiple disciplines, the Airalogy Platform also promotes the intelligent utilization of these electronic Records. Each Airalogy Record (generated from an Airalogy Protocol) has a well-defined data structure that directly corresponds to its parent Protocol (**Supplementary Method S2.3**), enabling AI to easily understand and interpret historical Records. Consequently, Aira provides a conversational Q&A feature that allows users to select past Records and pose questions about them. Aira automatically analyzes the selected Records, delivering insights in real time (**Fig. 4e; Supplementary Fig. S49**). This process enables users to rapidly identify trends and gain insights from historical data, further enhancing the value of data digitization. For example, after Bob generates five Records using Alice's Protocol (**Fig. 2c**), he can add these Records to Aira's conversation context and ask, "What trends do these data reveal?" Aira interprets the Records and their associated Protocol, replying, "These five Records indicate that for Compound A in HeLa cells, the tumor inhibition rate increases with concentration." (**Fig. 4e'**).

## 2.4 AI-driven research workflow automation

A single Airalogy Protocol typically corresponds to a specific experimental procedure, yet most research goals require combining multiple Protocols into a comprehensive workflow, possibly containing diverse Airalogy Protocols from various disciplines (**Fig. 5a**). Each Protocol focuses on a specific experimental or research activity and can be reused in similar contexts, much like individual building blocks. By linking these Protocols through a logical pipeline, researchers form an Airalogy Protocol Workflow tailored to a particular research goal (**Fig. 5b**), thus enabling systematic data recording across the entire process and streamlining the corresponding research endeavor.

Consider a real-world scenario involving carbon nanotube (CNT) dispersion[36,37] aimed at achieving a target average tube diameter. To meet this research goal, four Protocols can be assembled into a Workflow (**Fig. 5b**):

- Protocol 1: Preparing a CNT dispersion from CNT powder
- Protocol 2: Ultrasonic dispersion of CNTs
- Protocol 3: Diluting dispersed CNTs
- Protocol 4: Characterizing dispersion results via scanning electron microscopy (SEM)

Unlike a simple Pipeline, a Workflow does not necessary follow a strictly linear sequence, but can contain loops; Protocols may be invoked repeatedly depending on evolving research outcomes and the goals. For instance, aiming to "obtain a high-concentration CNT dispersion with a diameter of around 20 nm" might involve multiple cycles of ultrasonic dispersion and characterization (i.e., Protocol 4 → Protocol 2 → Protocol 4) or switching to dilution and further ultrasonication (Protocol 4 → Protocol 3 → Protocol 2 → Protocol 4) (**Fig. 5b**). Determining whether to use Protocol 2 or Protocol 3 next depends on the experimental results. If repeated ultrasound dispersion (Protocol 4 → Protocol 2 → Protocol 4) does not reduce the CNT diameter further, switching to Protocol 3 (dilution) may be necessary before repeating Protocol 2 again.

Aira employs LLM agents to guide the execution of arbitrary Workflows and achieves universal AI research automation—if automated research devices are provided (**Supplementary Case Study S1**; in this article, the participating labs performed the AI-suggested experiments manually). Here, although each Protocol within a Workflow may be independently designed by different experts, even across disciplines (and thus may not always align seamlessly, much like mismatched building blocks), Aira's LLM-based foundation provides it with broad cross-disciplinary knowledge, enabling it to serve as a universal "adapter" that integrates these Protocols (**Fig. 5c**). Regardless of the Protocols' configurations, Aira can automatically propose research goals, plan strategies, dynamically select the next suitable Protocol based on real-time results, set parameters, analyze the emerging data, and continue iterating until the research goal is achieved. Once the goal is met, Aira produces a final research conclusion (**Fig. 5d**).

When running this Workflow in a Westlake laboratory (illustrated in **Fig. 5e** and detailed in **Supplementary Case Study S1**), suppose the goal is to disperse CNTs in m-cresol solution to an average diameter of 10–30 nm. Aira can design the research strategy by starting from an initial 1 mg/mL CNT suspension and recommending repeated ultrasonication until no further improvement is observed, followed by dilution and additional ultrasonication to reach the target diameter. As the experiment proceeds, Aira automatically analyzes the characterization results from each run of Protocol 4 to determine whether the 10–30 nm range has been achieved, then selects the subsequent Protocol or parameter adjustments. Upon meeting the target, Aira terminates the research (End) and generates a final research conclusion. Additionally, each time a Protocol is applied during the Workflow, a corresponding Airalogy Record is created, yielding a fully standardized list of Records. Through Aira-enabled research automation, Airalogy not only greatly facilitates researchers' systematic data collection but also lowers barriers and improves efficiency in conducting systematic scientific studies that meet real-world research needs—thereby accelerating scientific progress.

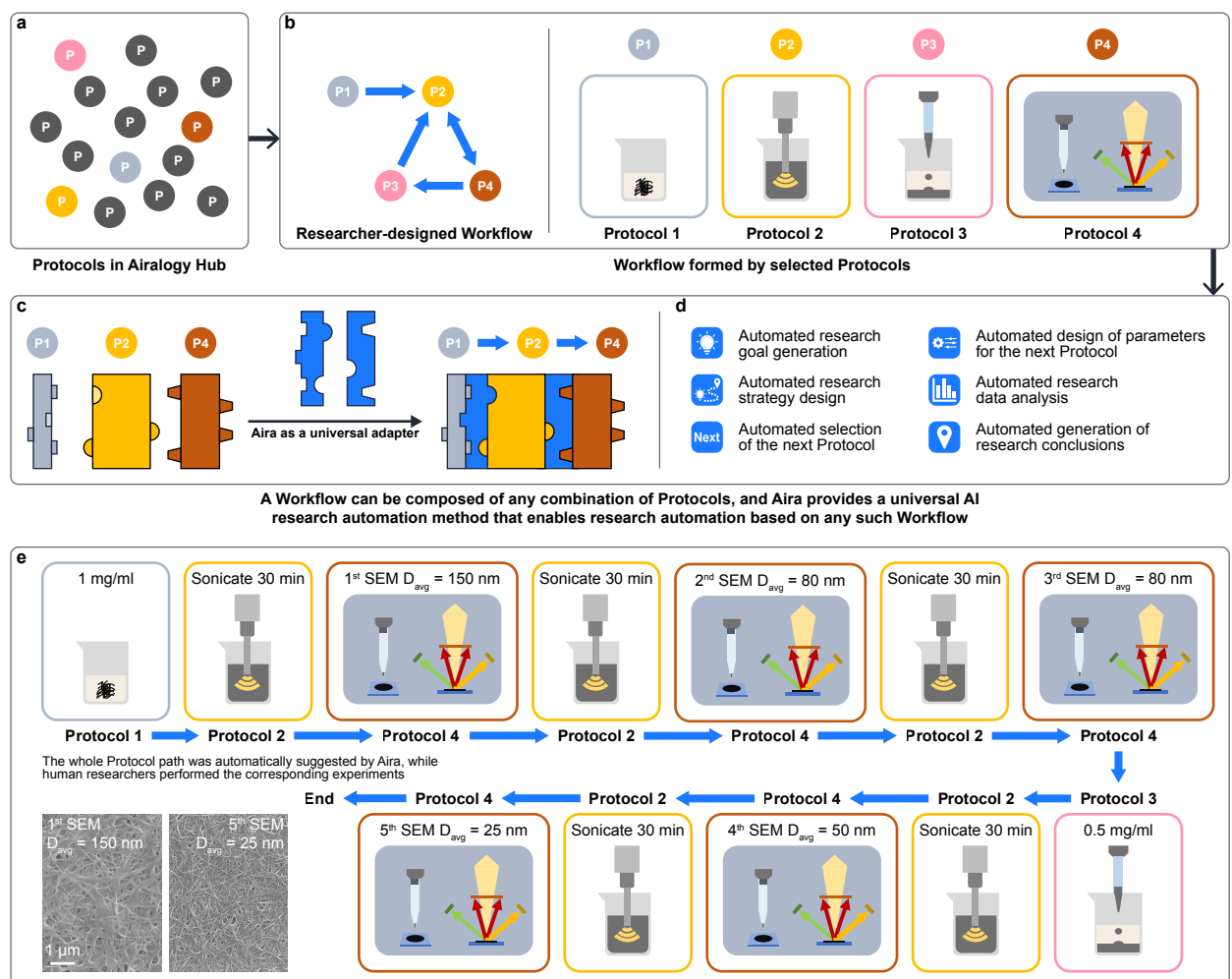

**Fig. 5: Systematic, automated research through the Airalogy Protocol Workflow.**
**a,** The Airalogy Hub hosts Airalogy Protocols customized by users from various disciplines. **b,** For a given research goal, these Protocols can be combined into an Airalogy Protocol Workflow. Shown here is a Workflow for dispersing carbon nanotubes to a target average diameter using four Protocols: Protocol 1 (Preparing a CNT dispersion from CNT powder), Protocol 2 (Ultrasonic dispersion of CNTs), Protocol 3 (Diluting dispersed CNTs), and Protocol 4 (Characterizing dispersion results via SEM). Because these Protocols may be designed by different experts, they can be difficult to integrate directly; however, Aira (the AI research copilot of the Airalogy Platform) provides a AI-based method to act as a universal adapter (**c**), bridging the Protocols and automating key Workflow steps (**d**). **e,** An example of how this Workflow is applied in a real laboratory setting.

# 3 Discussions

Through Airalogy, we explore a novel paradigm for electronic research data that aligns with current AI development trends. Anchored in a data-centric foundation, this AI- and community-driven approach goes beyond mere digitization to deliver far broader impacts. Already adopted in all four schools at Westlake University (Engineering, Life Sciences, Science, and Medicine), the broader impact of this paradigm is illustrated through six representative case studies from various laboratories (**Fig. 6**). These include: promoting the global sharing of cutting-edge research (**Case Study 1; Fig. 6a**), enabling intelligent teaching (**Case Study 2; Fig. 6b**), developing AI applications to empower research (**Case Study 3; Fig. 6c**), fostering Protocol–Record co-evolution to refine specific SOPs (**Case Study 4; Fig. 6d**), providing a one-stop solution for data digitization, management, and utilization (**Case Study 5; Fig. 6e**), and introducing a new paradigm of global AI–human collaboration and division of labor (**Case Study 6; Fig. 6f**). We anticipate that Airalogy will significantly advance research digitization and automation across multiple disciplines, thereby accelerating scientific and technological innovation, ultimately benefiting humanity as a whole.

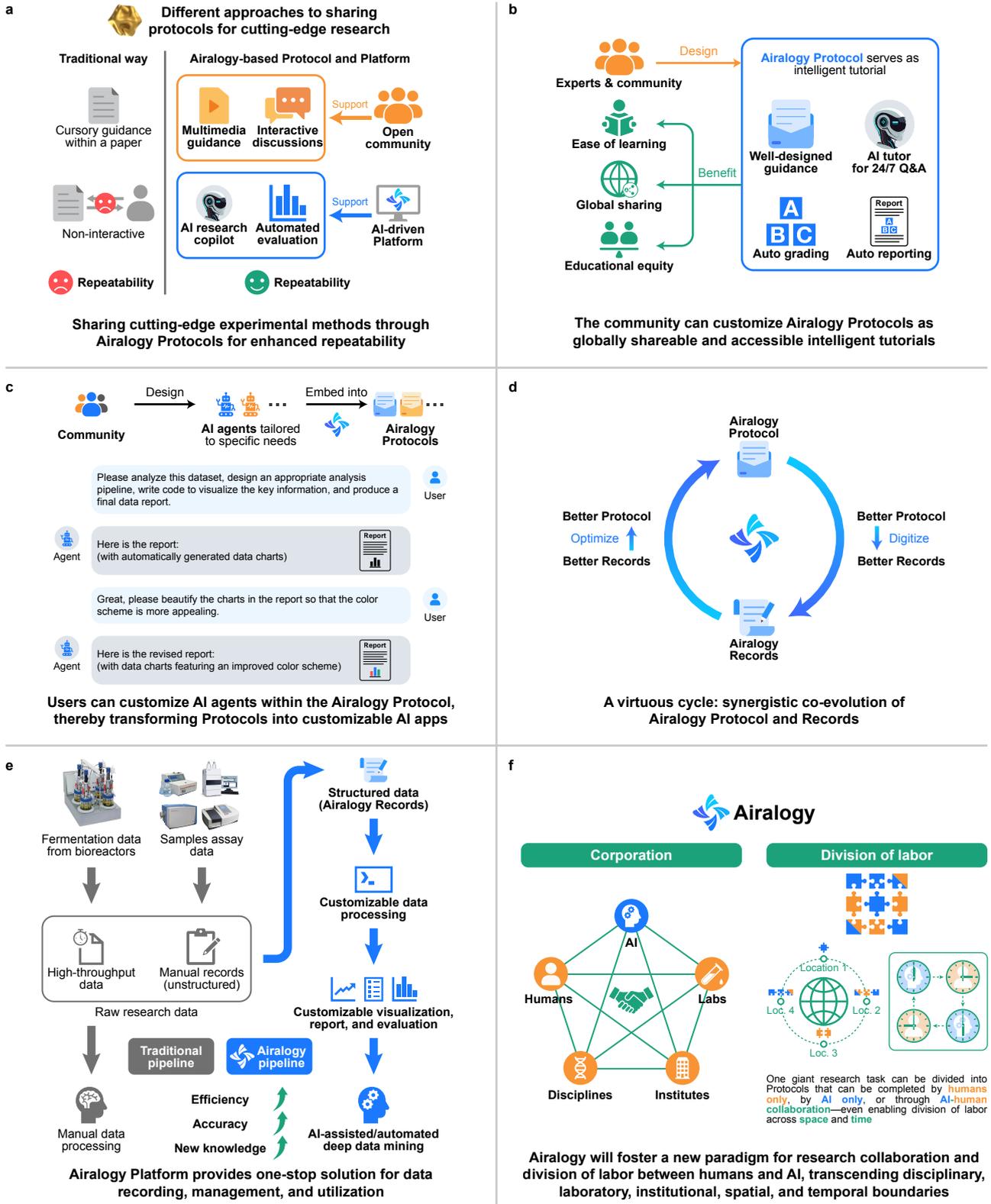

**Fig. 6: Broader impacts of Airalogy.**
Using six representative case studies from laboratories across the four schools of Westlake University, we illustrate the wide-reaching influence of Airalogy. Panels **a–f** correspond to **Case Studies 1–6** (and **Supplementary Case Studies 2–7**), with further details provided in each respective case study.

## Case Study 1: Sharing cutting-edge scientific research

Reproducibility is a cornerstone of scientific research[38]. Only by independently verifying experimental results can researchers ensure credibility and build a strong foundation for continued innovation. However, in practice, many researchers struggle to replicate published findings. Surveys have shown that over 70% of scientists fail to reproduce other researchers' results[39]. Two major contributors to this reproducibility crisis are (1) insufficient detail in published methods and (2) the static, text-based presentation of protocols, which prevents researchers from troubleshooting in real time if questions arise.

Existing solutions only address part of the problem. For example, specialized journals like *Nature Protocols*[40] offer peer-reviewed protocols with detailed experimental instructions, but lengthy review processes can delay sharing of the latest methods. Other emerging platforms (e.g., Protocol.io[41]) allow rapid uploads and user feedback, yet the absence of interactive elements still makes it difficult for researchers to evaluate the success of their experiments.

In contrast, the Airalogy Platform provides new approaches to resolving reproducibility challenges:

1. **Multimedia guidance:** Users can upload fully customized Airalogy Protocols, including text, images, audio, and video, to provide robust, detailed guidance.
2. **Interactive discussions:** Each shared Airalogy Protocol has a built-in Discussions section where other researchers can ask questions; the Protocol's creators/maintainers can respond promptly, fostering global collaboration.
3. **AI research copilot:** By linking Protocols with experimental Records, the AI research copilot, Aira, can serve as a virtual mentor—offering round-the-clock guidance and automated data analysis. This feature supports novice researchers and provides immediate, data-driven troubleshooting.
4. **Automated evaluation:** Because users can flexibly define data fields (Airalogy Fields) and the computational relationships among them, they can promptly enter experimental results in these Fields. Protocol creators can integrate automated analysis and evaluation methods (even leveraging AI/LLMs) to provide real-time feedback on experimental outcomes. As a result, users receive not only methodological instructions but also immediate guidance on whether the experiment has succeeded.

As more researchers adopt and refine a given Protocol, it evolves from a static document into a continuously improving resource (see **Case Study 4**). This "living" Protocol paradigm surpasses the limitations of single-publication models.

In **Supplementary Case Study S2**, we demonstrate these advantages by converting our newly published protocol for synthesizing gold nanopropellers[26] into an Airalogy Protocol for global sharing. Notably, the Protocol integrated a multimodal large language model that allows users to upload images of their synthesized gold nanopropellers. The model automatically assesses whether the experiment was successful based on the uploaded images, providing immediate feedback and enhancing reproducibility. This example demonstrates how Airalogy's approach not only accelerates the dissemination of cutting-edge methodologies but also elevates the overall quality of shared protocols.

## Case Study 2: Bridging research and education

Research and education are inseparable: high-quality research hinges on cultivating innovative talent, and education directly shapes future generations of scientists. Modern research often requires newcomers to master many experimental methods and operate increasingly complex instruments[42], but in a real laboratory setting, busy supervisors or senior researchers may not always be able to provide hands-on teaching or answer questions in real time—especially when working with multiple students. Consequently, newcomers can struggle with new experiments or equipment, delaying research progress. Airalogy offers a novel solution to these challenges: by leveraging Airalogy Protocols' flexible customization, users can design "intelligent tutorials" that address specific educational scenarios and enable AI-powered, efficient scientific training.

In **Supplementary Case Study S3**, we present a real-world example focusing on instrument-based teaching in the shared instrumentation facility at Westlake University. This facility houses numerous large-scale instruments, used daily by many students and faculty with varying levels of expertise—yet the limited number of technical staff can make training difficult. As a solution, we created a teaching-focused Airalogy Protocol for a flow cytometer, helping new users quickly learn flow cytometry experiments. First, the teaching Protocol can include detailed instructions (text, images, demonstration videos) to give novices a clear operational framework for the instrument. By adding appropriate Airalogy Fields, instructors can record and track key instrument parameters and user-generated experimental results, which can then be analyzed to assess a student's progress. Second, because Protocols support custom AI integrations, users can embed automated assessment features, which evaluate each training session based on data recorded in the Fields, providing immediate insights into each learner's proficiency. For instance, the Protocol can automatically generate a radar chart that visualizes a student's skill in multiple dimensions, allowing them to understand their progress more intuitively. Third, every training session recorded through the teaching Protocol generates an Airalogy Record that serves as a learning report, helping students easily review their development. Fourth, thanks to the AI-driven nature of the Airalogy Platform, its 24/7 AI copilot, Aira, serves as a virtual tutor, offering real-time troubleshooting and adaptive feedback. Students can ask questions at any time (without concern about asking "stupid questions") and reduce the teaching burden on trainers.

Notably, every Airalogy Protocol can be globally shared and reused across the Platform, extending high-quality instruction to labs with fewer training resources and helping level the educational playing field. Laboratories can adopt (and adapt) expertly crafted tutorials from others, while Aira's consistent, high-level support further reduces teaching disparities.

## Case Study 3: Customizing AI research agents and AI scientist

As AI capabilities advance, researchers increasingly seek deeper, automated assistance to accelerate their work. To meet this demand, Airalogy already provides Aira—its AI research copilot or co-scientist that supports widely needed tasks such as Protocol search and design, real-time guidance, and intelligent data recording and analysis (**Fig. 4**). Yet, because Airalogy is a multidisciplinary platform, it cannot possibly anticipate and tailor AI methods for every discipline-specific or highly specialized research need. Fortunately, Airalogy's community-driven paradigm and the Protocols' well-defined syntax—which let users flexibly specify computational relationships among Airalogy Fields (including calls to LLMs; **Supplementary Code S9**)—enable community members to harness their creativity to build specialized "AI applications" on top of these Protocols, serving as AI research agents or "AI scientists" that address niche as well as broadly applicable scientific challenges.

In **Supplementary Case Study S4**, we showcase how Airalogy Protocols can be designed as AI applications. One example features a no-code data analysis tool: users upload a CSV file containing research data and specify their analytical goals; the Protocol's embedded LLM-based AI agent then identifies the dataset's structure, generates Python scripts for analysis, executes them automatically, and outputs a report with data charts. If the user is unsatisfied, they can provide further feedback in natural language (e.g., requesting new chart colors), and the agent updates the report accordingly. Another case involves integrating DeepReview[43] (our recently published AI agent which can automatically generate review comments for a given paper), into an Airalogy Protocol. After uploading a paper, the agent generates reviewer comments automatically without requiring human intervention. Finally, we also demonstrate a Protocol for controlling automated lab equipment, such as the Opentrons OT-2 liquid-handling robot. Users simply describe their pipetting requirements in natural language; the embedded AI agent translates these instructions into device-specific commands, sends them to the OT-2, and automatically logs the resulting experimental data in the Protocol once the task is complete.

Crucially, because these AI applications adhere to Airalogy Protocol syntax, they can be shared and reused globally like any other Protocol. As community members with diverse AI backgrounds create and share more specialized AI Protocols within the Airalogy Hub, the Hub effectively becomes an "AI app marketplace". We anticipate that, as the Airalogy community continues to flourish, an expanding array of scientific needs will be addressed with intelligent, automated solutions built upon Airalogy Protocols and AI. Moreover, thanks to the shareability of Airalogy Protocols, these AI applications will significantly benefit the global research community.

## Case Study 4: Standardizing and iteratively refining operating procedures

Protocols are crucial for guiding experiments; an inadequately designed protocol can undermine researchers' ability to reproduce or achieve consistent, high-quality results. Traditionally, protocols are disconnected from the data gathered through their use, making it difficult to improve protocols by learning from past experimental outcomes.

In contrast, Airalogy establishes a new paradigm by tightly linking Airalogy Protocols with the Airalogy Records generated from them. Because every Airalogy Record is produced by applying a specific Protocol, the quality of those Records indirectly but strongly reflects the effectiveness of the Protocol itself. A robust Protocol typically helps researchers perform experiments more reliably, leading to higher-quality data. By analyzing the Records produced from a Protocol, researchers can identify shortcomings and introduce targeted improvements, creating a virtuous cycle: Protocol refinements yield better Records, which in turn provide insights for further optimization. Over time, these iterative improvements may elevate a Protocol to the level of a reliable standard operating procedure (SOP).

In **Supplementary Case Study S5**, we illustrate this approach using an Airalogy Protocol for purifying lipoylated H protein[44]. The initial version of the Protocol was less than optimal. However, by applying this preliminary Protocol and generating multiple Records (which documented any discrepancies or problems encountered during the experimental steps, as well as the data produced at each stage), we were able to evaluate these Records to identify inefficiencies and opportunities for improvement. We then introduced targeted modifications, such as more precise instructions or additional Airalogy Fields to capture critical parameters. Repeating this optimization cycle over several iterations rendered the Protocol increasingly robust and transparent, ultimately yielding a high-reproducibility Protocol that not only supports high-quality experiments but also produces correspondingly high-quality Records.

Notably, this Framework not only strengthens laboratory research but also holds promise for industrial applications, where SOPs demand strict validation and ongoing updates. Through Airalogy Protocols, industrial teams can implement rapid, data-driven improvements, accelerating their move toward highly optimized and reliable Protocols for industrial use.

## Case Study 5: Automating research data management and analysis

Ongoing advances in automated, high-throughput instrumentation have triggered an exponential increase in the volume of scientific data, propelling research into a big data era. Moreover, these data are becoming increasingly complex, characterized by multimodality and unstructured formats. Effectively handling and standardizing data from diverse laboratories, instruments, and researchers is therefore paramount for modern research.

In **Supplementary Case Study S6**, we showcase how the Airalogy Platform supports industrial strain fermentation research[45], enabling the intelligent and automated transformation of traditional workflows, and offering a one-stop solution for recording, managing, utilizing research data. By using Airalogy Protocols specifically designed for this context, the Platform captures multimodal data and converts them into standardized, experimental Airalogy Records. These Records can then undergo subsequent handling—such as analysis, visualization, and reporting—in a flexible, automated, and scalable manner. Through analysis-oriented Airalogy Protocols (integrated with custom code and embedded LLM-based methods; see **Case Study 3**), key parameters from fermentation experiments can be automatically processed, visualized, and summarized into insights before being consolidated into a comprehensive research report. Researchers can then quickly and intuitively grasp the relevant findings simply by reviewing this data report. Although designing these Protocols requires substantial data-analysis expertise (due to the complexity of data processing methods), the Airalogy Platform's shareability ensures that, once created, such automated pipelines for data recording, preprocessing, analysis, and reporting become easily accessible to other lab members. Even those without programming skills can utilize these pipelines without difficulty, drastically reducing the time and labor associated with manual data collection, analysis, visualization, and organization. This Airalogy-based research approach significantly boosts the speed, efficiency, and quality of data-driven science.

## Case Study 6: Fostering global collaboration and division of labor

It has been a well-established economic principle that dividing labor into specialized tasks dramatically increases productivity[46], which profoundly shaped modern society. Similarly rules also apply to modern scientific research[47]. The Airalogy Platform offers a novel paradigm for achieving global scientific cooperation (see **Supplementary Case Study S7** for a detailed discussion).

The core of division of labor lies in breaking down a large task into discrete modules, then coupling them through standard interfaces or coordination mechanisms to create a synergistic "1 + 1 > 2" effect[48,49]. Within the Airalogy Platform, each Airalogy Protocol functions as a minimal unit or module for a specific research activity; these modules can then be assembled into a Pipeline (**Fig. 3**) or Workflow (**Fig. 5**), using data (Airalogy Records) as the binding interface. This approach allows a large-scale research undertaking to be subdivided into multiple, relatively independent Protocols that still function cohesively.

1. **Collaboration beyond isolation:** Traditionally, researchers work in relative isolation, limiting synergy. In contrast, Airalogy consolidates data and knowledge (e.g., Protocols, Records, Discussions) from diverse individuals, disciplines, laboratories, and institutions across the globe into a single Platform, thereby enabling seamless sharing across the community—surpassing individual, spatial, and temporal barriers.
2. **Cross-disciplinary expertise:** Different Protocols can be authored and shared by researchers with varied, cross-disciplinary skill sets, allowing simultaneous collaboration and harnessing a broad spectrum of expertise.
3. **Flexible task allocation:** Each Protocol's independence lets separate parts of a project be performed by different human researchers or AI (potentially at different locations and times) before reintegration.
4. **AI-driven future:** As AI continues to advance (already automating tasks like literature reviews[50], idea generation[51], experimental design[6], manuscript drafting[17], and peer-review feedback[16]), future collaborations will no longer be confined to human–human interactions but will increasingly involve AI–human and even AI–AI partnerships. Accordingly, Airalogy's AI copilot, Aira (**Fig. 4**), and the ability to embed user-defined AI into Protocols (**Case Study 3**) mark initial steps toward integrating AI fully into research workflows, relieving researchers' workloads and speeding scientific discovery. Airalogy is thus well-positioned for an AI-driven research landscape.

Taken together, Airalogy's impact goes beyond functioning as a research platform for digitizing and automating; it compels a fundamental rethink and redesign of how scientific work is organized in the AI era. With AI poised to become the masterbrain of research[13], Airalogy lays a foundation for future paradigms of AI–human collaboration and division of labor, ultimately redefining research practices[52], boosting the efficiency, quality, and speed of academic and industrial research, and elevating the scientific competitiveness of universities, enterprises, and even nations—thereby benefiting the global research community.

# 4 Methods

The methods related to this study are detailed in **Supplementary Methods**.

# 5 Data availability

Data from the Airalogy Platform can be accessed at https://airalogy.com.

# 6 Code availability

The Airalogy suite and the Platform's AI research copilot, Aira, are available on GitHub (https://github.com/airalogy/airalogy, https://github.com/airalogy/masterbrain); Relevant Python packages have been released on PyPI (https://pypi.org/project/airalogy, https://pypi.org/project/masterbrain). All Airalogy Protocols used in this study are available on GitHub (https://github.com/airalogy/examples) and the Airalogy Platform (https://airalogy.com/labs/airalogy/projects/paper/protocols).

# 7 Acknowledgments

We thank Zhao Wang, Ronglu Zhou, Qian Yu, and Hongji Zhang for their support in the Airalogy Platform's development and artistic design; Fan Zhang, Li Huang, Mingdu Chen, Jingyu Miao, and Qiulin Ding for their support in launching the Platform; Jiawen Xu, Linyi Yang, and Yingjie Li for providing financial management support; Xue Cai, Xuedong Zhang for assisting with the configuration of the Opentrons OT-2 used in Supplementary Case Study S4; Minjun Zhu for providing the API for DeepReview, the AI-automated peer-review system used in Supplementary Case Study S4; Shuhan Zhao for assisting with the implementation of the Airalogy Protocol for clinical-dialogue audio transcription used in Supplementary Case Study S4; Fan Guo, Jingyi Li, Yiming Sun, Yukai Wang, Yongfei Liu, Wei Zhuang, Chunjun Zhan, An Su, Jie Huang, Qiuping Wu, Xuan Ding, Qingkai Min, Yuanshi Wang, Yinghao Yang, Yifeng Lu, and Xi Jin for their discussions and feedback on this project. This work was supported by the Research Program No. WU2023C020 of Research Center for Industries of the Future, Westlake University. We gratefully acknowledge Westlake University, its Research Center for Industries of the Future, and all four of its schools (School of Engineering, Life Sciences, Science, and Medicine) for their support of this work.

# 8 Contributions

Zijie Yang, Fang Guo, Qiji Zhou, and Yue Zhang conceived this study. Zijie Yang designed and implemented the Airalogy-related framework, syntax, architecture, and AI research automation methods. Zijie Yang, Qijie Zhou, Fang Guo, Panzhong Lu, and Junshu Pan designed the AI methods for Aira; Zijie Yang, Panzhong Lu, Junshu Pan, Qijie Zhou, and Fang Guo completed the implementation of Aira. Zijie Yang led the inter-laboratory collaboration, division of labor, and design of the case studies; Sijie Zhang, Yexun Xi, and Mingtao Chen designed and completed the Supplementary Case Study S1; Liping Huang and Yingming Pu designed and completed the Supplementary Case Study S2; Xiaoyu Ma designed and completed the Supplementary Case Study S3; Yonghe Xia designed and completed the Supplementary Case Study S4; Yudian Zhu designed and completed the Supplementary Case Study S5; Jinglei Nie designed and completed the Supplementary Case Study S6; Chou Wu and Jinglei Nie designed and completed the Supplementary Case Study S7; Panzhong Lu and Junshu Pan assisted with the coding for the Airalogy Protocols used in these case studies. Tiannan Guo, Yanmei Dou, Hongyu Chen, Anping Zeng, Jiaxing Huang, Tian Xu, and Yue Zhang supervised the study. All authors contributed to writing the manuscript.

# 9 Conflict of interests

Zijie Yang, Qijie Zhou, Fang Guo, Panzhong Lu, Junshu Pan, Tian Xu, and Yue Zhang hold the relevant patents (202510086469X, 2025100864670) for the Airalogy. The other authors declare no competing interests.

# 10 References

1      Jumper, J. *et al.* Highly accurate protein structure prediction with AlphaFold. *Nature* **596**, 583-589 (2021).

2      Lin, Z. *et al.* Evolutionary-scale prediction of atomic-level protein structure with a language model. *Science* **379**, 1123-1130 (2023).

3      Stokes, J. M. *et al.* A deep learning approach to antibiotic discovery. *Cell* **180**, 688-702. e613 (2020).

4      Du, Y. *et al.* Machine learning-aided generative molecular design. *Nature Machine Intelligence* **6**, 589-604 (2024).

5      Granda, J. M., Donina, L., Dragone, V., Long, D.-L. & Cronin, L. Controlling an organic synthesis robot with machine learning to search for new reactivity. *Nature* **559**, 377-381 (2018).

6      Boiko, D. A., MacKnight, R., Kline, B. & Gomes, G. Autonomous chemical research with large language models. *Nature* **624**, 570-578 (2023).

7      Szymanski, N. J. *et al.* An autonomous laboratory for the accelerated synthesis of novel materials. *Nature* **624**, 86-91 (2023).

8      Merchant, A. *et al.* Scaling deep learning for materials discovery. *Nature* **624**, 80-85 (2023).

9      Xu, H. *et al.* A whole-slide foundation model for digital pathology from real-world data. *Nature* **630**, 181-188 (2024).

10     Lu, M. Y. *et al.* A multimodal generative AI copilot for human pathology. *Nature* **634**, 466-473 (2024).

11     Yang, Z. *et al.* ShennongAlpha: an AI-driven sharing and collaboration platform for intelligent curation, acquisition, and translation of natural medicinal material knowledge. *Cell Discovery* **11**, 32, doi:10.1038/s41421-025-00776-2 (2025).

12     Novikov, A. *et al.* AlphaEvolve: A coding agent for scientific and algorithmic discovery. *Google DeepMind* (2025).

13     Yang, Z., Wang, Y. & Zhang, L. AI becomes a masterbrain scientist. *bioRxiv*, 2023.2004. 2019.537579 (2023).

14     Lu, C. *et al.* The ai scientist: Towards fully automated open-ended scientific discovery. *arXiv preprint arXiv:2408.06292* (2024).

15     Gottweis, J. *et al.* Towards an AI co-scientist. *arXiv preprint arXiv:2502.18864* (2025).

16     Weng, Y. *et al.* Cycleresearcher: Improving automated research via automated review. *arXiv preprint arXiv:2411.00816* (2024).

17     Yamada, Y. *et al.* The AI Scientist-v2: Workshop-Level Automated Scientific Discovery via Agentic Tree Search. *arXiv preprint arXiv:2504.08066* (2025).

18     Kitano, H. Nobel Turing Challenge: creating the engine for scientific discovery. *NPJ systems biology and applications* **7**, 29 (2021).

19     Van Noorden, R. & Perkel, J. M. AI and science: what 1,600 researchers think. *Nature* **621**, 672-675 (2023).

20     Hanson, B. *et al.* Garbage in, garbage out: mitigating risks and maximizing benefits of AI in research. *Nature* **623**, 28-31 (2023).

21     Lu, Y. *et al.* Unleashing the power of AI in science-key considerations for materials data preparation. *Scientific Data* **11**, 1039 (2024).

22      Musslick, S. *et al.* Automating the practice of science: Opportunities, challenges, and implications. *Proceedings of the National Academy of Sciences* **122**, e2401238121 (2025).

23      Harvard Longwood Medical Area Research Data Management Working Group. *Electronic Lab Notebook Comparison Matrix*, <https://doi.org/10.5281/zenodo.4723753> (2021).

24      Higgins, S. G., Nogiwa-Valdez, A. A. & Stevens, M. M. Considerations for implementing electronic laboratory notebooks in an academic research environment. *Nature Protocols* **17**, 179-189 (2022).

25      Benchling. *Benchling*, <https://www.benchling.com/> (2025).

26      Zheng, Y. *et al.* Homochiral Nanopropeller via Chiral Active Surface Growth. *J. Am. Chem. Soc.* **146**, 410-418, doi:10.1021/jacs.3c09652 (2024).

27      Wikipedia. *Wikipedia*, <https://www.wikipedia.org> (2025).

28      Wikipedia contributors. *Encyclopædia Britannica*, <https://en.wikipedia.org/wiki/Encyclop%C3%A6dia_Britannica> (2025).

29      Achiam, J. *et al.* Gpt-4 technical report. *arXiv preprint arXiv:2303.08774* (2023).

30      Jaech, A. *et al.* Openai o1 system card. *arXiv preprint arXiv:2412.16720* (2024).

31      Guo, D. *et al.* Deepseek-r1: Incentivizing reasoning capability in llms via reinforcement learning. *arXiv preprint arXiv:2501.12948* (2025).

32      Yang, A. *et al.* Qwen2. 5 technical report. *arXiv preprint arXiv:2412.15115* (2024).

33      Team, G. *et al.* Gemma 3 technical report. *arXiv preprint arXiv:2503.19786* (2025).

34      Grattafiori, A. *et al.* The llama 3 herd of models. *arXiv preprint arXiv:2407.21783* (2024).

35      xAi. *Grok 3 Beta — The Age of Reasoning Agents*, <https://x.ai/news/grok-3> (2025).

36      Chiou, K., Byun, S., Kim, J. & Huang, J. Additive-free carbon nanotube dispersions, pastes, gels, and doughs in cresols. *Proceedings of the National Academy of Sciences* **115**, 5703-5708 (2018).

37      Chiou, K. & Huang, J. Cresol-carbon nanotube charge-transfer complex: stability in common solvents and implications for solution processing. *Matter* **3**, 302-319 (2020).

38      McNutt, M.   Vol. 346   679-679 (American Association for the Advancement of Science, 2014).

39      Baker, M. 1,500 scientists lift the lid on reproducibility. *Nature* **533**, 452-454, doi:10.1038/533452a (2016).

40      Nature Publishing Group. *Nature Protocols*, <https://www.nature.com/nprot/> (2025).

41      protocols.io. *protocols.io*, <https://www.protocols.io> (2025).

42      Fantner, G. E. & Oates, A. C. Instruments of change for academic tool development. *Nature Physics* **17**, 421-424 (2021).

43      Zhu, M., Weng, Y., Yang, L. & Zhang, Y. in *ACL 2025*.

44      Zhang, X., Li, M., Xu, Y., Ren, J. & Zeng, A.-P. Quantitative study of H protein lipoylation of the glycine cleavage system and a strategy to increase its activity by co-expression of LplA. *Journal of biological engineering* **13**, 1-9 (2019).

45      Chen, L. & Zeng, A.-P. Rational design and metabolic analysis of Escherichia coli for effective production of L-tryptophan at high concentration. *Applied microbiology and biotechnology* **101**, 559-568 (2017).

46      Smith, A. An Inquiry into the Nature and Causes of the Wealth of Nations. *Readings in economic sociology*, 6-17 (2002).

47      Council, N. R. *Enhancing the Effectiveness of Team Science*.   (The National Academies

Press, 2015).

48    Simon, H. A. The Architecture of Complexity. *Proceedings of the American Philosophical Society* **106**, 467-482 (1962).

49    Baldwin, C. Y. & Clark, K. B. *Design rules, Volume 1: The power of modularity*.    (MIT press, 2000).

50    Wang, Y. *et al.* Autosurvey: Large language models can automatically write surveys. *Advances in Neural Information Processing Systems* **37**, 115119-115145 (2024).

51    Si, C., Yang, D. & Hashimoto, T. Can llms generate novel research ideas? a large-scale human study with 100+ nlp researchers. *arXiv preprint arXiv:2409.04109* (2024).

52    Gao, J. & Wang, D. Quantifying the use and potential benefits of artificial intelligence in scientific research. *Nature human behaviour* **8**, 2281-2292 (2024).

# Supplementary Information

## Contents

# S1 Supplementary Case Studies

## S1.1 Supplementary Case Study S1: Achieving systematic, AI-automated research via an Airalogy Protocol Workflow

### S1.1.1 Representing a research process with the Airalogy Protocol Workflow

In real-world research, realizing a specific research goal often requires a series of interconnected activities or experiments. To achieve that goal, researchers may need to establish a workflow that performs multiple iterative validation steps, forming a logical chain of experiments that ultimately yield conclusions. Moreover, such workflows frequently exhibit non-linear and dynamic feedback loops: specific experimental tasks may be repeated at various times, and the outcome of one step typically guides the design of the next, linking otherwise independent experiments into a coherent research pipeline. To systematically capture these processes, we need a method for representing such workflows.

Taking the ultrasonic dispersion of carbon nanotubes (CNTs) as an example. CNTs possess excellent electrical, thermal, and mechanical properties, which endow them with broad application prospects[53]. Currently, CNT powders can be produced in bulk quantities on the ton scale. However, their high surface energy and the strong $\pi$–$\pi$ interactions caused by their conjugated structures make them prone to aggregation. Industrially produced CNTs are often bundled or entangled, which significantly compromises their processability, limiting their practical applications[54,55]. To explore CNT dispersion, the procedure can be divided into four key steps, each represented by an Airalogy Protocol:

- Airalogy Protocol 1 ($AP_1$): Preparing a CNT dispersion from CNT powder
- $AP_2$: Dispersing the CNTs using tip sonication
- $AP_3$: Diluting a high-concentration CNT dispersion to obtain a low-concentration dispersion
- $AP_4$: Characterizing the dispersion results via scanning electron microscopy (SEM), wherein the mean CNT diameter is statistically derived from SEM images

In the general ideal case, we would sequentially execute these Protocols to disperse CNTs. For example, a typical and simplest process might involve preparing a dispersion from CNT powder ($AP_1$), followed by ultrasonic dispersion of the liquid ($AP_2$), and characterization ($AP_4$). This process can be represented as an Airalogy Protocol Path (APP):

$$APP\ 1: AP_1 \rightarrow AP_2 \rightarrow AP_4 \rightarrow End$$

The above APP is linear and can serve as an abstract representation of this research process. However, in real-world research, such linear processes are typically not exhaustive. For example, in the workflow of CNT dispersion, $AP_4$ (i.e., the characterization protocol) is not just a procedural step, but the key quality control for evaluating the dispersion results and guiding the next steps. Furthermore, $AP_4$ also serves as a criterion for determining whether to repeat preceding steps or to conclude the entire experimental process. To be more detailed, based on the characterization results from $AP_4$, one might need to repeat the ultrasonic dispersion and SEM characterization (i.e., repeat $AP_2 \rightarrow AP_4$), or dilute the dispersion before repeating the ultrasonic dispersion and SEM characterization (i.e., repeat $AP_3 \rightarrow AP_2 \rightarrow AP_4$). This process could continue iteratively until expected results are achieved. As a result, the APP can involve multiple applications of the same AP, meaning the possible APPs are infinite. For example, the following are also valid APPs:

- APP 2: $AP_1 \rightarrow AP_2 \rightarrow AP_4 \rightarrow AP_2 \rightarrow AP_4 \rightarrow$ End
- APP 3: $AP_1 \rightarrow AP_2 \rightarrow AP_4 \rightarrow AP_2 \rightarrow AP_4 \rightarrow AP_2 \rightarrow AP_4 \rightarrow$ End
- APP 4: $AP_1 \rightarrow AP_2 \rightarrow AP_4 \rightarrow AP_3 \rightarrow AP_2 \rightarrow AP_4 \rightarrow$ End
- APP 5: $AP_1 \rightarrow AP_2 \rightarrow AP_4 \rightarrow AP_3 \rightarrow AP_2 \rightarrow AP_4 \rightarrow AP_3 \rightarrow AP_2 \rightarrow AP_4 \rightarrow$ End
- APP 6: $AP_1 \rightarrow AP_2 \rightarrow AP_4 \rightarrow AP_2 \rightarrow AP_4 \rightarrow AP_3 \rightarrow AP_2 \rightarrow AP_4 \rightarrow$ End
- APP 7: $AP_1 \rightarrow AP_2 \rightarrow AP_4 \rightarrow AP_3 \rightarrow AP_2 \rightarrow AP_4 \rightarrow AP_2 \rightarrow AP_4 \rightarrow$ End

Clearly, the number of such APPs is infinite, and it would be impractical to list all linear paths to represent a research process. To address this, we introduce the concept of the Airalogy Protocol Workflow (APW), where users can define a research process as a graph, referred to as the Airalogy Protocol Workflow Graph (APWG). The APWG uses directed edges to represent the logical relationships between Protocols, allowing it to handle logical relationships with cyclic topologies. An APWG can be considered as a "path set" consisting of all valid APPs for the given research workflow. For example, the aforementioned study on carbon nanotube dispersion can be represented as **Supplementary Fig. S1** (also see **Fig. 5b**), with APP 1–7 conforming to the APWG's topology.

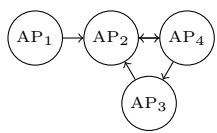

**Supplementary Fig. S1: Airalogy Protocol Workflow Graph for a carbon nanotube dispersion study (also see Fig. 5b).**
Each AP represents an Airalogy Protocol in this Workflow: $AP_1$: Preparing the dispersion liquid from CNT powder; $AP_2$: Ultrasonic dispersion of the CNT solution; $AP_3$: Preparation of a low-concentration CNT dispersion from a high-concentration solution; $AP_4$: Characterization of CNT dispersion. The directed edges (known as Airalogy Protocol Edges) indicate the logical sequence between the Protocols; for example, $AP_1 \rightarrow AP_2$ means that $AP_2$ can be performed as the next step after completing $AP_1$.

In some cases, the APWG alone may not fully describe the logic of an APW. In such cases, we can add additional descriptions to the Graph to further clarify the logic. This description is referred to as the Airalogy Protocol Workflow Graph Logic (APWGL). For example, we can add the following APWGL to **Supplementary Fig. S1** (**Supplementary Code S1**):

```
1. The entire dispersion process must occur within a solution system, and the preparation
of the dispersion from solid powder can only be the first step of the experiment: Protocol
1 must be the starting point of the Workflow.
2. Every dispersion system must go through the stages of preparation, ultrasonication, and
characterization: a Protocol Path must include at least one instance of (Protocol 1 -> Protocol
2 -> Protocol 4), and this sequence is irreversible.
3. Based on the characterization results, it is determined whether: 1. The sample needs to
be re-sonicated (Protocol 4 -> Protocol 2), or 2. The dispersion solution needs to be further
diluted before sonication (Protocol 4 -> Protocol 3 -> Protocol 2). After repeating either
of these two paths, characterization (Protocol 4) must be performed again to confirm the
subsequent results. These two paths can be iteratively followed based on the outcomes of
Protocol 4.
4. The characterization process (Protocol 4), as the only quality control step in the
experiment, can appear in the middle of the steps but must always be the final step in a Protocol
Path.
5. When the characterization results (Protocol 4) meet the research goal, the Workflow can
be terminated.
```

**Supplementary Code S1: Airalogy Protocol Workflow Graph Logic for a carbon nanotube dispersion study shown in Supplementary Fig. S1.**

### S1.1.2 Systematic acquisition of research data Records via Workflow application

Using the approach described above, we can effectively represent a workflow for a specific research objective via an Airalogy Protocol Workflow. Moreover, by encapsulating this Workflow into a dedicated Airalogy Protocol (**Supplementary Method S2.2.6**), one can share or apply it in much the same way as any other Protocol on the Platform. As a result, each use of this Workflow automatically generates a list of structured, standardized Airalogy Records (**Supplementary Fig. S2**), systematically advancing the goal of full digitization for a particular research workflow.

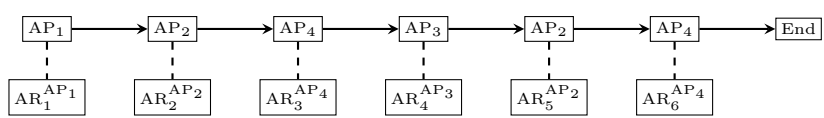

**Supplementary Fig. S2: An illustration of an Airalogy Protocol–Record Path.**
This figure demonstrates how, through applying an Airalogy Protocol Workflow, a corresponding list of Airalogy Records is generated. Taking the Workflow shown in **Supplementary Fig. S1** as an example, suppose that in one instance of using the Workflow, a series of Airalogy Protocols are executed in the order: $AP_1 \rightarrow AP_2 \rightarrow AP_4 \rightarrow AP_3 \rightarrow AP_2 \rightarrow AP_4 \rightarrow$ End. In practice, each application of an Airalogy Protocol in this sequence produces a corresponding Airalogy Record. The notation $AR_i^{AP_x}$ indicates that this is the $i$ th Airalogy Record in the Path, generated from the $x$ th

Protocol in the Workflow. Essentially, one application of an Airalogy Protocol Workflow results in an Airalogy Protocol–Record Path. For instance, based on the above Airalogy Protocol Path, the corresponding Airalogy Protocol–Record Path is: $AP_1 \rightarrow AR_1^{AP_1} \rightarrow AP_2 \rightarrow AR_2^{AP_2} \rightarrow AP_4 \rightarrow AR_3^{AP_4} \rightarrow AP_3 \rightarrow AR_4^{AP_3} \rightarrow AP_2 \rightarrow AR_5^{AP_2} \rightarrow AP_4 \rightarrow AR_6^{AP_4} \rightarrow End$. The list of Airalogy Records generated along the way (i.e., $[AR_1^{AP_1}, AR_2^{AP_2}, AR_3^{AP_4}, AR_4^{AP_3}, AR_5^{AP_2}, AR_6^{AP_4}]$) constitutes the structured, standardized research data obtained from this particular Workflow application.

### S1.1.3 A universal AI-driven research automation method that enables automated research using any Airalogy Protocol Workflow formed by arbitrary combinations of Airalogy Protocols

Achieving fully AI-automated research has the potential to dramatically accelerate innovation[6,7,14,16,17,56,57] and spark a paradigm shift in how science is conducted[13]. Although the Airalogy Protocol Workflow already allows us to systematically structure research activities and digitize experimental data for a given research goal, its possibilities extend far beyond these capabilities. Here, we developed a method called "AI Research Automation" (AIRA), which provides a universal approach for realizing full AI automation (encompassing automated research goal generation, research strategy design, next-step Protocol selection, parameter design, data analysis, and final research conclusion generation; **Fig. 5d**) for any Airalogy Protocol Workflow composed of arbitrarily combined Airalogy Protocols.

The theoretical underpinnings of the AIRA method are detailed below (readers who find the following abstract representations difficult to understand may skip ahead to **Section S1.1.4**, where we present a real-world case study on CNT dispersion research that applies AIRA in practice):

Assume we are given an arbitrary Airalogy Protocol Workflow (APW), which can be represented as:

$$APW = (APWG, APWGL) \tag{S1}$$
$$APWG = (APs, APEs) \tag{S2}$$
$$APs = AP_1, AP_2, \cdots, AP_m \tag{S3}$$

This indicates that an APW is defined by the Airalogy Protocol Workflow Graph (APWG) and the Airalogy Protocol Workflow Graph Logic (APWGL). In this workflow, there are *m* Airalogy Protocols (APs); APEs denotes the set of Airalogy Protocol Edges.

Given a specific APW for AI research automation, our goal is to derive a Research Conclusion (RC) from a given Research Goal (RG) via an automatable process. This relationship can be expressed as:

$$RG \rightarrow An\ automatable\ process? \rightarrow RC \tag{S4}$$

The AIRA method is designed to specify precisely what this "automatable process" entails and how it can be implemented.

In practice, applying an APW for a given RG involves sequentially executing the Protocols within the APW to form an Airalogy Protocol–Record Path (APRP) (**Supplementary Fig. S2**). This can be represented as follows:

$$\begin{aligned} \text{APRP} = \text{AP}_{x_1} &\to \text{AR}_1^{\text{AP}_{x_1}} \to \\ &\cdots \\ \text{AP}_{x_i} &\to \text{AR}_i^{\text{AP}_{x_i}} \to \\ &\cdots \\ \text{AP}_{x_n} &\to \text{AR}_n^{\text{AP}_{x_n}} \to \\ &\text{End} \end{aligned} \tag{S5}$$

$$\text{AP}_{x_i} \in \text{APs} \tag{S6}$$

$$\text{ARs}_{1 \to n} = [\text{AR}_1^{\text{AP}_{x_1}}, \cdots, \text{AR}_i^{\text{AP}_{x_i}}, \cdots, \text{AR}_n^{\text{AP}_{x_n}}] \tag{S7}$$

Here, $n$ represents the total number of Protocols traversed along the Path. $\text{AP}_{x_i}$ denotes the $i$ th Protocol in the APRP (which may be any Protocols from the set APs); note that $i$ serves merely as a positional index rather than a specific Protocol identifier. $\text{AR}_i^{\text{AP}_{x_i}}$ represents the Airalogy Record produced by the $i$ th Protocol in the APRP. In this manner, we obtain a sequence of Airalogy Records $\text{ARs}_{1 \to n}$, capturing the Records of each step along the APRP.

Then, by comprehensively analyzing both the APW and the sequence $\text{ARs}_{1 \to n}$—which, given that the order of elements in the list mirrors that of the APRP (and for simplicity, can be assumed to encapsulate the APRP information)—we can derive the RC corresponding to a given RG. This derivation can be performed automatically using AI analysis, which can be expressed as follows:

$$\text{RC} = \text{AI}_1(\text{APW}, \text{RG}, \text{ARs}_{1 \to n}) \tag{S8}$$

Thus, the key to implementing **Supplementary Eq. S4** reduces to automatically generating a reasonable APRP for a given RG.

To address this challenge, there are two critical automation problems:

- **Automation Problem 1:** Under a given APW, the number of possible Airalogy Protocol–Record Paths is infinite, and the number of Protocols traversed in any single APRP can be arbitrary. Therefore, a crucial challenge is to automatically identify a reasonable APRP. In other words, to dynamically generate an APRP, we need an automated method capable of selecting the next

Airalogy Protocol to execute.

- **Automation Problem 2:** Even if we can automatically select the next Airalogy Protocol, another key challenge is to automatically apply that Protocol and obtain its corresponding Airalogy Record.

These challenges form the core of our AIRA method for transforming a RG into a RC through an automatable process.

To address the aforementioned issues—while temporarily setting aside Automation Problem 2—we can design **Supplementary Eq. S4** as follows to solve Automation Problem 1:

$$
\begin{aligned}
&RG \rightarrow RS \\
&\qquad \rightarrow AP_{x_1} \rightarrow AR_1^{AP_{x_1}} \rightarrow PRC_1 \\
&\qquad \cdots \\
&\qquad \rightarrow AP_{x_i} \rightarrow AR_i^{AP_{x_i}} \rightarrow PRC_i \\
&\qquad \cdots \\
&\qquad \rightarrow AP_{x_n} \rightarrow AR_n^{AP_{x_n}} \rightarrow PRC_n \\
&\rightarrow End \rightarrow RC
\end{aligned}
\tag{S9}
$$

In this process, when given a RG, the AI analyzes the RP together with the information provided by the APW and designs a corresponding Research Strategy (RS). The primary objective of this step is for the AI to comprehensively assess both the APW and the RG by performing two key actions: (1) determine whether a reasonable application of the APW can adequately address the RG; and (2) if not, immediately inform the user that the given RG is unsuitable for resolution via the APW, or if it is, propose a specific RS (i.e., an overall experimental design) to achieve the RG. This RS, in turn, provides high-level guidance for the subsequent automated selection of the appropriate Airalogy Protocol.

The critical role of the RS in automating the APW becomes evident from two simple observations. First, a given APW is applicable only to certain types of RGs; it cannot address every conceivable RG. For example, the APW illustrated in **Supplementary Fig. S1** for CNT dispersion is clearly unsuitable for an RG such as "investigating how cells undergo mitosis". Without first assessing the suitability of the RG, blindly generating an APRP would be futile and would waste valuable research resources. Second, different RGs inherently require different strategies. For instance, for the CNT dispersion APW shown in **Supplementary Fig. S1**, one RG might be "to disperse carbon nanotubes to an average diameter of 10–30 nm using m-cresol with minimal ultrasonic treatment", whereas another might be "to achieve the same dispersion with minimal dilution". Each RG demands a distinct RS.

The generation of the RS can be achieved using the following AI automation method:

$$
RS \lor End = AI_2(APW, RG)
\tag{S10}
$$

This equation indicates that AI Method 2 ($AI_2$) automatically evaluates whether a given APW can address the RG. If it can, the AI produces a corresponding RS; if not, the process terminates (End).

Once an RS is obtained, as illustrated in **Supplementary Eq. S9**, we can automatically select the first Airalogy Protocol using:

$$AP_{x_1} = AI_3(\text{APW, RG, RS}) \tag{S11}$$

By applying $AP_{x_1}$, we obtain its corresponding Airalogy Record ($AR_1^{AP_{x_1}}$). This Record is then analyzed to derive an Phased Research Conclusion (PRC). The PRC may convey rich insights beyond a simple summary—for example, it can indicate: (1) what preliminary conclusion the current Protocol–Record Path provides for the RG; (2) whether that conclusion is sufficient to address the RG; (3) if not, what further steps should be planned for subsequent research; (4) whether the RS is reasonable, and if not, how it might be adjusted given the inherent uncertainties in research; and (5) whether any notable anomalies or exceptional observations have been discovered. This process can be achieved using the following AI automation method:

$$PRC_1 = AI_4(\text{APW, RG, RS}, [AR_1^{AP_{x_1}}]) \tag{S12}$$

Clearly, for subsequent steps, the above process can be repeated:

$$
\begin{aligned}
AP_{x_2} \vee \text{End} &= AI_3(\text{APW, RG, RS}, [PRC_1], [AR_1^{AP_{x_1}}]) \\
PRC_2 &= AI_4(\text{APW, RG, RS}, [AR_1^{AP_{x_1}}, AR_2^{AP_{x_2}}], [PRC_1]) \\
AP_{x_3} \vee \text{End} &= AI_3(\text{APW, RG, RS}, [PRC_1, PRC_2], [AR_1^{AP_{x_1}}, AR_2^{AP_{x_2}}]) \\
PRC_3 &= AI_4(\text{APW, RG, RS}, [AR_1^{AP_{x_1}}, AR_2^{AP_{x_2}}, AR_3^{AP_{x_3}}], [PRC_1, PRC_2]) \\
&\cdots
\end{aligned}
\tag{S13}
$$

Thus, the Airalogy Protocols along the APRP are sequentially selected until reaching "End". This process can be generally expressed as:

$$PRCs_{1 \to i-1} = [PRC_1, \cdots, PRC_{i-1}] \tag{S14}$$

$$AP_{x_i} \vee \text{End} = AI_3(\text{APW, RG, RS}, PRCs_{1 \to i-1}, ARs_{1 \to i-1}) \tag{S15}$$

$$PRC_i = AI_4(\text{APW, RG, RS}, ARs_{1 \to i}, PRCs_{1 \to i-1}) \tag{S16}$$

In the above equations, when $i = 1$, we define $PRCs_{1 \to 0} = []$, and $ARs_{1 \to 0} = []$; thus, **Supplementary Eq. S15** takes the same form as **Supplementary Eq. S11**, and **Supplementary Eq. S16** is identical to **Supplementary Eq. S12**.

Furthermore, since the process produces a sequence of Phased Research Conclusions $PRCs_{1 \to n}$, these Conclusions can assist in generating the final Research Conclusion. Therefore, we can extend

**Supplementary Eq. S8** as follows:

$$RC = AI_1(APW, RG, ARs_{1 \to n}, PRCs_{1 \to n}) \tag{S17}$$

Thus, by following the process outlined above, we can effectively resolve Automation Problem 1.

However, Automation Problem 2 remains unresolved: choosing the next Airalogy Protocol alone is insufficient to automatically apply that Protocol and obtain its corresponding Airalogy Record. This is because each Airalogy Protocol defines an experimental plan with adjustable parameters, referred to as Airalogy Fields. For instance, an ultrasonic-dispersion AP might include a Field for "ultrasonic duration". Without specifying its value in advance, the Protocol cannot be properly executed, nor can its Record be generated.

To analyze how to obtain each Record, note that every Airalogy Record fundamentally consists of a set of Airalogy Fields. Formally,

$$AR_i^{AP_{x_i}} = AFs_i^{AP_{x_i}} \tag{S18}$$

$$AFs_i^{AP_{x_i}} = \{AF_{i,1}^{AP_{x_i}}, AF_{i,2}^{AP_{x_i}}, \cdots, AF_{i,k_{x_i}}^{AP_{x_i}}\} \tag{S19}$$

Here, $AR_i^{AP_{x_i}}$ denotes the $i$ th Airalogy Record generated on the current APRP from the $x_i$ th Airalogy Protocol ($AP_{x_i}$) within a given APW. $AFs_i^{AP_{x_i}}$ represents the set of all Airalogy Fields in that Record, comprising $k_{x_i}$ Fields that capture all key data items recorded during the research activity.

Furthermore, these Airalogy Fields can be classified into two categories. The first category, the Parameter-type Fields, guides how an Airalogy is executed and must be specified before the experiment begins. The second category, the Feedback-type Fields, consists of values obtained through executing the Protocol that contains the Parameter Fields, effectively serving as feedback from the Protocol application. This process can be expressed as:

$$AFs_i^{AP_{x_i}} = AFs_i^{AP_{x_i},P} \cup AFs_i^{AP_{x_i},F} \tag{S20}$$

$$AFs_i^{AP_{x_i},P} = \{AF_{i,1}^{AP_{x_i},P}, AF_{i,2}^{AP_{x_i},P}, \cdots, AF_{i,k_{x_i,P}}^{AP_{x_i},P}\} \tag{S21}$$

$$AFs_i^{AP_{x_i},F} = \{AF_{i,1}^{AP_{x_i},F}, AF_{i,2}^{AP_{x_i},F}, \cdots, AF_{i,k_{x_i,F}}^{AP_{x_i},F}\} \tag{S22}$$

$$k_{x_i} = k_{x_i,P} + k_{x_i,F} \tag{S23}$$

$$AFs_i^{AP_{x_i},F} = AP_{x_i}(AFs_i^{AP_{x_i},P}) \tag{S24}$$

Here, $\mathrm{AFs}_i^{\mathrm{AP}_{x_i},\mathrm{P}}$ and $\mathrm{AFs}_i^{\mathrm{AP}_{x_i},\mathrm{F}}$ denote the sets of Parameter-type and Feedback-type Airalogy Fields within $\mathrm{AFs}_i^{\mathrm{AP}_{x_i}}$, respectively. In total, there are $k_{x_i,\mathrm{P}}$ Parameter Fields and $k_{x_i,\mathrm{F}}$ Feedback Fields.

Hence, once we know the Parameter Fields $\mathrm{AFs}_i^{\mathrm{AP}_{x_i},\mathrm{P}}$ for $\mathrm{AP}_{x_i}$, the remaining Feedback Fields $\mathrm{AFs}_i^{\mathrm{AP}_{x_i},\mathrm{F}}$ can be derived by applying $\mathrm{AP}_{x_i}$. Thus,

$$\mathrm{AR}_i^{\mathrm{AP}_{x_i}} = \mathrm{AFs}_i^{\mathrm{AP}_{x_i},\mathrm{P}} \cup \mathrm{AFs}_i^{\mathrm{AP}_{x_i},\mathrm{F}} \tag{S25}$$

$$= \mathrm{AFs}_i^{\mathrm{AP}_{x_i},\mathrm{P}} \cup \mathrm{AP}_{x_i}(\mathrm{AFs}_i^{\mathrm{AP}_{x_i},\mathrm{P}}) \tag{S26}$$

These equations reveal the core requirement for Automation Problem 2: to generate an Airalogy Record automatically, we must determine the Parameter Fields of the chosen Protocol prior to execution. Once these Parameter Fields are known, the Protocol can run automatically to produce its Feedback Fields and thus the complete Record.

For this purpose, given a specific $\mathrm{AP}_{x_i}$, we can derive its corresponding $\mathrm{AFs}_i^{\mathrm{AP}_{x_i},\mathrm{P}}$ using an AI-based method:

$$\mathrm{AFs}_i^{\mathrm{AP}_{x_i},\mathrm{P}} = \mathrm{AI}_5(\mathrm{APW}, \mathrm{RG}, \mathrm{RS}, \mathrm{AP}_{x_i}, \mathrm{PRCs}_{1 \to i-1}, \mathrm{ARs}_{1 \to i-1}) \tag{S27}$$

Based on this, **Supplementary Eq. S9** can be extended to the following form:

$$\begin{aligned}
\mathrm{RG} &\to \mathrm{RS} \\
&\to \mathrm{AP}_{x_1} \to \mathrm{AFs}_1^{\mathrm{AP}_{x_1},\mathrm{P}} \to \mathrm{AFs}_1^{\mathrm{AP}_{x_1},\mathrm{F}} \to \mathrm{AR}_1^{\mathrm{AP}_{x_1}} \to \mathrm{PRC}_1 \\
&\cdots \\
&\to \mathrm{AP}_{x_i} \to \mathrm{AFs}_i^{\mathrm{AP}_{x_i},\mathrm{P}} \to \mathrm{AFs}_i^{\mathrm{AP}_{x_i},\mathrm{F}} \to \mathrm{AR}_i^{\mathrm{AP}_{x_i}} \to \mathrm{PRC}_i \\
&\cdots \\
&\to \mathrm{AP}_{x_n} \to \mathrm{AFs}_n^{\mathrm{AP}_{x_n},\mathrm{P}} \to \mathrm{AFs}_n^{\mathrm{AP}_{x_n},\mathrm{F}} \to \mathrm{AR}_n^{\mathrm{AP}_{x_n}} \to \mathrm{PRC}_n \\
&\to \mathrm{End} \to \mathrm{RC}
\end{aligned} \tag{S28}$$

Thus, we obtain the complete chained process of the AIRA method, where any node in the chain can be derived using the equations presented above:

$$RS \vee End = AI_2(APW, RG) \tag{S10}$$

$$AP_{x_i} \vee End = AI_3(APW, RG, RS, PRCs_{1 \to i-1}, ARs_{1 \to i-1}) \tag{S15}$$

$$AFs_i^{AP_{x_i},P} = AI_5(APW, RG, RS, AP_{x_i}, PRCs_{1 \to i-1}, ARs_{1 \to i-1}) \tag{S27}$$

$$AFs_i^{AP_{x_i},F} = AP_{x_i}(AFs_i^{AP_{x_i},P}) \tag{S24}$$

$$AR_i^{AP_{x_i}} = AFs_i^{AP_{x_i},P} \cup AFs_i^{AP_{x_i},F} \tag{S25}$$

$$PRC_i = AI_4(APW, RG, RS, ARs_{1 \to i}, PRCs_{1 \to i-1}) \tag{S16}$$

$$RC = AI_1(APW, RG, ARs_{1 \to n}, PRCs_{1 \to n}) \tag{S17}$$

The process represented by the above equations can be implemented through various automation methods; therefore, the chained process described in **Supplementary Eq. S28** can, in theory, be completely automated.

- **Supplementary Eqs. S10, S15, S27, S16, S17:** Automated through AI; this study has demonstrated specific implementation methods (**Supplementary Method S2.7**).
- **Supplementary Eq. S24:** Implemented using automated scientific equipment, robotics, and similar methods.
- **Supplementary Eq. S25:** Achieved via simple set arithmetic operations.

In summary, the AIRA method enables AI-driven research automation in a universally applicable manner for any given Airalogy Protocol Workflow and Research Goal.

Additionally, this method can be extended to scenarios where no specific Research Goal is provided, because using AI, we can automatically generate a potential RG based solely on the APW:

$$RG = AI_6(APW) \tag{S29}$$

This implies that AI could potentially propose its own scientific questions or goals and conduct research in an automated fashion.

### S1.1.4 A practical example of CNT dispersion research using the AIRA method

The complete research process is illustrated in **Supplementary Fig. S3** (also see **Fig. 5e**). In this study, we set the following research goal:

Disperse CNTs in m-cresol to achieve an average nanotube diameter of 10–30 nm, with as few dilution steps as possible.

Leveraging the Airalogy Platform's AI research co-pilot, Aira (which implements the AIRA method), designed a research strategy based on this goal:

1. Define the starting point and basic conditions

- Protocol 1 (preparing a high-concentration CNT dispersion from powder) must be performed first.
- m-cresol is used as the dispersing medium.

2. Establish the basic experimental path
   - Execute Protocol 1 → Protocol 2 → Protocol 4 in sequence: prepare a high-concentration dispersion (Protocol 1), ultrasonically disperse it (Protocol 2), then characterize the dispersion (Protocol 4) to measure the average CNT diameter.
   - If this first characterization step already meets the 10–30 nm diameter target, the workflow terminates.

3. Decision logic based on characterization results
   - If the measured diameter remains too large, two approaches are possible:
     A. Perform another round of ultrasonic dispersion (Protocol 2) followed by another measurement (Protocol 4).
     B. Dilute the dispersion (Protocol 3) before ultrasonic dispersion (Protocol 2), then measure again (Protocol 4), aiming to reach the target diameter by lowering the CNT concentration.
   - Approach A is prioritized to minimize the number of dilutions. If repeated ultrasonic dispersion fails to reduce the diameter further, Approach B is selected.

Based on this goal and strategy, Aira chose Protocol 1 as the starting step, specifying an initial concentration of 1 mg/ml. The human researchers prepared a 1 mg/ml CNT dispersion by mixing 10 mg of single-walled CNT powder with 10 ml of m-cresol. Next, Aira chose Protocol 2 with a 30-minute ultrasonic treatment. The researchers used a Qsonica Q125 sonicator (90% power amplitude, 2 s on/2 s off pulse mode) for 30 minutes. Aira then selected Protocol 4 to characterize the dispersion, where the researchers deposited 10 µl of the dispersion onto a silicon wafer for SEM imaging, obtaining an average diameter of approximately 150 nm. Seeing that 150 nm was well above the 10–30 nm target, Aira automatically chose Protocol 2 again for further ultrasonic dispersion. In this manner, Aira continuously selected the subsequent Protocol and designed the necessary experimental parameters, while the human researchers carried out each step. Importantly, Aira adjusts decisions in real time according to results. At step 7, Aira noted that the new diameter (~80 nm) was unchanged from step 5, despite further ultrasonic dispersion (Protocol 2). Recognizing that further sonication at this concentration likely would not improve dispersion, Aira added a dilution step by selecting Protocol 3 (step 8) before Protocol 2 (step 9). Finally, by step 12, the average CNT diameter reached around 25 nm, meeting the specified target range, and Aira terminated the Workflow (End), providing the final research conclusion:

In this research workflow, we leveraged the iterative strategy offered by Protocols 1, 2, 3, and 4 for CNT dispersion, ultimately achieving the user-defined goal of maintaining an average CNT diameter of 10–30 nm in an m-cresol–based dispersion. Specifically:

1. Goal Fulfillment:
   The final characterization (Protocol 4) showed that after five rounds of ultrasonic dispersion (Protocol 2) and one dilution step (Protocol 3), the average nanotube diameter reached approximately 25 nm, within the user-defined target range of 10–30 nm. This outcome demonstrates that the proposed strategy can achieve the desired dispersion state with minimal dilution steps.

2. Notable Observations and Process Insights:

   I observed that continuing ultrasonic dispersion alone no longer reduced the average diameter after a certain point. However, once a dilution step (Protocol 3) was introduced, an additional round of ultrasonic dispersion (Protocol 2) further decreased the average diameter. This indicates that lowering the CNT concentration facilitates more effective interaction between m-cresol and the nanotube surfaces, thereby enhancing the dispersion effect.

3. Future Directions and Strategy Optimization:

   Further investigation into various initial CNT concentrations may help identify the optimal or maximum feasible concentration for achieving this goal.

In summary, by automating protocol selection and iterative strategy adjustments, we successfully met the targeted dispersion objective and gathered new data and insights for future refinement. These findings validate the workflow and offer valuable guidance for deeper explorations of m-cresol as a dispersing medium.

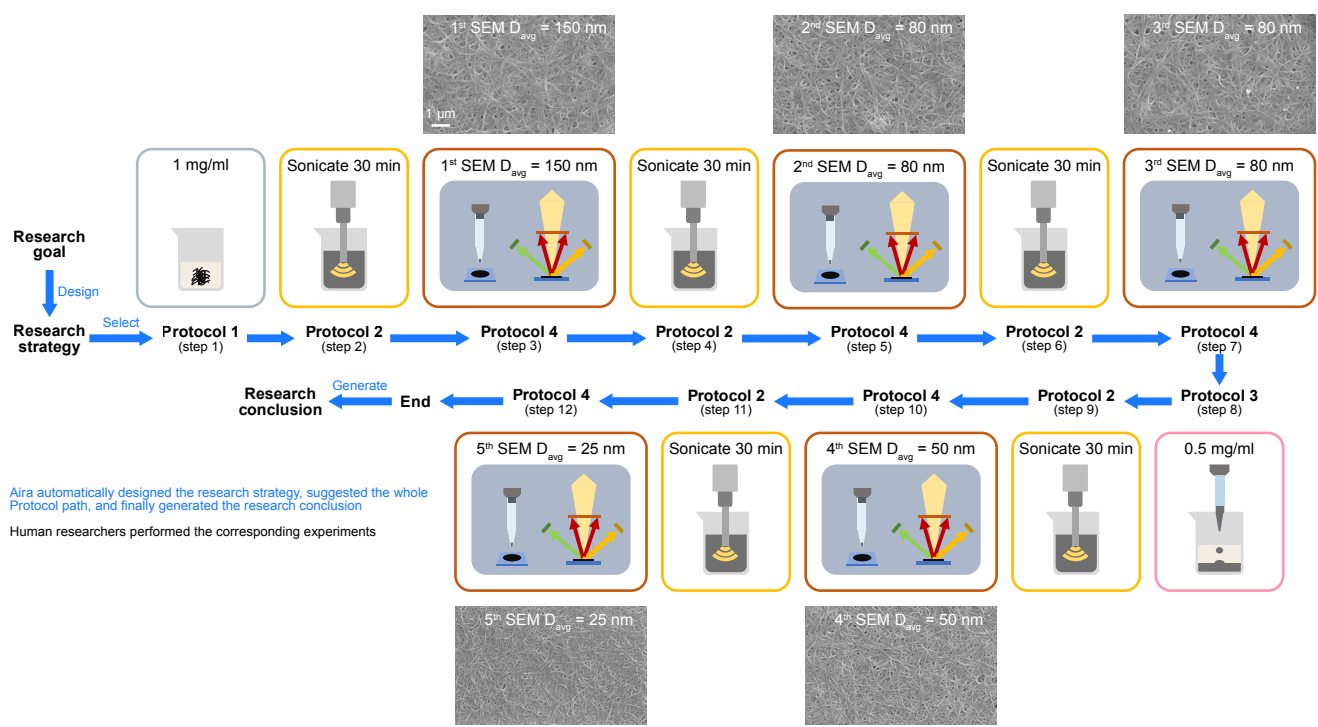

**Supplementary Fig. S3: A practical example of CNT dispersion research using the AIRA method.**

## S1.1.5 Significance of the Airalogy Protocol Workflow and the AIRA Method for real-world research

1. **Balancing modularity and systematic organization of experiments:** In the Airalogy Platform, each Airalogy Protocol functions as a relatively self-contained experimental step. These individual Protocols can be invoked independently in similar scenarios, lending them a modular

"building block" quality. However, real research often involves multiple interconnected steps. By assembling various Protocols into an Airalogy Protocol Workflow, one can sequence seemingly disjoint or unordered experiments according to the demands of a specific research goal. In essence, the individual Protocols act like individual building blocks that, when combined into a Workflow, yield a cohesive and systematic approach to a given research question.

2. **Aira as a universal adapter for flexible assembly:** Even with well-designed modules (i.e., individual Airalogy Protocols), assembling them into a functioning Workflow can be challenging. Each Protocol may be contributed by different researchers with distinct data structures or standards. This situation is analogous to trying to fit together building blocks with mismatched studs and connectors (**Fig. 5c**). However, with the rise of LLMs offering extensive cross-disciplinary knowledge, the Aira (which implements the AIRA method) leverages AI as a universal "adapter" bridging these heterogeneous modules (**Fig. 5c**). Through Aira, any Airalogy Protocol Workflow (constructed from arbitrarily chosen Protocols) can still achieve AI-driven automation.

3. **Streamlined data collection and utilization:** Once individual Protocols are logically assembled into a Workflow, the Workflow can be used to seamlessly collect structured, standardized electronic records for each experimental step. This design facilitates subsequent data utilization. Moreover, chaining Protocols together provides insight into trends and logical interdependencies across multiple steps. Since automated approaches often require not just static endpoints but evolving data streams, converting disjointed Protocol-level data ("point data") into a continuous Workflow-based dataset ("line data") enables dynamic experimental design and real-time adjustments (as demonstrated in **Supplementary Fig. S3**, where Aira autonomously determines whether to proceed with direct ultrasonication or to dilute first).

4. **Goal-oriented sharing and reuse:** An Airalogy Protocol Workflow can be shared or reused in the Airalogy Hub just like an individual Airalogy Protocol (**Supplementary Method S2.2.6**). Consequently, the Platform's community can exchange not only single-step Protocols describing isolated experiments but also entire Workflows designed to address specific research goals. This expands the Hub into a rich repository of systematic, goal-driven research assets that others can adopt, adapt, and enhance for their own scientific inquiries.

### S1.1.6 Perspectives

In the example described above, the Airalogy Protocol Workflow was assembled by manually selecting and arranging suitable Protocols. This effectively imparted a specific research goal or background context to the Workflow, thereby constraining and guiding the subsequent research. However, we also envision a more flexible, goal-oriented strategy that forgoes pre-defined Workflows in favor of on-the-fly research automation (**Supplementary Fig. S4**).

As the Airalogy community grows, we anticipate two important developments:

1. The Airalogy Hub will accumulate an ever-expanding library of Airalogy Protocols contributed by researchers from diverse fields;
2. AI capabilities for designing Airalogy Protocols from scratch will continue to advance (for

example, **Supplementary Method S2.6.1** demonstrates a method for enabling AI to generate an effective Airalogy Protocol based on a given design objective).

In this scenario, we can develop AI-driven methods to search the Hub for suitable Airalogy Protocols that match a particular research goal or question—whether that goal is proposed by human scientists or by AI itself. If no fully compatible Protocol is found, the AI could design a new one from scratch to meet the specified requirements. The AI would then automatically assemble these Protocols into an Airalogy Protocol Workflow and conduct the research using the AIRA method, generating conclusions for the specified goal. Moreover, the AI could dynamically reconfigure the Workflow based on real-time results. For instance, an initially designed Workflow might lack certain Protocols necessary to address unforeseen complexities; in response, the AI could search for and incorporate additional Protocols from the Hub, thereby refining the Workflow. The AI could also optimize individual Protocols within the Workflow (e.g., **Supplementary Case Study S5** illustrates how a Protocol was iteratively refined until it functioned effectively as a standard operating procedure). By enhancing individual Protocols in this manner, the overall Workflow improves. Ultimately, the AI continues the research cycle following the AIRA method until the conclusions derived from the current Workflow sufficiently address the given research goal or question.

Through this approach, we aim to realize a universally applicable, AI-driven method of full research automation—one capable of tackling virtually any research goal/question without relying on manually curated Workflows.

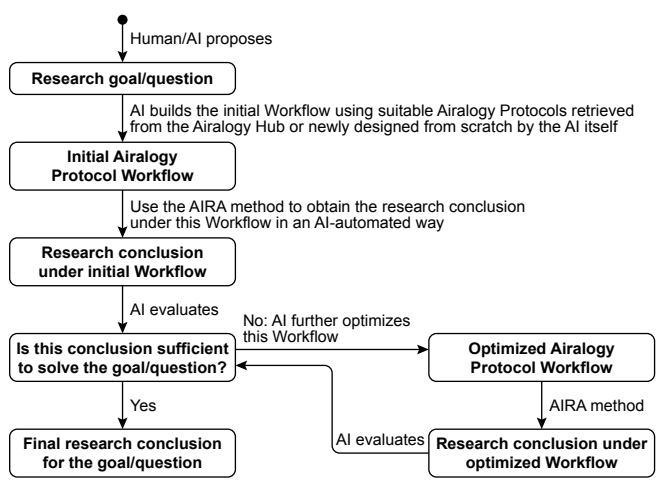

**Supplementary Fig. S4: A potential method for achieving fully AI-driven, automated research for any specific research goal or question based on the Airalogy Protocol Workflow and the AIRA method.**

## S1.2 Supplementary Case Study S2: Sharing cutting-edge scientific research via the Airalogy Platform

### S1.2.1 Challenges in sharing experimental protocols

In scientific research, the reproducibility of experimental results serves as a critical criterion for evaluating the reliability of research outcomes. Moreover, the ease of replicating published experimental data is a fundamental prerequisite for fostering scientific communication and progress. However, in practice, not all researchers can successfully reproduce results using the protocols provided in publications. A survey published in *Nature* revealed that in the field of biology alone, over 70% of researchers failed to reproduce other scientists' findings, while approximately 60% struggled to replicate their own work[39]. Even in chemistry—a discipline renowned for its high reproducibility due to standardized synthesis and analytical methods—nearly 90% and 63% of chemists reported difficulties in fully reproducing others' or their own experiments, respectively[39]. Concurrently, the volume of scientific publications continues to grow exponentially each year. This widespread irreproducibility inevitably leads to significant financial and temporal waste, diminishes scientific productivity, and erodes public trust in scientific research.

The causes of irreproducibility, beyond rare instances of intentional data fabrication, are predominantly attributed to the following factors[58-60]:

1. Insufficient description of crucial experimental details ("tricks") in methodology sections;
2. Limited access to raw research data;
3. Inadequate management and analysis of complex datasets;
4. Researchers reporting erroneous data they believe to be valid;
5. Cognitive biases stemming from researchers' preconceptions;
6. Selective reporting of positive results driven by publication pressures.

Acquiring comprehensive and authentic experimental details is essential for achieving better reproducibility. Currently, several platforms aim to address this challenge. Publications such as *Nature Protocols* and *Bio-protocol* focus on experimental methodology and technical standardization, offering detailed descriptions of newly developed experimental methods to enhance global research reproducibility and transparency. However, like other peer-reviewed journals, these publishers enforce rigorous editorial standards, often resulting in lengthy review cycles that delay the timely dissemination of cutting-edge protocols. Additionally, readers encounter limited opportunities to interact with authors to resolve technical queries arising during experiments.

In recent years, online protocol-sharing platforms (such as Protocol.io[41]) have emerged, enabling users to upload experimental protocols and details independently while facilitating question-and-answer exchanges. Analogous to preprint and formal publication platforms, these tools accelerate knowledge-sharing by improving timeliness and flexibility. Nevertheless, it is noted that such platforms primarily focus on instructing users how to perform experiments, offering only a unidirectional guidance function. As highlighted earlier, irreproducibility issues extend beyond insufficient methodological details to include inadequate dataset management, flawed analytical approaches, and cognitive biases.

However, Airalogy Protocols go far beyond merely supplying step-by-step instructions. In addition to offering comprehensive methodological details, each Protocol is intrinsically linked to its corresponding experimental Records, thereby enabling systematic dataset management. Protocols can also incorporate custom automated data analysis, substantially improving the reproducibility of downstream data processing and interpretation. Furthermore, each Protocol's built-in Discussions section allows users to consult with the Protocol's creator whenever questions arise, fostering constructive interaction between users and authors. Beyond these features, the Airalogy Platform's AI research copilot, Aira, provides 24/7 intelligent support, further enhancing the Protocol's interactivity. Together, these characteristics make Airalogy Protocols a powerful tool for sharing and disseminating research methods, while simultaneously improving reproducibility.

Below, we present a case study demonstrating how our latest method for synthesizing gold nanopropellers[26] was shared via an Airalogy Protocol. This example illustrates the advantages of using Airalogy Protocols to distribute cutting-edge research methods.

## S1.2.2 Sharing a cutting-edge gold nanopropeller synthesis method via Airalogy Protocol: advantages over traditional protocol formats

In nanoparticle synthesis, experimental outcomes are highly sensitive to slight variations in parameters; even small changes can significantly affect crystal morphology, size, and yield[61-63]. Ensuring reproducibility is therefore crucial in nanoscience. When disseminating our gold nanopropeller synthesis method[26] through an Airalogy Protocol, we adopted several strategies to improve experimental reproducibility:

### 1. Providing more comprehensive experimental details

In traditional publications, the experimental methods for cutting-edge research are typically shared through concise text descriptions in the "Methods" section. For instance, due to space limitations in our *Journal of the American Chemical Society* article on gold nanopropeller synthesis[26], we could only provide a very condensed overview (**Supplementary Fig. S5a**). Such brevity can make it challenging for newcomers to replicate the experiment. Additionally, purely text-based guidance often fails to convey the finer nuances of certain procedures, which may require images, videos, or even audio clips to illustrate effectively. By contrast, an Airalogy Protocol (**Supplementary Fig. S 5b**) is not constrained by page limits. Authors can therefore include exhaustive instructions and record critical details or "know-how" to guide users step-by-step. For elements that are difficult to capture in text alone—such as subtle manipulations or auditory cues—multimedia content (images, videos, audio) can be seamlessly integrated to provide more intuitive instruction.

**a**

**Synthesizing Au decahedra:** Gold decahedra were synthesized following literature. [1,] [2] In brief, gold seeds were obtained by reduction of HAuCl₄ with NaBH₄ in CTAC solution, in the presence of citric acid at 80 °C. Then 68 μL seed was added to a growth solution containing BDAC (100 mL, 100 mM), HAuCl₄ (1 mL, 50 mM) and AA (0.75 mL, 100 mM) at 30 °C. The mixture was left for 30 minutes, then washed with Milli-Q water twice and concentrated (×10 concentration, estimated $1.1 \times 10^{13}$ particles/L), before finally stored in 1 mM CTAB.

**Synthesizing Homochiral Nano-propeller:** CTAB (0.8 mL, 100 mM), HAuCl₄ (10 mM, 0.2 mL) and H₂O (3.95 mL) were placed into a 20 ml vial. Then 0.1 M, 0.475mL AA were added to the above solution quickly. After shaking the vial to mix the reactant, color of the mixture changes quickly from yellow to colorless which we note as growth solution. Then 40 μL GSH solution (1.5 mg GSH powder dispersed in 10 mL water) and 1 μL Au decahedra were added to the growth solution for the second time growth. The growth was maintained at 30 °C for 2 h.

**b**

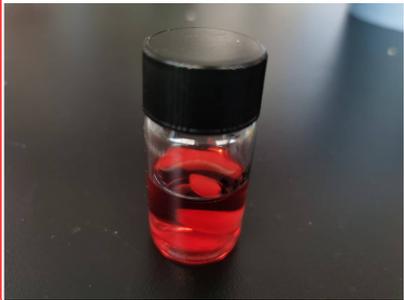

**Supplementary Fig. S5: Sharing detailed and intuitive methods for gold nanopropeller synthesis via Airalogy Protocol.**

**a,** A concise description of the experimental procedure from the original publication[26]. **b,** A screenshot of the method shared through an Airalogy Protocol. For example, section ① expands upon the brief instructions of seed synthesis from the original paper (yellow underlined) to offer more detailed guidance and highlight critical steps. In section ②, an image is embedded to provide a more intuitive, visual explanation of specific techniques and key operational points.

## 2. Beyond sharing experimental methods: also sharing how experimental data are recorded and managed

Under the "publish or perish" mentality and rising academic pressures of the last decade[64], researchers often report only successful, positive outcomes. Negative or inconclusive findings remain largely hidden, despite being equally important for improving experimental reproducibility. By intrinsically linking experimental Records to Protocols, the Airalogy Protocol not only facilitates the dissemination of experimental methods but also promotes transparent data-recording practices.

For instance, when sharing our latest Protocol on synthesizing gold nanopropellers, we went beyond simply providing the experimental procedure. We also emphasized potential pitfalls that could lead to unsuccessful synthesis. The Protocol includes strategically designed Airalogy Fields that prompt users to record real experimental parameters and conditions (**Supplementary Fig. S6 ①**). In addition, specific Fields capture the key experimental data generated during the synthesis (**Supplementary Fig. S6 ②**). Users can log both "positive" and "negative" results; after each experiment, the Airalogy Platform produces a unique Airalogy Record containing the protocol steps, parameters, and outcomes.

This approach not only ensures detailed documentation of the experimental process but also facilitates standardized data recording. Records can be retained for later analysis and shared with the community once proper permissions are granted. Each Record is assigned a unique ID for both traceability and broader dissemination. Encouraging the community to upload a more diverse range of Records—regardless of their success—helps improve the transparency and reliability of research data. Moreover, as Airalogy is a multidisciplinary platform, datasets from various fields can be hosted, enabling researchers to perform secondary analyses, combine data from multiple experiments, and spark cross-disciplinary collaborations.

**Supplementary Fig. S6: Recording key experimental parameters and outcomes via an Airalogy Protocol.**

①, Example Fields for documenting synthesis parameters directly within the Protocol. The `h2o_volume`, `haucl4_conc`, `ctab_conc`, `aa_conc`, and `gsh_volume` record five parameters that can critically affect the experiment: water volume, $HAuCl_4$ concentration, CTAB concentration, ascorbic-acid concentration, and GSH volume, respectively. ②, An example Field for illustrating how real experimental data can be entered and stored in the Protocol. Users may enter the actual outcome (whether positive or negative) under the specified parameters and conditions in the `au_nanopropeller_result` Field.

## 3. Providing automated, AI-driven data analysis and assessment

When sharing experimental methods, the ability to offer targeted, automated analysis of the resulting data is equally important—particularly for newcomers, who may not yet recognize what constitutes a "successful" outcome. In practice, researchers often replicate someone else's protocol, gather the results, and then wonder whether their data truly indicate success. By embedding customized data-analysis and evaluation functionalities in an Airalogy Protocol, these uncertainties can be significantly reduced.

For instance, in our gold nanopropeller synthesis Protocol, we include an automatic scoring feature that evaluates the synthesis results. Users simply upload scanning electron microscopy (SEM) images of their synthesized particles; a built-in vision LLM processes the images and assesses nanopropeller quality from multiple angles (e.g., contour symmetry, structural clarity), returning an overall score or evaluation in seconds (**Supplementary Fig. S7**).

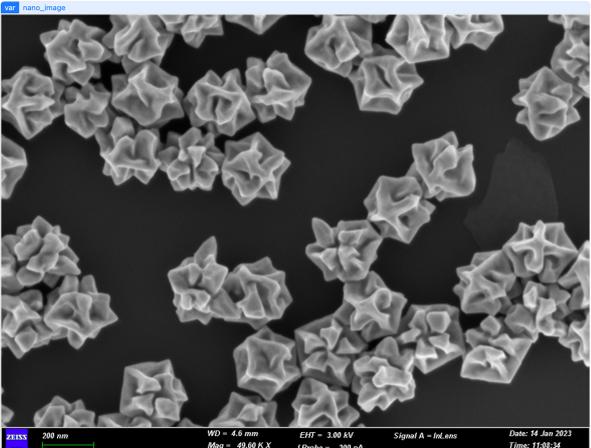

**Supplementary Fig. S7: Automated quality assessment of synthesized nanopropellers via AI.**
In the `nano_image` Field (①), users upload SEM images of the gold nanopropeller particles. The Protocol then calls an embedded, designer-configured vision LLM, which analyzes the images and populates the `nano_eval_results` Field (②) with an automated assessment of nanopropeller quality.

## 4. Discussions section for improved communication

Traditional protocols (e.g., printed manuals) offer few avenues for feedback or clarifications from users to the original authors. On the Airalogy Platform, however, each Airalogy Protocol has a dedicated Discussions section, enabling convenient, direct interaction between users and creators. For example, if someone encounters confusion while using our gold nanopropeller synthesis Protocol, they can post a query in the Discussions section, where the Protocol's author or experienced community members can provide solutions (**Supplementary Fig. S8a**). Conversely, if community users discover new insights or optimizations after applying the Protocol in their own labs, they can share their suggestions in the same forum (**Supplementary Fig. S8b**). The Protocol's author can then incorporate these ideas into future revisions. This iterative feedback cycle not only fosters a dynamic, collaborative scientific community—reducing communication barriers—but also aggregates expert input from diverse users, continually refining and improving each Protocol.

**Supplementary Fig. S8: Discussions section as a communication channel for Airalogy Protocol authors, users, and the wider scientific community.**
**a,** A user (e.g., the novice researcher Bob) runs into a question about the Protocol and posts it, inviting the author (e.g., Liping Huang) or other knowledgeable community members to respond. **b,** Community users (e.g., the expert researcher Alice) share further insights gained after applying the Protocol in their own work.

## 5. AI-driven real-time support

Even the most detailed Protocol cannot cover every possible question—especially those tangential to the main experimental focus. Consequently, newcomers might encounter unexpected issues that are not explicitly addressed in the Protocol. Moreover, while they can post inquiries in the Discussions board, the original Protocol author may not always be available to respond promptly.

By contrast, on the Airalogy Platform, the AI research copilot, Aira, is designed to adapt to each Airalogy Protocol. Aira functions as an "AI tutor", offering around-the-clock assistance specific to the Protocol in question. For instance, if users encounter problems while following the gold nanopropeller synthesis Protocol, they can pose their questions directly to Aira, which then provides immediate guidance based on the Protocol's contents (**Supplementary Fig. S9**).

How does the reaction temperature affect the experimental results in this case?

The reaction temperature plays a pivotal role in the synthesis of homochiral gold nanopropeller particles, and its impact on the experimental results can be understood through several key aspects:

1. **Seed Solution Synthesis (80°C):**
- The seed solution is synthesized at 80°C for 90 minutes. This elevated temperature ensures efficient reduction of gold ions, leading to the formation of stable gold seeds.
- If the temperature is too low, the reduction process may be inefficient, resulting in smaller or less uniform seeds. Conversely, if the temperature is too high, it could lead to rapid nucleation and uncontrolled growth, producing irregularly shaped nanoparticles.

2. **Gold Decahedra Synthesis (30°C):**
- The decahedra synthesis occurs at 30°C. This lower temperature is critical for maintaining the structural integrity and symmetry of the decahedra, which serve as precursors for the nanopropeller formation.
- Deviations from this temperature can affect the growth dynamics, potentially leading to structural defects or altered morphologies in the decahedra, which could propagate to the final nanopropeller structures.

3. **Homochiral Gold Nanopropeller Synthesis (30°C):**
- The final step of nanopropeller synthesis also takes place at 30°C. This temperature is crucial for controlling the orientation and growth of gold atoms, ensuring the development of the desired chirality and morphology.
- A higher temperature might cause faster reaction rates, potentially leading to less controlled growth and reduced chiral purity. A lower temperature might slow down the reaction, resulting in incomplete growth or poor yield.

4. **Impact on Chirality and Optical Properties:**
- The temperature influences the rate of reaction and the alignment of gold atoms during the growth process. These factors directly affect the chirality and optical properties of the nanopropellers.
- For instance, an optimal temperature ensures that the nanopropellers develop the correct handedness (chirality), which is essential for their applications in sensing, biomedicine, and catalysis.

5. **Potential Consequences of Temperature Variations:**
- **Higher Temperatures:** Could result in rapid nucleation, leading to aggregation or irregular shapes, thus reducing the quality and yield of the nanopropellers.
- **Lower Temperatures:** Might slow down the reaction, causing incomplete reduction of gold ions and poor formation of the nanopropeller structures.

In summary, precise temperature control is essential throughout the synthesis process. It ensures the formation of stable gold seeds, well-defined decahedra, and homochiral nanopropellers with the desired optical properties. Any deviation from the optimal temperatures can significantly affect the experimental outcomes, including the morphology, chirality, and overall performance of the synthesized nanoparticles.

**Supplementary Fig. S9: Users consult Aira—Airalogy Platform's virtual AI tutor—with questions about the gold nanopropeller protocol, receiving real-time answers.**

## 6. Global sharing from a single design—promoting a thriving scientific community and elevating authors' recognition

Once a cutting-edge experimental Protocol is published on the Airalogy Platform, it becomes immediately accessible worldwide. This enables the broader scientific community to rapidly adopt and apply the Protocol to their own research, amplifying its impact. At the same time, by sharing new methods as Airalogy Protocols, authors increase the visibility and reach of their associated paper publications. Community members can also "star" these Protocols (**Supplementary Fig. S 41**)—similar to GitHub's star feature—signaling popularity and offering recognition and motivation to Protocol creators.

Driven by this feedback mechanism, researchers from diverse fields are encouraged to continually submit their latest methods to the Airalogy Platform, fostering an ever-expanding, mutually supportive community. In this way, Airalogy helps create a vibrant global network that accelerates knowledge exchange, boosts reproducibility, and benefits scientists worldwide.

## S1.3 Supplementary Case Study S3: Transforming Airalogy Protocols into intelligent tutorials

Talented individuals drive technological innovation, and education forms the cornerstone of future scientific breakthroughs. In experimental sciences, hands-on training is especially crucial; mastering an experiment often requires students to perform it successfully themselves, thereby demonstrating genuine competency. However, the current landscape of laboratory education faces several challenges:

1. **Limited instructor resources:** In university lab courses, a single instructor may oversee dozens of students, leaving little time for individualized attention. As a result, many student questions go unanswered or receive insufficient clarification.
2. **Lack of student initiative:** Traditional instructor-led lab teaching can suppress students' motivation to learn independently. Students often wait passively for step-by-step guidance instead of taking the initiative. Moreover, even when they struggle with certain aspects of an experiment, they may hesitate to seek help due to fear or embarrassment.
3. **Difficulty in sharing and reusing high-quality tutorials:** Experienced instructors typically develop their own effective teaching methods, but these practices are rarely disseminated. Current tutorials (often in PDF format) provide only limited information, and even if shared, it can be challenging for other teachers to adapt them directly for their own classes.

To address these challenges, we present a case study on transforming an Airalogy Protocol into an intelligent tutorial that supports real-world teaching scenarios.

### S1.3.1 Applying the Airalogy Protocol to training on public shared laboratory instruments

Public shared laboratory instruments serve as essential resources for both research and education. However, mastering the use of high-end instruments (many of which are structurally complex and operationally intricate) typically requires extensive instruction before a user is granted access rights. This training can be time-consuming and resource-intensive, and even after formal instruction, achieving true proficiency often demands significant hands-on practice. In the meantime, improper usage can reduce experimental quality and risk damaging sensitive equipment.

While conventional usage protocols can offer basic guidance, they usually do not meet the real-time needs of beginners, who often encounter unique, context-specific challenges. These static documents may also become outdated or fail to address complex scenarios. Consequently, there is a pressing need for a more flexible, interactive training approach to ensure the efficient utilization of shared instruments, minimize equipment damage, and promote responsible sample usage.

As a solution, we propose transforming Airalogy Protocols into continuously updated, intelligent tutorials. This approach is demonstrated here using the BD FACSAria Fusion Flow Cytometer as an example. By providing new users with structured, step-by-step guidance through an Airalogy Protocol, instrument managers and educators can ensure that beginners operate the instrument correctly, maintain its integrity, and consistently achieve reliable experimental outcomes.

## S1.3.2 Unique advantages of using the Airalogy Protocol for teaching

### 1. Step-by-step instrument instructions with rich multimedia

Inside an Airalogy Protocol, every action required to operate an instrument can be laid out in clear, sequential steps (**Supplementary Fig. S10a**). Unlike traditional paper manuals (which rely almost exclusively on text and static images), Airalogy Protocols support multiple media formats, giving learners a far more intuitive grasp of each procedure. Consider the start-up sequence for a BD FACSAria Fusion Flow Cytometer, where several subsystems (main console, laser module, fluidics module, etc.) must be activated in a precise order. Instead of describing these actions only in words, the Protocol can embed:

- **Images** that highlight the appearance and location of each subsystem, helping beginners identify components instantly (**Supplementary Fig. S10b**).
- **Audios** that capture the normal operating sound of the fluidics compressor, enabling users to compare live noise with a "healthy" reference and verify proper function (**Supplementary Fig. S10c**).
- **Videos** that show what a stable fluidics stream looks like, so novices can visually confirm readiness for downstream cell-sorting steps (**Supplementary Fig. S10d**).

With these interactive elements, an Airalogy Protocol becomes a realistic, accessible teaching tool—conveying critical details that conventional text and images alone cannot deliver.

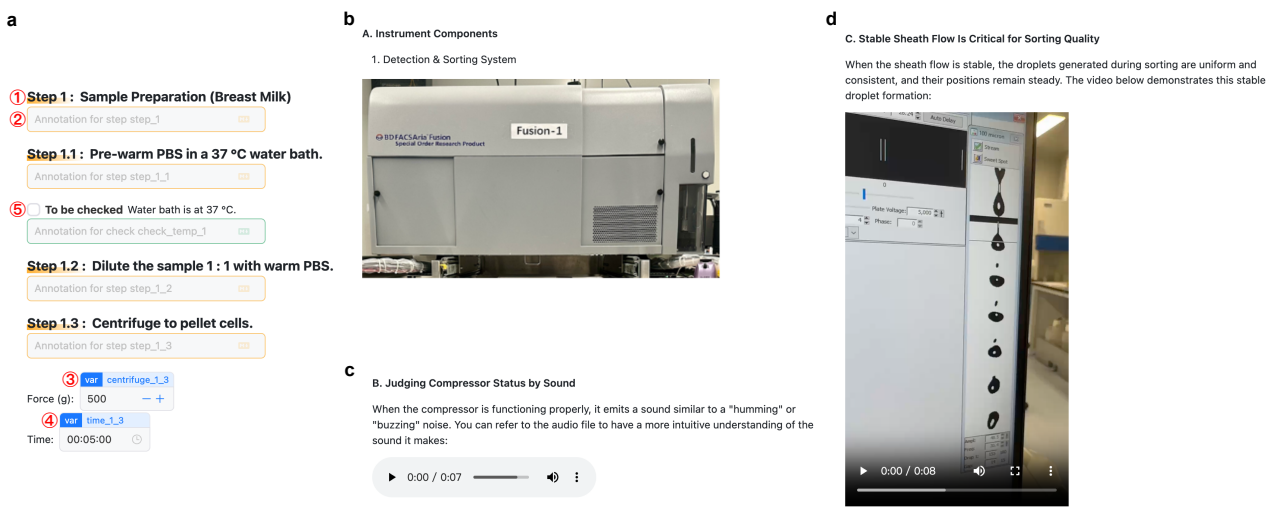

**Supplementary Fig. S10: Detailed, multimedia teaching guidance delivered through an Airalogy Protocol.**

**a,** The Protocol's step-by-step structure. A major step—e.g., Step 1 (①)—can be broken into finer sub-steps (Step 1.1, 1.2, 1.3 …) to give learners granular instructions. Each step is paired with an annotation box (②) in which learners may log deviations or anomalies observed during execution. Protocol designers can embed Variable-type Fields for recording key experimental parameters or results; here, two Var Fields capture the centrifugation force (③) and duration (④) used while preparing flow-cytometry samples. Checkpoint-type Fields (⑤) prompt users to confirm completion of critical actions. **b–d,** Beyond text, designers may add images, audios, or videos to convey practical details that are difficult to express verbally—such as identifying instrument components, recognizing normal operating sounds, or visually verifying fluid-flow stability.

## 2. Capturing learner-generated data within the Protocol

Traditional lab manuals merely deliver instructions and fail to log a learner's actual actions or results. An Airalogy Protocol, however, can record real-time inputs and return dynamic feedback. For every step (**Supplementary Fig. S10a ①**), a dedicated annotation box lets users note deviations from the standard procedure (**Supplementary Fig. S10a ②**). Designers can also add Variable-type Fields (Var Fields) to log key parameters and outcomes—for example, in the flow-cytometry training Protocol, two Var Fields can be used to capture centrifugation force and duration when preparing cell samples (**Supplementary Fig. S10a ③, ④**); and Checkpoint-type Fields (Check Fields) to flag critical checkpoints that learners must confirm (**Supplementary Fig. S10a ⑤**). By combining these elements, a Protocol can systematically capture diverse data on learner behavior, creating a rich foundation for further automated assessment of each training session.

## 3. Automated learning assessment through real-time operational and outcome records

Because an teaching Protocol can capture every action and result produced by the learner, it can also deliver data-driven, automated assessment. In the flow-cytometry training Protocol, for example, the final step (Step 5) requires the learner to clean and shut down the instrument correctly. A series of Var Fields records whether the student: (1) flushed the fluidics lines, (2) stopped the fluid flow, (3) retracted the nozzle, (4) completed the cleaning program, and (5) powered down the hardware (**Supplementary Fig. S11a ①–⑤**). An Assigner then tallies these inputs (awarding one point per completed task) and generates both a numerical score for Step 5 (**Supplementary Fig. S11a ⑥**) and a transparent justification (**Supplementary Fig. S11a ⑦**).

Extending this logic, the Protocol defines five core steps (sample preparation, cell staining, flow-cytometry analysis, cell sorting, and post-run cleanup) each scored out of five points. After the learner finishes all steps, the Protocol automatically calculates per-step scores and an overall total (**Supplementary Fig. S11b ⑧**), presenting the results in a radar plot for easy visualization (**Supplementary Fig. S11b ⑨**). An integrated LLM then summarizes the student's performance in natural language, highlighting strengths and areas for improvement (**Supplementary Fig. S11b ⑩**). By enabling automated scoring, the Protocol evolves from a static instruction set into an interactive, feedback-rich teaching tool, closing the loop between Protocol and learner and enabling a more adaptive learning experience.

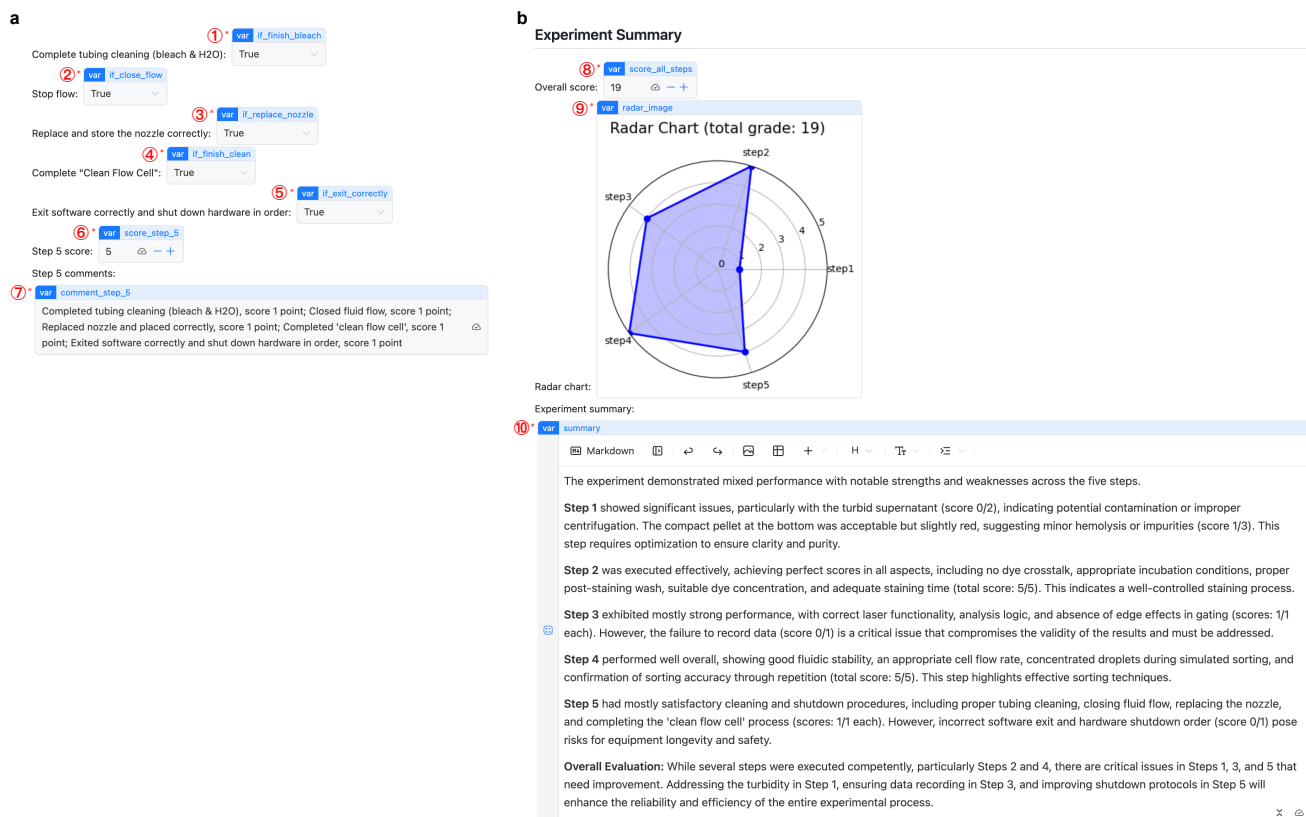

**Supplementary Fig. S11: Embedding automated assessment within an Airalogy Protocol.**
**a,** Example for Step 5: Post-run cleaning and shutdown. The Protocol designer adds five assessment Fields that capture whether the learner has: (1) flushed the fluidics lines, (2) stopped the fluid flow, (3) retracted the nozzle, (4) completed the cleaning program, and (5) powered down the hardware (①–⑤). An embedded Assigner automatically computes the learner's score for Step 5 (⑥) and records the rationale (⑦), based strictly on the values entered in these Fields—providing objective, behavior-based evaluation. **b,** Generating an overall performance summary. Using additional Assigners, the Protocol aggregates the scores from all five main steps (5 points each; total = 25 points) to yield a composite score (⑧). Custom plotting code then renders a radar chart that visually displays the learner's performance across all steps (⑨). Finally, an embedded LLM produces an AI-generated summary (⑩) that highlights strengths, identifies weaknesses, and offers tailored suggestions for improvement.

## 4. Real-time instrument guidance with Aira

One of the biggest advantages of an Airalogy Protocol–based tutorial over a traditional instrument manual is its interactive nature. In addition to static instructions, learners can consult Aira, the Airalogy Platform's AI research copilot, for continuous, on-demand support—acting like a virtual tutor. Whereas a human instructor may have limited time to answer questions, Aira is available 24/7 to provide immediate feedback.

For example, if a student forgets to switch off the flow cytometer's laser after completing the experiment, they can simply ask Aira for help. Aira not only offers reassuring feedback to ease any anxiety about the mistake but also supplies clear, step-by-step instructions for rectifying it (**Supplementary Fig. S12**). This not only spares human instructors from routine queries but also encourages beginners to ask any question (no matter how basic, or how "stupid") without hesitation. Because Aira's responses are always available and endlessly patient, students benefit from a less intimidating learning environment, ultimately promoting greater initiative and engagement.

I forgot to turn off the flow cytometer's laser after I finished my experiment, and now I'm super anxious. Will my professor get mad at me? What should I do? 😬

It sounds like you're feeling really stressed about forgetting to turn off the flow cytometer's laser. First, take a deep breath — it's not the end of the world! Here's what you can do:

What You Should Do Now:

1. **Turn Off the Laser**: If possible, go back to the machine and turn off the laser immediately. This will prevent any unnecessary wear or damage to the equipment.
2. **Inform Your Professor**: It's important to be honest. Let your professor know that you forgot to turn off the laser and explain what happened. Most professors understand that mistakes happen, especially if it's an oversight.
3. **Apologize Sincerely**: A simple apology goes a long way. Show that you're aware of the mistake and are taking steps to avoid it in the future.
4. **Learn From It**: Make sure to set a reminder or note for yourself next time to always double-check that all equipment is turned off after use.

What Might Happen Next:
- Your professor might be upset initially, but they'll likely appreciate your honesty and willingness to learn from the mistake.
- They may offer guidance on how to avoid similar issues in the future, which could help you grow as a researcher.

Final Thoughts:
Remember, everyone makes mistakes. The key is how you handle them. By being proactive and honest, you show responsibility and maturity. You've got this! 😊

If you'd like, I can also help you draft a message to your professor. Just let me know!

**Supplementary Fig. S12: Learner interacting with Aira, the Airalogy Platform's AI co-pilot.** At any point during the tutorial, users can pose questions to Aira, which functions as a virtual AI tutor. Drawing on the Protocol's context, Aira delivers adaptive, real-time answers—available 24/7.

## 5. Design once, reuse globally: advancing educational access and equity

Because an intelligent tutorial built on an Airalogy Protocol is itself still an Airalogy Protocol, it can be shared worldwide on the Airalogy Platform just like any other Protocol. This greatly enhances accessibility and educational fairness. On one hand, creating a high-quality, interactive teaching Protocol typically requires deep expertise; when such a Protocol is globally shared, regions with limited educational resources or weaker teaching faculties gain instant access to top-tier instructional materials. On the other hand, because Aira (the AI virtual tutor) is available through the Airalogy Platform wherever internet access exists, learners can receive consistent, high-level feedback and answers around the clock. Even if local educators are unable to address certain questions, Aira can fill the gap with reliable, 24/7 assistance—substantially improving both the accessibility and fairness of educational opportunities worldwide.

## 6. Community-driven ecosystem for sustainable improvement

- **Broad access to quality teaching Protocols:** As the Airalogy community expands, we anticipate the creation and sharing of Airalogy Protocols tailored for various disciplines (not limited to experimental teaching, but also for literature, history, language, computer programming, mathematics, the natural sciences, etc.). For instance, **Supplementary Case Study S4** demonstrates how an Airalogy Protocol can be applied to English-language teaching. In the future, anyone seeking to learn a specific skill can search the Airalogy Hub for a well-crafted teaching Protocol, enabling rapid and efficient onboarding. This approach fundamentally improves access to high-quality instruction across multiple fields, boosting both teaching effectiveness and learning outcomes. Furthermore, as the diversity of teaching-oriented Protocols expands to encompass a growing array of subjects and specific educational needs, we foresee the possibility of implementing a certificate system. Each time a learner completes a particular Protocol—or a series of Protocols—and passes the associated assessments, Airalogy could award a qualification certificate. Such certificates would validate that the learner has mastered a particular skill, qualifying them for further pursuits that require that competency. Because Protocols can be flexibly created and customized by the community, these certificates could also be issued through a decentralized, community-driven process. In this way, Airalogy Protocols could cover a wide spectrum of disciplines and specialized areas, potentially simplifying how schools, businesses, and other institutions recognize, evaluate, and certify various skill sets.

- **Continuous improvement through feedback:** An Airalogy Protocol can continually evolve based on real-world feedback (as discussed in **Supplementary Case Study S5**). When a teaching Protocol is widely used by learners worldwide, their experiences and suggestions (could be posted in the corresponding Discussions section) help the Protocol's maintainers refine it. This creates a "living" tutorial that evolves over time. By pooling insights from the community, more and more high-quality tutorials emerge. The Platform's collaborative environment is especially beneficial for resolving common issues in experimental Protocols, such as reproducibility challenges on different instrument models or variations in specific lab conditions. Users can share practical workarounds and adaptations, further enhancing each Protocol's value.

- **Incentives for stakeholders beyond researchers:** The Airalogy Platform's community-driven approach motivates a broader range of stakeholders—beyond scientists—to participate in the Protocol ecosystem. For example, reagent, kit, or instrument manufacturers can publish carefully designed tutorials (in the form of an Airalogy Protocol) to help users quickly master their products. If researchers find these tutorials effective in obtaining reliable experimental results, they will naturally gravitate toward those reagents, kits, or instruments in the future. This creates a win–win situation: manufacturers boost sales by providing superior onboarding Protocols, while researchers gain rapid, reliable access to high-quality experimental guidance. Such mutual benefits further accelerate the growth and vitality of the Airalogy ecosystem.

## S1.4 Supplementary Case Study S4: Airalogy Protocols serving as specialized AI applications

### S1.4.1 Common AI research services and personalized AI needs

Although scientific research are often highly specialized and complex, many fundamental tasks and requirements in research share common features. To meet these universal needs, the Airalogy Platform—through its AI research co-pilot, Aira—offers built-in AI services such as automated generation of Airalogy Protocols, searching and recommendation, conversational Q&A, and data analysis (**Fig. 4**). Nevertheless, because Airalogy is designed to be a multidisciplinary platform serving a vast range of fields, users in different disciplines, laboratories, or individual roles may have highly diverse AI requirements. Obviously, as the platform provider, we cannot possibly deliver a tailored solution for every niche AI demand.

To address this challenge, we designed the Airalogy Protocol framework to let users easily create custom AI functionalities by embedding AI modules into their Protocols (for detailed instructions, see **Supplementary Method S2.2.4**). More importantly, each AI-augmented Protocol automatically renders a visually appealing and user-friendly Recording Interface (**Fig. 2b**). This means users can focus solely on defining the input–output relationships and data-processing logic required by their AI features, without worrying about how to implement or design a front-end user interface. By dramatically lowering the barrier to developing AI applications on Airalogy Protocols, we anticipate a flourishing ecosystem of creative AI-driven tools contributed by individuals with various areas of expertise.

Equally important, these AI applications inherit the standard data format of ordinary Airalogy Protocols. Consequently, each time an AI application is used, its outputs are captured as Airalogy Records, allowing for straightforward archiving or further analysis. This not only ensures immediate benefits from the AI's capabilities but also helps collect valuable usage data, which can be used to train more robust AI models in the future. Moreover, like any other Protocol, these AI applications can be shared globally on the Airalogy Platform, enabling their creators to easily distribute their work and innovations to a worldwide audience.

Within the Airalogy team, we have already developed numerous AI-driven applications based on Airalogy Protocols and adopted them for daily use. In the sections that follow, we present several examples from different fields to demonstrate the flexibility and power of AI applications built on this framework.

### S1.4.2 Case studies of AI applications built on Airalogy Protocols

### S1.4.2.1 AI Application 1: Assisting English teachers with essay grading

In **Supplementary Case Study S3**, we demonstrated how the Airalogy Platform can facilitate intelligent teaching for experiments. However, AI-powered tools can also benefit other academic fields. For example, grading student essays is often one of the most time-consuming tasks for English teachers. A typical teacher may need to correct dozens of essays, each requiring several minutes (up to more than ten minutes) to review manually. To address this, we developed an AI-based essay-grading application through an Airalogy Protocol (**Supplementary Fig. S13**). Teachers can simply paste a student's essay into the Protocol, and the integrated AI automatically provides a grade and suggested revisions, reducing the correction time for each essay to just a few seconds and significantly saving teachers' time and energy. Moreover, each time this Protocol is used for grading, the outcome is saved as an Airalogy Record. Thus, once the teacher has graded a number of student essays, these Records can be easily reviewed or exported to enable further analysis of individual student performance or the class's overall writing trends.

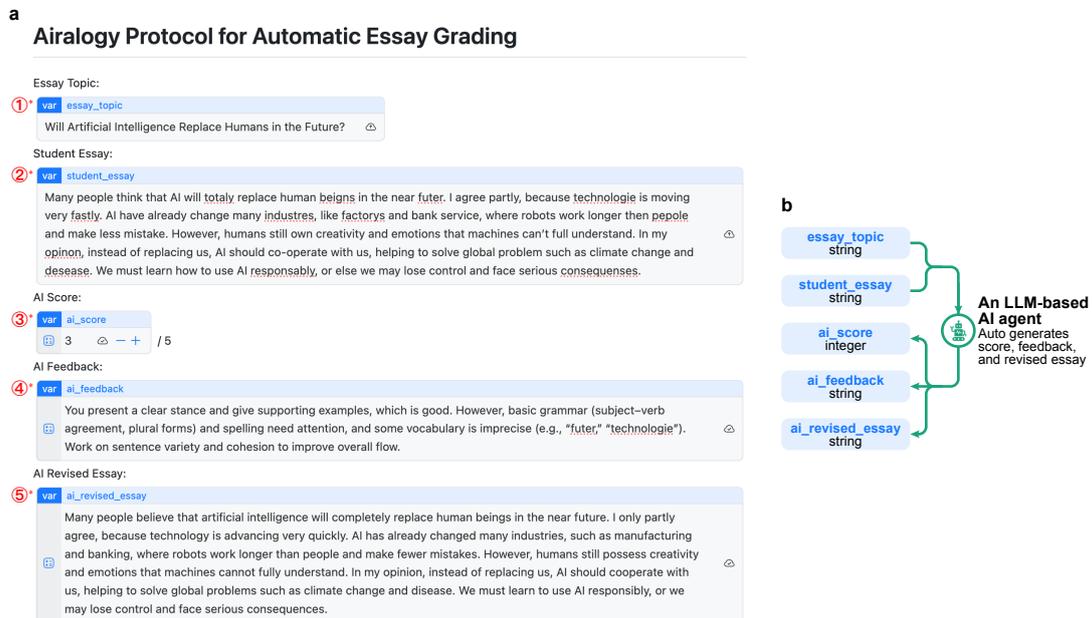

**Supplementary Fig. S13: Using an Airalogy Protocol to assist English teachers with student essay grading.**
**a,** Screenshot of the Protocol Recording Interface. We designed 5 key Airalogy Fields: `essay_topic` (①), `student_essay` (②), as well as `ai_score` (③), `ai_feedback` (④), and `ai_revised_essay` (⑤). When using the Protocol, a teacher only needs to fill in the first two Fields with the essay topic and the student's essay. The Protocol then automatically invokes its embedded LLM to generate a grade and suggestions—populating the remaining Fields (③, ④, ⑤) with the essay score, feedback, and a revised version, respectively. **b,** Diagram showing how each Field is logically linked within the Protocol through the embedded LLM.

## S1.4.2.2 AI Application 2: Helping doctors transcribe physician–patient audio recordings into text

In clinical research settings, doctors often need to transform patient–doctor conversations into text, a task that can be time-consuming and labor-intensive if done manually. To address this issue, we developed an Airalogy Protocol–based AI application that automatically transcribes physician–patient audio files. By simply uploading an audio file recorded a consultation, the Protocol's embedded AI differentiates between the speakers and organizes their statements into a coherent transcript (**Supplementary Fig. S14**). This approach can convert an hour of audio into written text in just a few minutes.

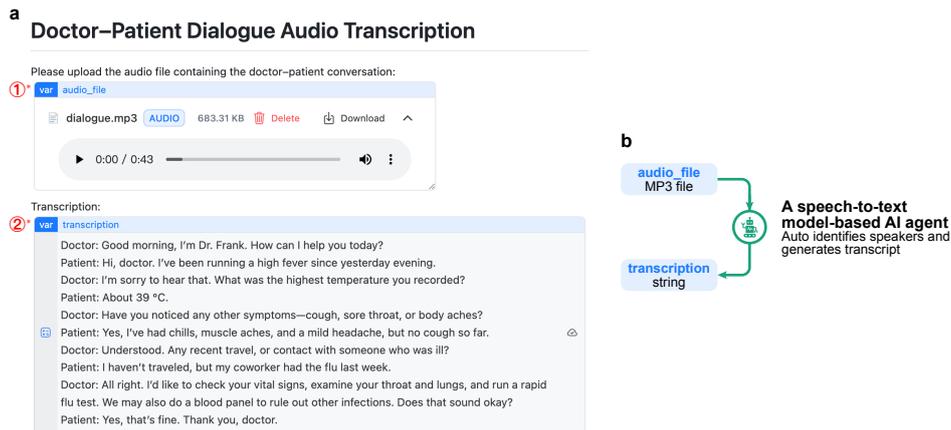

**Supplementary Fig. S14: Using an Airalogy Protocol to help clinical doctors transcribe physician–patient audio recordings into text.**
**a,** Screenshot of the Protocol Recording Interface. This Protocol defines two Airalogy Fields: `audio_file` (①) and `transcription` (②). Once the user uploads an MP3 file containing a physician–patient conversation, the Protocol invokes a speech-to-text AI model to convert the recording into a structured transcript. The resulting text is assigned to `transcription` and displayed in the interface. **b,** A diagram illustrating how these Fields are linked by AI logic.

### S1.4.2.3 AI Application 3: Code-free generation of data reports

As LLMs continue to advance, data analysis may increasingly shift from manually executing hard-coded scripts to having AI autonomously interpret the structure of a dataset and generate the necessary code for analysis[65]. This transition could drastically lower technical barriers, allowing researchers without programming skills to analyze data effortlessly. To showcase this possibility, we developed an Airalogy Protocol that demonstrates automated data-report generation (**Supplementary Fig. S15**).

In this Protocol, users upload a CSV file and provide relevant instructions or analysis goals. The Protocol's embedded AI agent then interprets both the data structure and the user's stated objectives, automatically drafting Python code to visualize the CSV data and create a report. The code is executed within the Protocol, producing a final data analysis output. If the user finds the report unsatisfactory, they can submit further feedback, prompting the AI agent to refine its approach and generate an updated report.

![Airalogy]

**a** — Upload a CSV File ①

Please upload your csv file here:

csv_file

Predict Hair Fall.csv  CSV  84.83 KB   ⊙ Preview  🗑 Delete  ⬇ Download

| # | Id | Genetics | Hormonal Changes | Medical Conditions | Medications & Treatments | Nutritional Deficiencies |
|---|-----|----------|------------------|--------------------|--------------------------|--------------------------|
| 1 | 133992 | Yes | No | No Data | No Data | Magnesium deficiency |
| 2 | 148393 | No | No | Eczema | Antibiotics | Magnesium deficiency |
| 3 | 155074 | No | No | Dermatosis | Antifungal Cream | Protein deficiency |
| 4 | 118261 | Yes | Yes | Ringworm | Antibiotics | Biotin Deficiency |
| 5 | 111915 | No | No | Psoriasis | Accutane | Iron deficiency |

Total: 999 rows  [1] 2 3 4 5 6 7 … 200 ›  5 / page ⌄

**📊 Data Analysis Intent**

Please describe how you would like to analyze the data in the CSV file:

analysis_intent

What is the proportion of patients with hair loss in different age groups? It can be presented through a bar chart or a pie chart.

Columns:
- Genetics: Indicates whether the individual has a family history of baldness (Yes/No).
- Hormonal Changes: Indicates whether the individual has experienced hormonal changes such as Alopecia Areata, Thyroid Problems, Scalp Infection, Psoriasis, Dermatitis, etc.
- Medical Conditions: Lists specific medical conditions that may contribute to baldness, such as Alopecia Areata, Thyroid Problems, Scalp Infection, Psoriasis, Dermatitis, etc.
- Medications & Treatments: Lists medications and treatments that may lead to hair loss, such as Chemotherapy, Heart Medication, Antidepressants, Steroids, etc.
- Nutritional Deficiencies: Lists nutritional deficiencies that may contribute to hair loss, such as Iron deficiency, Vitamin D deficiency, Biotin deficiency, Omega-3 fatty acid deficiency, etc.
- Age: Represents the age of the individual.
- Stress: Indicates the stress level of the individual (Low/Moderate/High).
- Poor Hair Care Habits: Indicates whether the individual practices poor hair care habits (Yes/No).
- Environmental Factors: Indicates whether the individual is exposed to environmental factors that may contribute to hair loss (Yes/No).
- Smoking: Indicates whether the individual smokes (Yes/No).
- Weight Loss: Indicates whether the individual has experienced significant weight loss (Yes/No).
- Baldness (Target): Binary variable indicating the presence (1) or absence (0) of baldness in the individual.

**b** — Analysis Report Generator ②

Here is the analysis report for the data in the CSV file:

initial_report

⊞ Markdown …

# Analysis of Hair Loss Proportions Across Age Groups

## Introduction

This report aims to analyze the proportion of individuals experiencing hair loss across different age groups. The analysis is based on a dataset containing various factors that may contribute to baldness, including genetics, hormonal changes, medical conditions, medications, nutritional deficiencies, stress levels, and lifestyle choices. The data was processed using Python's Pandas library, and visualizations were created using Matplotlib.

## Method Description

The data processing involved the following steps:

1. **Data Loading:** The dataset was loaded from a CSV file.
2. **Age Group Creation:** Individuals were categorized into age groups using predefined bins.
3. **Proportion Calculation:** The mean of the binary hair loss variable was calculated for each age group to determine the proportion of individuals with hair loss.
4. **Visualization:** A bar chart was generated to visually represent the proportions of hair loss across age groups.

## Results Display

### Proportion of Patients with Hair Loss in Different Age Groups

**📊 User Feedback** ③

Please provide your feedback after you have reviewed the analysis report:

user_feedback

Could you please make the text appear on top of the bar and use a colorful color scheme?

**📝 Improved Report Version**

Here is the improved analysis report:

improved_report

⊞ Markdown …

## Analysis of Hair Loss Proportions Across Age Groups

### Introduction

This report aims to analyze the proportion of individuals experiencing hair loss across different age groups based on a dataset containing various factors related to baldness. The dataset includes variables such as genetics, hormonal changes, medical conditions, medications, nutritional deficiencies, stress levels, age, hair care habits, environmental factors, smoking status, and weight loss. The analysis employs statistical methods and visualizations to present insights into the relationship between age and hair loss.

### Method Description

The data processing and analysis involved the following steps:

1. **Data Loading:** The dataset was loaded from a CSV file using Pandas.
2. **Age Group Creation:** Age was categorized into predefined groups using the `pd.cut()` function.
3. **Proportion Calculation:** The mean of the 'Hair Loss' binary variable was calculated for each age group to determine the proportion of individuals with hair loss.
4. **Visualization:** A bar chart was created using Matplotlib to visually represent the proportions of hair loss across different age groups.

### Results Display

### Proportion of Patients with Hair Loss in Different Age Groups

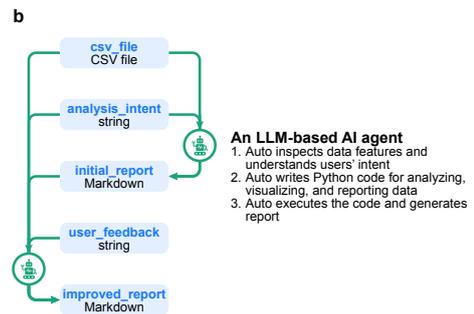

An LLM-based AI agent
1. Auto inspects data features and understands users' intent
2. Auto writes Python code for analyzing, visualizing, and reporting data
3. Auto executes the code and generates report

**Supplementary Fig. S15: An Airalogy Protocol implementing code-free data-report generation. a,** Screenshot of the Protocol Recording Interface. This Protocol defines 5 Airalogy Fields: `csv_file`, `analysis_intent`, `initial_report`, `user_feedback`, and `improved_report`. After uploading a CSV file, the user enters a desired analysis goal (①). The Protocol's LLM-based AI agent then interprets the CSV file, writes Python code to visualize and analyze the data, and executes that code—ultimately generating a data-analysis report (②). If needed, the user can provide additional feedback, prompting the AI agent to incorporate those requests and produce an improved report (③).The CSV data source used in the example shown in the figure is from Kaggle[66]. **b,** A diagram illustrating how these Fields are interconnected through the AI logic embedded in the Protocol.

### S1.4.2.4 AI Application 4: AI-automated peer review

The rapid rise in published research papers places an immense burden on peer reviewers. To address this challenge, the AAAI 2026 conference on AI plans to introduce an AI-powered peer-review assessment system, aiming to automate the paper-review process[67]. Here, we showcase an Airalogy Protocol (**Supplementary Fig. S16**) that embeds our recently published automated review system—DeepReview[43]. Users simply upload a PDF of their paper into the Protocol, and the embedded DeepReview service returns a review outcome, which are then displayed in the Protocol interface.

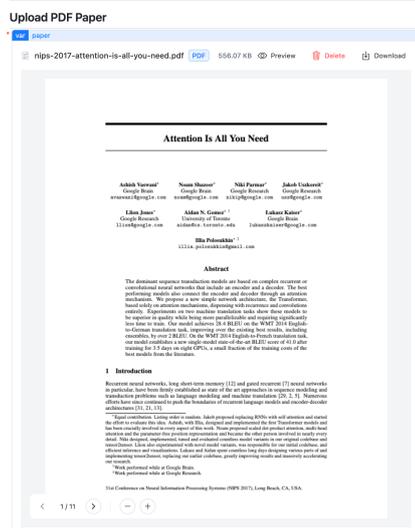

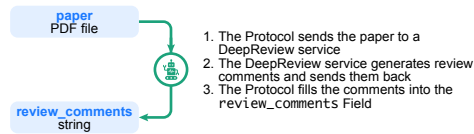

1. The Protocol sends the paper to a DeepReview service
2. The DeepReview service generates review comments and sends them back
3. The Protocol fills the comments into the `review_comments` Field

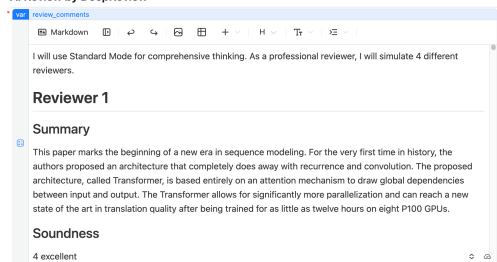

**Supplementary Fig. S16: An Airalogy Protocol for AI-automated peer review.**
**a,** Screenshot of the Protocol Recording Interface. The Protocol defines two Airalogy Fields: `paper` (①) and `review_comments` (②). After uploading a PDF in Field ①, the Protocol automatically invokes the DeepReview AI service, which analyzes the paper and returns its review. That feedback is then displayed in Field ②. **b,** A schematic showing how these Fields are interconnected through AI logic within the Protocol.

### S1.4.2.5 AI Application 5: Controlling automated lab equipment through natural language

To enable fully AI-automated research via the Airalogy Platform (discussed in **Supplementary Case Study S1**), an Airalogy Protocol must be able to communicate with automated lab devices—sending instructions and receiving experimental outcomes. In this case study, we demonstrate a Protocol designed for proof of concept, using the Opentrons OT-2 liquid-handling robot as an example.

The OT-2 can be networked to listen for incoming pipetting instructions (with a specific data structure[68]) on a designated port. Upon receiving these instructions, it executes the pipetting task and returns the results over the internet. However, one main barrier typically prevents widespread adoption by non-experts: users lacking programming experience often cannot craft pipetting instructions that conform to the OT-2 syntax.

To address these challenges, we developed the Airalogy Protocol (**Supplementary Fig. S17**). In this Protocol, users simply express their pipetting requirements in natural language. The built-in LLM automatically interprets these instructions and translates them into executable pipetting commands for the OT-2 robot. The Protocol then sends these commands to the OT-2, which performs the pipetting operation and returns the results to the Protocol.

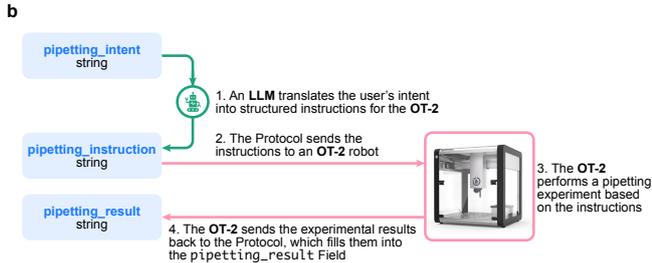

**Supplementary Fig. S17: Using an Airalogy Protocol to control an automated pipetting robot. a,** Screenshot of the Protocol Recording Interface. This Protocol defines three Airalogy Fields: `pipetting_intent` (①), `pipetting_instruction` (②), and `pipetting_result` (③). Users simply describe their pipetting needs in Field ① using natural language. The Protocol then employs an embedded LLM AI agent to translate that description into valid OT-2 instructions, which are transmitted to the robot. Once the OT-2 completes the task, it returns the experimental results, which the Protocol records in Field ③. **b,** A schematic showing how the Fields, LLM, and OT-2 interact to achieve automated pipetting.

## S1.4.3 Airalogy Protocol as a new paradigm for future AI application development

A standardized, open protocol that is widely adopted by the community can both promote a vibrant ecosystem and stimulate the creative potential of its users. For instance, model context protocol (MCP; initially proposed by Anthropic)[69] has emerged as an open standard for unifying communication between LLMs and external data sources or tools. Thanks to the growing MCP ecosystem, thousands of MCP-based services have been developed (for example, the interfaces for GitHub repository management and file operations).

Likewise, our proposed Airalogy Protocol (see **Supplementary Method S2.2** for detailed specifications) could serve not only as a standard for recording data in a structured electronic format but also as a new paradigm for developing AI applications. Key advantages include:

1. **Automatic interface rendering enables users to focus on business logic:** Within each Airalogy Protocol, users only need to create an Airalogy Markdown file (**Supplementary Method S2.2.2**) and define the appropriate Airalogy Fields that will receive inputs and generate outputs for AI operations. By linking these Fields through Assigners (**Supplementary Method S2.2.4**), which encapsulate AI-related logic, the Airalogy Platform automatically renders an interactive user interface. Users with limited (or no) front-end experience can therefore build AI applications simply by deciding how to retrieve data from upstream Fields, process the data with AI agents, and store the resulting output in downstream Fields.

2. **Seamless standardized record-keeping:** Using an AI application built on an Airalogy Protocol essentially involves populating the relevant upstream Airalogy Fields (such as `essay_topic` and `student_essay` in the automated essay-grading app) triggering the AI agent's processes (via Assigners), and then retrieving the outputs (e.g., `ai_score`, `ai_feedback`, and `ai_revised_essay`). After each AI operation, these inputs and outputs can be saved as an Airalogy Record. This approach provides a straightforward method for capturing, managing, and analyzing all user interactions and results from an AI application. The accumulated data also serve as valuable resources for ongoing refinement—following the method illustrated in **Supplementary Case Study S5**, where Airalogy Records drive improvements in the corresponding Airalogy Protocol.

3. **Easy coordination of external tools and resources:** An Airalogy Protocol can easily call upon external services, tools, or devices through internet-based APIs (via code logic in Assigners), effectively distributing various operational tasks across these external resources (**Supplementary Fig. S18**). For example, in the five AI applications presented within this case study, the LLM functions do not reside directly in the Protocol. Instead, the Protocol calls an external LLM service provided by a third-party vendor. As the capabilities and diversity of global AI services continue to expand, users can integrate these high-quality external resources through Airalogy Protocols, rapidly designing powerful AI applications.

4. **Global reusability from a single design:** Any AI application developed on top of an Airalogy Protocol is, at its core, still an Airalogy Protocol. Thus, it can be shared on the Airalogy Platform just like a normal Protocol. Users can effortlessly disseminate their carefully crafted AI applications to a worldwide audience, receiving feedback that drives further improvements while gaining recognition from a global community.

5. **Chaining multiple AI applications:** Because each AI application is itself an Airalogy Protocol, multiple applications can be assembled into a larger pipeline or workflow under the Airalogy Protocol Pipeline (**Fig. 3**) or Airalogy Protocol Workflow (**Fig. 5**) paradigm. We foresee users creating increasingly sophisticated "AI Application Pipelines" or "AI Application Workflows" by integrating tools from different Protocol designers, enabling more complex and powerful functionalities.

Looking ahead, as the Airalogy community grows, the Airalogy Protocol development standard will unleash the creativity and enthusiasm of its developer community. We anticipate a wide array of well-designed, robust AI applications—addressing both general and specialized needs—will be developed and shared across academia, industry, and society at large, benefiting countless users worldwide.

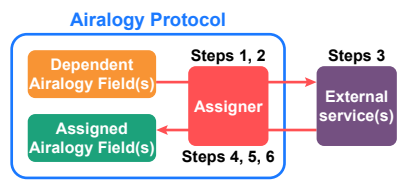

**Step 1:** The Assigner processes the data in the dependent Field(s) into the request format required by the external service(s)

**Step 2:** The Assigner sends the request to the external service(s) over the internet

**Step 3:** The external service processes the request and generates result data

**Step 4:** The external service returns the result data to the Assigner over the internet

**Step 5:** The Assigner processes the data returned by the external service(s) into the format required by the assigned Field(s)

**Step 6:** The Assigner populates the corresponding assigned Field(s) with this processed data

**Supplementary Fig. S18: A general architecture for integrating external resources into an Airalogy Protocol.**

Using a meticulously designed Assigner, data (including instructions) from upstream Airalogy Fields (the dependent Fields) can be sent via an internet-based API to external services (e.g., external computing services, LLMs, automated instruments, robots, etc.). After processing the input data, the external service returns its results to the Assigner, which then populates the appropriate downstream Fields (the assigned Fields) within the Protocol.

## S1.5 Supplementary Case Study S5: Evolving Airalogy Protocols into standard operating procedures through continuous use

Scientific experiments require protocols that are both precise and adaptable[70,71]. Yet, traditional static protocols often fail to meet the dynamic demands of modern research[72]. These protocols are designed as rigid guides, disconnected from the data they generate. A significant limitation is their inability to refine protocol with past experimental records, which are often scattered across diverse contexts both in space and time[73,74].This fragmented record-keeping deprives researchers of essential insights for refining experimental methods and adjusting research objectives.

This disconnection poses challenges in two key areas. First, in discovery-driven research, where protocols are uncertain and subject to frequent adjustments, reproducibility is often compromised by the absence of a structured framework for recording and learning from experimental variations[75]. Second, mature protocols, despite their established structures, frequently require modifications to suit specific applications. These adjustments are typically made through *ad hoc* methods, relying on individual expertise or incomplete historical data, resulting in inefficiencies and inconsistent outcomes.

To address these limitations, we introduce a novel protocol paradigm—the Airalogy Protocol—which can be iteratively evolved into robust standard operating procedures (SOPs) through continuous use. By systematically linking each Protocol version to its corresponding experimental Airalogy Records and refining it based on real-time feedback, the Airalogy Protocol establishes a closed-loop system for ongoing improvement. This integrated approach ensures that every experimental outcome and modification is captured and actively utilized, thereby enhancing both the reproducibility and adaptability of scientific research.

### S1.5.1 Coupling protocols with their experimental data: a key innovation of the Airalogy Protocol

A critical shortcoming of traditional protocols lies in their detachment from the experimental data they generate. To illustrate, a PDF-based protocol typically exists as a stand-alone document, while the associated experimental data—produced by following that protocol—are maintained separately, perhaps in another software system or on a researcher's local computer. This setup frequently makes it difficult to trace how a protocol leads to the specific dataset obtained under its guidance.

By contrast, in the Airalogy Protocol paradigm, each experimental outcome is recorded as an Airalogy Record, which is structurally linked to the Airalogy Protocol itself. Specifically, the data fields within each Record (e.g., process annotations, key results, and checkpoint validations) derive from the Airalogy Fields defined in the Protocol (see below). These Fields serve as the "bridge" between a Protocol and its Records, ensuring that all experimental information is captured and contextualized in a standardized manner.

Below, we provide an example from this case study: a specialized Airalogy Protocol designed to purify lipoylated H protein (**Supplementary Fig. S19**). By defining various Airalogy Fields—such as Steps, Vars, and Checks (**Supplementary Table S1**)—within the Protocol, users can directly embed relevant experimental data (e.g., procedural notes, critical result values, checkpoint confirmations) while executing each step of the experiment. Taken together, these entries form a single Airalogy Record, which, in practical terms, "lives" within the Protocol. By storing the experimental data in this integrated format, each Record remains tightly coupled to the Protocol that generated it (see **Supplementary Method S2.3**) for details on the data structure of an Airalogy Record). Consequently, researchers no longer have to manage protocols and experimental data as isolated files but can instead preserve the complete workflow and results in a unified, accessible format.

**Supplementary Fig. S19: Partial screenshots of the Airalogy Protocol used in this case study, illustrating different types of Airalogy Fields.**
**a,** The appearance of each Field before any experimental data are entered. This Protocol defines several Field types for recording distinct dimensions of experimental outcomes. For example, Step (①) denotes an experimental action step and provides an annotation box where users can document any deviations from the predefined procedure. Check (②) highlights critical steps requiring special attention. Var (③) captures key results that are recorded once the Protocol is executed. **b,** The same Protocol after experimental data have been filled into each Field.

**Supplementary Table S1: How different types of Airalogy Fields in a Protocol establish close data associations with Airalogy Records.**

| Field type | Role in research data recording |
|---|---|
| **Step** (the structural backbone) | **1. Flexible structure:** Steps outline the sequential instructions of an experimental Protocol and can be freely inserted, removed, or reordered as conditions change. **2. Transparent modification:** Any procedural modifications, such as adding an extra wash step upon impurity detection during protein purification, are explicitly recorded along with their rationales. **3. Comprehensive logs:** Step data serve as procedural logs, capturing all modifications or deviations from default steps, ensuring transparent tracking of Protocol evolution. |
| **Var** (capturing critical data) | **1. Multimodal data capture:** Vars represent data points collected during each experimental step, encompassing string/numeric values (e.g., researcher's name, pH) as well as data in other modalities (e.g., images or videos). **2. Adaptable variables:** Researchers can dynamically add new variables or remove less relevant ones as needed. **3. Detailed observations:** Var data capture all measurements and observations made throughout the experiment, ensuring that both positive and negative outcomes are archived for later analysis and informed decision-making. |
| **Check** (ensuring consistency and accuracy) | **1. Validation checkpoint:** Checks serve as validation checkpoints to confirm that crucial conditions are met before proceeding, such as verifying the operational status of critical equipment, thus minimizing errors and boosting reproducibility. **2. Adaptable standards:** Checks can be revised or expanded as insights emerge, ensuring continuous enhancement of experiment quality. **3. Clear documentation:** Check data typically indicate whether key conditions (e.g., instrument settings) passed or failed the predefined settings. |

By integrating these components, the Airalogy Protocol ensures transparency and traceability. Every modification is meticulously logged and all outcomes are preserved, enabling the feedback system for continuous optimization.

### S1.5.2 Driving Airalogy Protocol improvement through a feedback loop

An Airalogy Protocol and its corresponding Records form a closed-loop system: each Protocol run produces a Record that reflects the Protocol's clarity, completeness, and reliability. By using the quality and consistency of these Records as indirect metrics of Protocol performance, researchers can identify ambiguities or bottlenecks and systematically address them. Repeated cycles of review and optimization ultimately produce clearer instructions, fewer errors, and more reproducible results (**Fig. 6d**).

Below is a general framework for feedback-driven evolution of an Airalogy Protocol (**Supplementary Fig. S20**):

1. **Define the initial Airalogy Protocol:** Using expert knowledge, compile the Protocol's core information and incorporate the appropriate Airalogy Fields (e.g., Steps, Vars, and Checks). This creates an initial version of the Airalogy Protocol that can both guide the experiment and capture its key outputs.
2. **Collect Airalogy Records:** Each Protocol execution generates an Airalogy Record, capturing detailed information about the outcomes and effectiveness of the Protocol.
3. **Analyze and provide feedback:** Researchers worldwide can individually or collaboratively analyze the Records to identify issues and discuss improvements in the Discussions section on the Airalogy Platform, akin to GitHub's issue-tracking and pull-request workflows. In addition to human expertise, the AI research co-pilot, Aira, can help uncover deeper insights by spotlighting critical details.
4. **Adjust the Protocol:** Insights from the analysis guide modifications to Steps, Vars, and Checks, thus refining the Protocol and improving the quality of new Records.
5. **Refine iteratively:** The updated Protocol is used again, generating new Records that drive further improvements. This cycle allows the Protocol to evolve continuously, accommodating shifting experimental demands while ensuring reproducibility and efficiency.

Through these iterative optimizations, every aspect of experimental practice is systematically refined. Real-time feedback not only boosts the precision and adaptability of Protocols but ultimately transforms them into robust SOPs that fuel sustained scientific discovery.

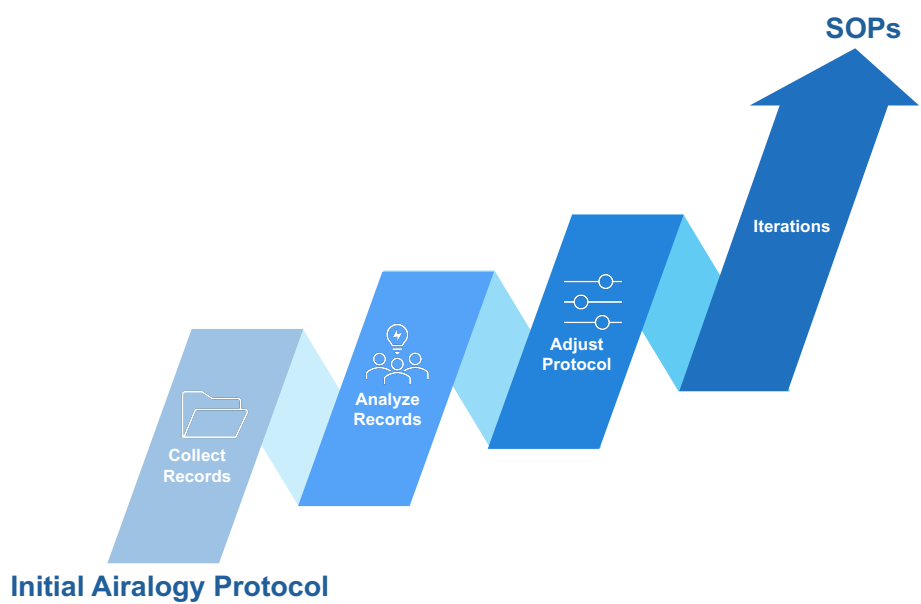

**Supplementary Fig. S20: A general framework for feedback-driven evolution of an Airalogy Protocol.**

### S1.5.3 A real-world example of an evolving Airalogy Protocol

In this case study, we applied the iterative optimization framework shown in **Fig. 6d** and **Supplementary Fig. S20** to refine an Airalogy Protocol for purifying lipoylated H protein[44]. Over six rounds of improvements (leading to six improved Protocol versions, **Supplementary Fig. S21a**), each new version was modified based on insights gleaned from the previous version's Airalogy Records (see **Supplementary Table S2**). **Supplementary Fig. S21b** compares key changes between the initial Protocol (V1) and the final Protocol (V7). By the sixth iteration, the Protocol effectively guided users to produce high-quality lipoylated H protein—essentially achieving SOP status for this experimental workflow.

Through these iterative refinements, the Protocol steadily matured until it provided clear guidance for every step, minimized experimental variability, and delivered reproducible results—underscoring the power of a feedback-driven approach for Protocol development.

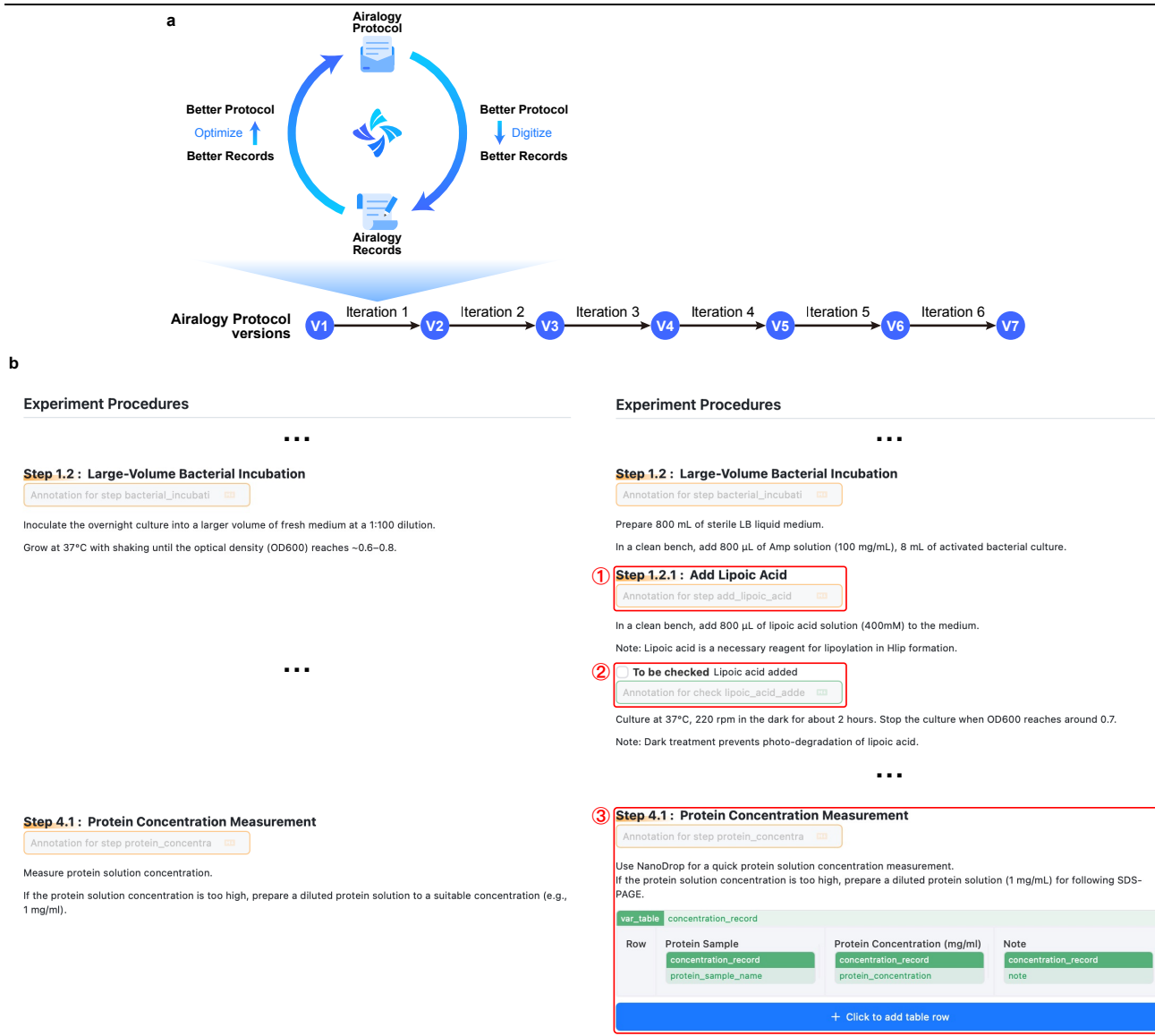

**Supplementary Fig. S21: Iterative optimization of an Airalogy Protocol for purifying lipoylated H protein.**

**a,** Starting from the initial protocol (V1), six iterative refinements resulted in a final version (V7) capable of reliably producing high-purity lipoylated H protein. **b,** Comparative overview of V1 (initial, left) and V7 (optimized, right), highlighting some critical improvements: ① Inserting a dedicated step for adding lipoic acid; ② Adding a validation check to ensure successful lipoic acid addition; ③ Establishing a variables table for systematic documentation of results.

**Supplementary Table S2: Key optimization points introduced in each iteration of the lipoylated H protein purification Protocol, along with the data-driven rationale for these changes.**

The "Records" column describes the problematic data observed in the corresponding Airalogy Records. The "Analysis" column outlines the conclusions drawn by human researchers and AI. The "Optimization" column summarizes the specific Protocol refinements implemented at each stage.

| Iteration | Records | Analysis | Optimization |
|---|---|---|---|
| 1 | Both lipoylated and un-lipoylated H protein detected on the gel | Low lipoylation efficiency | Insert Step: Co-express LplA and H protein; increase lipoic acid concentration during induction |
| | | ...... | |
| 3 | Temperature rose during cell homogenization | Cooling system malfunction | Add Check: Verify homogenizer cooling system is active and within temperature limits |
| | | ...... | |
| 6 | High-purity lipoylated H protein obtained | Protocol stable and reproducible | No further adjustments needed; Protocol finalized as a robust SOP |

## S1.5.4 Discussions

By linking Airalogy Protocols with detailed Records and a comprehensive feedback mechanism, this case study illustrates a novel framework for evolving a Protocol into a SOP. Moreover, because Airalogy Protocols are easily reused and adapted across different laboratories, projects, and researchers, a single initial Protocol can follow multiple optimization paths to meet diverse experimental needs (**Supplementary Fig. S22**). As Protocols are continually refined by a growing global user community, they become increasingly robust—ultimately enhancing the standardization, reproducibility, and success rate of research or experiments defined through Airalogy Protocols.

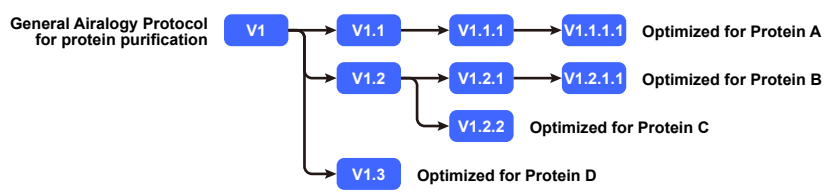

**Supplementary Fig. S22: An Airalogy Protocol evolution tree.**

A single initial Protocol can undergo iterative optimization and branching to meet varied research needs. For instance, consider a hypothetical generic protein purification Protocol (V1). This initial version can be optimized and branched for distinct protein targets—such as V1.1.1.1 for Protein A and V1.2.1.1 for Protein B. At each node, the Protocol can split again (for example, V1 might spawn V1.1, V1.2, and V1.3, which themselves could branch into V1.2.1, V1.2.2, and so on)—forming a continually expanding tree of iterative improvements.

While the Airalogy Protocol has demonstrated substantial benefits for the scientific community, its core principles extend well beyond experimental research. Fundamentally, the Protocol models physical interactions with the real world, and the Records document the outcomes of these interactions. This feedback-driven system of iterative optimization is broadly applicable to any domain requiring systematic processes and continuous improvement, such as healthcare and education.

In the healthcare domain, Airalogy's principle could revolutionize clinical procedures by continually adapting treatment procedures based on detailed patient outcomes, integrating personalized practice and medicine insights. Within education, iterative Protocol refinement could lead to tailored learning experience, with curriculum adjustments driven by direct, measurable feedback from student's scores.

Extending further, industry may experience greater potential benefit. For example, manufacturing processes can profit from treating workflows as adaptive Protocols. Assembly steps, quality-control procedures, and machine calibrations can be recorded in detailed Records, capturing all variations and outcomes. These data can then be analyzed to uncover inefficiencies, reduce waste, and optimize production variables. For instance, monitoring a machine's operational parameters (Vars) and incorporating verification points (Checks) to confirm calibration accuracy enhance productivity and reduce downtime. Iterative refinements ensure that production systems respond dynamically to real-time data, delivering consistent quality and improved resource utilization.

In summary, the Airalogy Protocol paradigm not only revolutionizes experimental approaches but also serves as a versatile framework with potential applications in industry and beyond. By transforming static procedures into living, self-improving systems, it offers a powerful route to ongoing progress and efficiency in an ever-changing world.

## S1.6 Supplementary Case Study S6: Airalogy Platform as a one-stop solution for research data recording, management, and utilization

Contemporary scientific research is entering the big data era. The extensive use of automated, high-throughput equipment has led to an exponential growth in data volume[76], rendering datasets increasingly complex, multimodal, and unstructured. Consequently, modern research is more data-intensive[77], posing unprecedented challenges. First, because experimental data often originate from multiple sources, they can become dispersed across various instruments and researchers, posing significant hurdles for data asset management. Second, the heightened complexity of scientific data complicates cross-instrument and cross-researcher integration, as records from diverse instruments and researchers must undergo standardization before they can be effectively combined. Structured, standardized data recording is therefore indispensable for preserving experimental resources and ensuring reproducibility and transparency. Third, extracting deep insights from these multimodal datasets demands advanced data-processing skills, requiring researchers to invest considerable time and effort in mastering statistical, analytical, and visualization techniques in addition to their domain expertise. Here, we use the development of industrial strains in bioengineering as a case study to show how the Airalogy Platform addresses these challenges by providing a one-stop solution for standardized data recording, management, integration, analysis, visualization, and report generation—ultimately empowering and accelerating scientific research.

### S1.6.1 Challenges in industrial strain research and development

Industrial strain research and development requires knowledge and skills spanning biology, chemistry, physics, engineering, and computer science to design strains suitable for industrial production. In our case study, we employed a DASGIP parallel bioreactor system to develop a strain aimed at improving both the yield and economic viability of a target product[45,78]. Traditionally, such research involves four major steps (**Supplementary Fig. S23**, gray pipeline):

1. Constructing the engineered strain;
2. Performing fermentations of this strain using the DASGIP system;
3. Analyzing tail gas and fermentation broth samples at multiple time points via mass spectrometry, liquid chromatography, and biosensors, etc.;
4. Integrating data from various instruments and summarizing the results into a test report, which then guides further strain and process optimization.

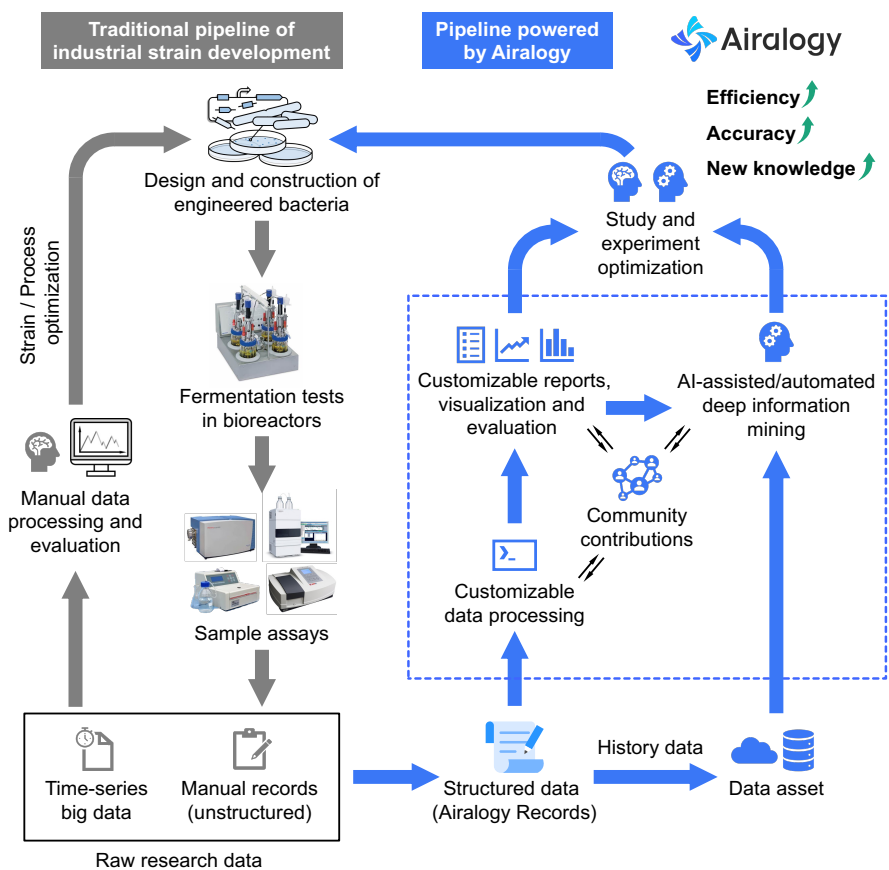

**Supplementary Fig. S23: Comparison of the traditional industrial strain development pipeline (gray pipeline) with an Airalogy-empowered pipeline (blue pipeline).**
The traditional workflow comprises strain construction, fermentation testing, sample analysis, manual data processing, and strain/process optimization. In contrast, the Airalogy-empowered approach integrates automated data processing, visualization, reporting, and evaluation through community-driven methods, culminating in AI-assisted/automated data mining that guides subsequent experiments.

These steps form a closed loop that is repeated until the strain meets industrial requirements. However, data integration and analysis often become the bottleneck in this process, primarily for two reasons:

1. **Large and heterogeneous datasets:** The DASGIP parallel bioreactor system can simultaneously conduct multiple small-scale fermentations while measuring parameters (e.g., pH, dissolved oxygen, temperature) at second-to-minute resolution. Fermentations may last from days to weeks, generating massive time-series datasets (potentially encompassing over 50 process parameters and more than 10,000 time points in our case). Real-time mass spectrometry further analyzes off-gas components, and various other instruments measure attributes like cell density, residual sugar, and product or by-product yields at multiple stages. Certain metrics also require manual sampling at specific times (e.g., product concentration, sugar content), which must be integrated to build a complete time profile of the fermentation. Because these measurements are gathered on different schedules and from various sources, time-alignment and analysis of these large, multimodal, and unstructured datasets demand extensive effort and advanced data-analysis skills.

2. **Difficulty in extracting deep insights:** This challenge arises at two levels. First, within a single fermentation batch, one must interpret the influence of numerous variables (potentially up to 50 process parameters) on the outcome. Second, across multiple batches (often performed under different conditions to optimize an industrial strain) uncovering patterns, correlations, and trends, and relating them to underlying biological mechanisms, is critical for guiding further strain engineering. Traditional analytical methods often struggle to reveal such nuanced information, necessitating more sophisticated techniques like AI/machine learning[79], which demand high computational capacity and specialized data science expertise.

Notably, challenges of this sort are not unique to industrial strain research; they also arise in many other fields, such as drug screening[80], clinical diagnostics[81], omics studies[82], materials synthesis[83]. Consequently, there is an urgent need for platforms that facilitate collaboration between scientists and data experts by providing automated data-analysis pipelines and advanced tools (such as AI-integration capabilities) for in-depth mining of complex experimental datasets, ultimately accelerating scientific research.

### S1.6.2 Using the Airalogy Platform for research data recording, management, and utilization

The Airalogy Platform, driven by both AI and community, offers an ideal environment for recording, managing, and utilizing research data. In our case study, we used the Airalogy Protocol to build a one-stop solution that handles standardized data collection, integration, analysis, visualization, and report generation (**Supplementary Fig. S23**, blue pipeline).

Structured, standardized data form the foundation of automated processing. Accordingly, we designed a dedicated Airalogy Protocol to electronically record all raw data in a consistent format (**Supplementary Fig. S24**). Despite of basic fermentation information (**Supplementary Fig. S24b**), our raw experimental data encompassed three main sources (**Supplementary Fig. S24a**):

1. Fermentation process monitoring data from the DASGIP parallel bioreactor system (CSV files);
2. Real-time off-gas composition data from mass spectrometry (CSV files);
3. Manually collected measurements (e.g., cell density, sugar concentration, and product yield) at various time points.

In the Airalogy Protocol, the first two data types were uploaded into corresponding Airalogy Fields which allow users to upload CSV files (**Supplementary Fig. S24c ①**), while the manually sampled data were entered into a table-type Variable Field (**Supplementary Fig. S24c ②**). All of these datasets represent time-series measurements which, in our prior traditional pipeline, required labor-intensive manual alignment. By contrast, the Airalogy Protocol lets users embed custom Python scripts and automatically assign data across different Fields via the "Assigner" feature (**Supplementary Method S2.2.4**). We therefore developed a Python function to automatically align these three time-series data sources (DASGIP data, mass spectrometry data, and manually sampled data) and integrated them in the Protocol through an Assigner. Consequently, once these data are uploaded/entered to the relevant Airalogy Fields, the Protocol automatically merges them into a single CSV file, stored in another Airalogy Field (**Supplementary Fig. S24d ③**). Additionally, the Protocol uses embedded user-defined analysis scripts to automatically visualize and plot key fermentation parameters (e.g., time-course growth curves, temperature changes; **Supplementary Fig. S24d ④**) and computes essential evaluation metrics (e.g., product yield and carbon balance; **Supplementary Fig. S24d ⑤**). Recognizing the complexity of handling 50+ parameters in our study, we further integrated a specialized LLM-based AI agent via another Assigner, enabling automated interpretation of the fermentation outcomes (**Supplementary Fig. S24d ⑥**).

Compared with traditional manual data processing, this Airalogy-based automated pipeline guarantees consistent analysis across different batches (enhancing reliability) and reduces processing time from hours or days to mere minutes (saving substantial research effort). Moreover, by allowing researchers to embed LLM-based AI into Protocols, data interpretation can shift from purely manual work to AI-assisted or automated analysis—an evolution that aligns with the growing prominence of AI in future scientific research[13].

**a**

**Raw research data**

Basic experimental information

Data from DASGIP

Data from mass spectrometer

Manual records

Align and merge according to timeline

Integrated time series data file

Fermentation evaluation parameters

Customized data visualization

Summary and insights based on LLM

**b** Fermentation Data Recording

▶ About this research protocol

**Basic Information**

Recorder: Jingke `var record_recorder`

Time: 2024-01-01 00:00:00 `var record_time`

**Basic Information of Fermentation Batch**

Batch number: 1 `var record_batch_number`

Fermenter number: 1 `var record_bioreactor_number`

Strain number: S001 `var record_strain_number`

Medium components: basic medium `var record_medium_component`

The initial glucose concentration in the medium: 8 `var record_conc_medium_glucose` g/L

Mass fraction of the fed glucose: 523.1 `var record_conc_feed_glucose` g/L

**c** Online Data Uploading

① Upload DASGIP Data (.csv Format): `var dasgip_file`

Upload Mass Data (.csv Format): `var mass_file`

Sampling and Offline Data Recording

② `var_table sample_record`

| Sample time `sample_record` `sample_timestamp` | Cumulative time (h) `sample _record _time` | OD600 value `sample _record _sample _bacterial _density` | Residual glucose conc. (g/L) `sample _record sample_conc _glucose` | Product A conc. (g/L) `sample_record sample_conc_product_a` | Product B conc. (g/L) `sample_record sample_conc_product_b` | Product C conc. (g/L) `sample_record sample_conc_product_c` |
|---|---|---|---|---|---|---|
| 2024-01-01 | 2 − + | 2.22 − + | 8.5 − + | Product A conc. (g/L − + | Product B conc. (g/L − + | Product C conc. (g/L − + |
| 2024-01-01 | 4 − + | 3.94 − + | 6.8 − + | 43.25 − + | 0.48 − + | Product C conc. (g/L − + |
| 2024-01-01 | 6 − + | 7.45 − + | 3.6 − + | Product A conc. (g/L − + | Product B conc. (g/L − + | Product C conc. (g/L − + |

**d** Experimental Data Integration, Processing, and Report Generation

③ Multi-source Data Integration

The three types of experimental data (sampling offline data, DASGIP data, and tail gas spectrometry data) are automatically combined into a CSV file.

`var output_file`

④ Fermentation process control

Growth curve: 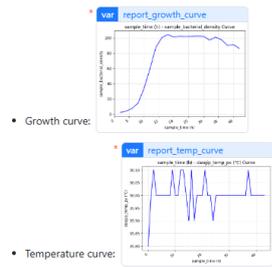 `var report_growth_curve`

Temperature curve: `var report_temp_curve`

⑥ Report generation

Report generated by LLM

`var report`

Markdown ⟲ ↵ ⤵ ⊞ + H ↹ T₉ ⋯

**Evaluation of Experimental Results**

**1. Real-time Control and Sensor Data Analysis**

**Temperature Curve:**

- The temperature shows significant fluctuations, especially in the initial stages (0-10 hours) and around 20 hours. These fluctuations could stress the cells and affect their metabolic activities.

- **Recommendation:** Stabilize the temperature control to minimize fluctuations. A more robust temperature control system or better insulation might be necessary.

**pH Curve:**

- The pH drops sharply in the first few hours and then stabilizes around 6.6. A lower pH can inhibit cell growth and product formation.

- **Recommendation:** Optimize the pH control strategy, possibly by adjusting the buffer capacity of the medium or the timing and rate of base addition.

Data analysis

⑤

| Test Reasults | | Units |
|---|---|---|
| Fermentation period | `var g1_batch_total_time` 42 − + | h |
| Titer A | `var g1_final_titer_a` 43 − + | g/L |
| Final bacterial density | `var g1_final_bacterial_density` 86 − + | - |
| Total glucose consumption | `var g2_total_glucose_consum` 251.68 − + | g |
| Conversion rate | `var g3_conversion_rate` 13.16 − + | % |
| Carbon recovery | `var g4_carbon_recovery` 86.75 − + | % |

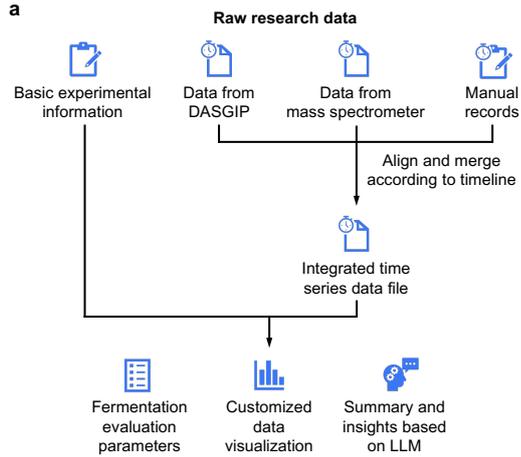

**Supplementary Fig. S24: The Airalogy Protocol for the fermentation experiment.**
**a,** Schematic overview of the Protocol's core functions. This Protocol allows users to record basic fermentation information and parameters, as well as data from three different sources (DASGIP, mass spectrometry, and manual sampling). It automatically aligns these data by time, compiles them into a single CSV file containing all fermentation parameters, and then generates visualizations and key performance metrics. Additionally, a specialized LLM can be embedded to produce an automated summary and insights for each fermentation run. **b,** Screenshot of the Protocol interface for recording the experiment's basic information and parameters. **c,** Screenshot of the interface for uploading or entering data from the three sources. CSV files from the DASGIP system and mass spectrometer are uploaded to variable-type (`var`) Airalogy Fields (`dasgip_file`, `mass_file`; ①), and manual sampling data are entered in a table-type (`var_table`) field (`sample_record`; ②). **d,** Screenshot of the automatic data-integration, visualization, and insight-generation interface. After the three raw datasets have been uploaded/entered, the Protocol merges them into a single CSV file (`output_file`; ③) and then uses the combined dataset to generate plots and visualizations (④) and to calculate key performance metrics (⑤). Researchers can also embed an LLM (⑥) for automated data interpretation.

The above discussion illustrates Airalogy's automated data analysis for individual fermentation batches. On the Airalogy Platform, all raw data and analysis outcomes are stored in structured, standardized Airalogy Records, preserving valuable experimental resources for long-term research and centralizing data asset management. Furthermore, this standardized Record structure ensures that both humans and AI can readily interpret the data, paving the way for more advanced applications. For instance, after multiple fermentation batches are completed, researchers can use Aira, Airalogy's AI research copilot, to conduct conversational data analysis across these history Records (e.g., comparing results, identifying patterns, discovering new insights, and flagging anomalies; **Supplementary Fig. S25**). This kind of knowledge mining would be extremely challenging under traditional manual pipelines. In addition, Airalogy's collaborative design allows all project members to access and reuse these data for custom processing, analysis, and visualization. Any scripts within a Protocol used for data analysis can also be shared and maintained by the community, promoting ongoing improvements and expanded functionality for each Protocol.

Please compare the diferences between the results of the two experiments and give a brief summary.

The two experiments conducted under the same protocol involved different batches (Batch 1 and Batch 2) with similar conditions but yielded distinct results. Here is a brief comparison:

Key Differences:

1. **Final Titer of Product A:**
   - **Batch 1:** Reached a final titer of **43.3 g/L.**
   - **Batch 2:** Achieved a higher final titer of **44.17 g/L.**

2. **Glucose Consumption:**
   - **Batch 1:** Consumed **251.92 g** of glucose.
   - **Batch 2:** Consumed **189.43 g** of glucose, indicating lower glucose utilization in Batch 2.

3. **Conversion Rate:**
   - **Batch 1:** Had a conversion rate of **13.23%.**
   - **Batch 2:** Showed a better conversion rate of **17.95%**, suggesting more efficient substrate utilization.

4. **Carbon Recovery:**
   - **Batch 1:** Recorded a carbon recovery of **86.81%.**
   - **Batch 2:** Achieved a significantly higher carbon recovery of **116.06%**, which might indicate some discrepancies or errors in data collection or calculation.

5. **Bacterial Density:**
   - **Batch 1:** Final bacterial density was **86 OD600**.
   - **Batch 2:** Final bacterial density was **73 OD600**, slightly lower than Batch 1.

6. **Product B Concentration:**
   - **Batch 1:** Product B concentration reached up to **0.45 g/L.**
   - **Batch 2:** Product B concentration was consistently lower, reaching up to **0.6 g/L** at certain time points.

Summary:

Batch 2 showed a better conversion rate and higher carbon recovery, which could suggest improved metabolic efficiency. However, the lower glucose consumption and slightly reduced bacterial density may imply differences in strain behavior or environmental factors. The higher final titer of product A in Batch 2 indicates better productivity. Overall, Batch 2 appears to be more efficient in terms of conversion and product yield, but further investigation into the discrepancies in carbon recovery and glucose utilization is recommended.

**Supplementary Fig. S25: AI-driven conversational analysis of historical data from multiple fermentation batches.**

### S1.6.3 Broad potential for automated research data processing and analysis with Airalogy

In summary, our case study on industrial strain development demonstrates how Airalogy Platform can serve as a one-stop solution for research data recording, management and utilization, encompassing: (1) standardized, electronic data recording; (2) automated data processing, analysis, and visualization through flexible custom scripts; (3) AI-assisted deep data mining; (4) secure archiving, sharing, and reuse of data assets; and (5) community-driven maintenance of research pipelines. Compared with our traditional pipelines, the Airalogy-based pipeline not only boosts efficiency and accuracy, and fosters new knowledge discovery (**Supplementary Table S3**), but also introduces a new paradigm of collaborative research (**Case Study 6**).

First, workflows for data recording, analysis, and visualization can be modularized and reused, lowering barriers to collaboration across researchers, labs, and disciplines. In our case, a data analysis expert implemented the automated data-processing scripts, while researchers with a bioengineering background could use these scripts without additional programming skills. Second, by designing Protocols in a modular fashion, data processing and downstream applications remain maximally flexible. Although for simplicity we consolidated data integration, analysis, visualization, and report generation into a single Airalogy Protocol in this case study, these functions can readily be split into several more specialized Protocols, each maintained by users with different expertise.

Through the Airalogy Protocol Pipeline paradigm (**Fig. 3**), the same experimental Airalogy Records can be analyzed by multiple Protocols defined by different researchers, and data outputs from one Protocol can be fed into other Protocols for further visualization or custom report generation (**Supplementary Fig. S26**). Collectively, this paradigm effectively promotes collaboration among experts with diverse skillsets, as well as collaboration between humans and AI, and across multiple laboratories and disciplines.

**Supplementary Table S3: Comparison between the traditional pipeline and the Airalogy-based pipeline for industrial strain research and development.**

| | Traditional pipeline | Airalogy-based pipeline |
|---|---|---|
| **Data recording & management** | Recorded separately in various files, devices, and user notes; data are scattered, difficult to manage, and easily lost | All data are uniformly recorded and uploaded to standardized Airalogy Protocols, then stored as structured, standardized Airalogy Records for centralized management |
| **Data analysis method** | Manual processing using common software (e.g., Excel) | Automated processing via Assigners in Airalogy Protocols |
| **Data analysis efficiency** | Time- and labor-intensive (single-batch analysis can take hours to days) | Fast and automated (single-batch analysis can be completed in minutes) |
| **Accuracy & consistency** | Errors are common among new personnel; even experienced researchers can make mistakes due to large data volumes; analysis details may vary by researcher | A unified, predefined Assigner logic ensures consistent, high-accuracy data processing, irrespective of user expertise |
| **AI integration** | Difficult to implement | Straightforward: users can embed AI/LLM-based data-analysis methods for AI-assisted/automated processing, insights, and knowledge discovery |
| **Data & analysis sharing** | Low shareability | High shareability: all Protocols and Records on the Airalogy Platform can be easily shared with the community and reused with one click |
| **Community involvement** | Minimal community support | All Protocols can be published to the Airalogy community, enabling continuous improvement and optimization |

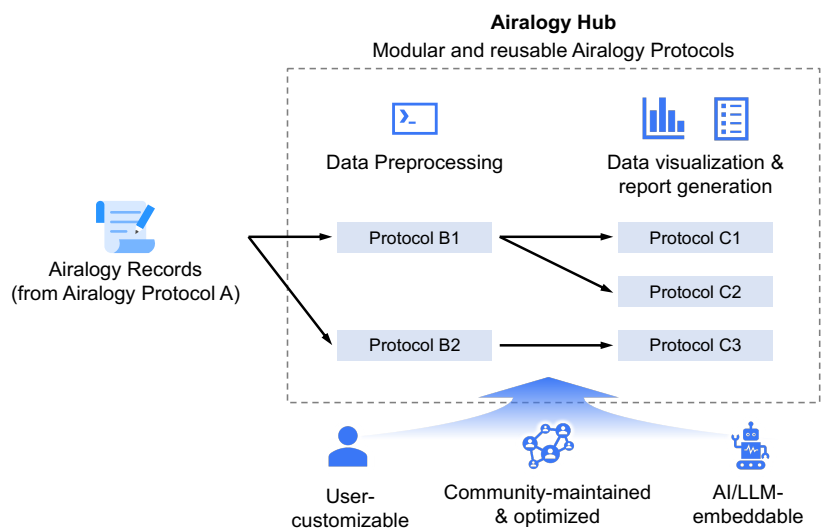

**Supplementary Fig. S26: The modularity and reusability of Airalogy Protocols provide researchers with considerable flexibility in data handling and usage.**

Under the Airalogy Protocol Pipeline architecture (**Fig. 3**), various analysis steps can be decomposed into specialized Protocols. For example, data analysis typically involves two major steps: (1) preprocessing data into a specific format for subsequent use, and (2) visualizing results and generating reports. Different users with distinct skillsets can each create and maintain Protocols for these tasks (embedding custom data-analysis logic, or even including AI agents/LLMs) to automate each step. These Protocols can then be shared on the Airalogy Hub for reuse or collaborative refinement. As the Hub expands, users can freely combine Protocols without needing programming or data-analysis expertise to generate high-quality visualizations or reports. For instance, a standardized Airalogy Record from Protocol A can be preprocessed by Protocol B1, followed by Protocol C1/C2 for visualization and reporting. Alternatively, Protocol B2 can preprocess the same data in a different way before passing it to Protocol C3 for another custom report.

## S1.7 Supplementary Case Study S7: Fostering collaboration and division of labor in research through Airalogy

Throughout human history, major leaps in productivity have been accompanied by transformative changes in organizational principles and structures. The increasing specialization of labor and the expansion of collaborative networks have consistently served as core drivers of efficiency. Adam Smith first proposed and systematically described the concept of "division of labor" in 1776[46], while Friedrich Hayek introduced and elaborated upon "division of knowledge" in 1937[84]. These seminal ideas laid the conceptual groundwork for modern social division of labor and have profoundly shaped human progress, leading to substantial gains in productivity.

Research is the engine of technological innovation and advancement, as well as a catalyst for leveraging science to drive societal progress. Today, research itself is increasingly evolving into a complex, socially driven undertaking. In this chapter, we build on the established concepts of labor and knowledge division to discuss the notion of "division of research". We also explore how Airalogy-mediated research collaboration and division of research may shape future paradigms of scientific work, potentially accelerating the pace of innovation and exerting far-reaching impacts on scientific productivity.

### S1.7.1 Breaking down large tasks into modular subtasks and being able to coordinate them in an orderly manner is the key to effective division of labor

In 1913, the world's first moving assembly line debuted at Ford[85], spearheading a transformation in industrial productivity and production methods, while laying the foundation for industrial standardization and global manufacturing. Several key factors underpinned the success of Ford's assembly line:

1. **Specialized division of labor and process standardization:** Instead of requiring workers to master multiple skills for full vehicle assembly, Ford broke down the process into hundreds of small, standardized steps. Each worker repeatedly performed a single, specialized task, thereby improving proficiency and product quality simultaneously.
2. **Efficient integration of production resources:** By using conveyor belts and moving assembly lines, Ford organized human labor and raw materials optimally—delivering parts directly to workers, who remained at fixed stations. This minimized time wasted in fetching parts and significantly increased production efficiency.

Today, most researchers still operate under a traditional research paradigm. Pursuing a scientific goal often forces researchers to acquire and execute every necessary experimental technique themselves, investing substantial time and energy in learning skills, conducting repetitive lab work, and performing data analysis. For example, consider a traditional genetics lab: A Ph.D. student investigating a genetics problem may need to master dozens of experimental methods—from constructing DNA fragments to generating transgenic mice—each of which must be performed nearly perfectly to avoid erroneous results that can delay the project. As a result, it can take years before a student becomes competent enough to conduct research independently, posing a major barrier to efficient, high-quality scientific output. In the context of an exponentially growing body of knowledge, this traditional approach is increasingly inadequate.

Much like society's quest to enhance productivity through better organizational structures, modern research faces the challenge of optimizing both its methodology and organizational forms to achieve a leap in efficacy.

Accompanying the rapid development of AI, which increasingly empowers all sectors of society, the Airalogy Platform has emerged as a community- and AI-driven solution for advancing division of research. This new research paradigm employs the following distinctive strategies:

1. **Modularizing research and coordinating modules:** On the Airalogy Platform, large-scale research projects can be systematically decomposed into standardized Airalogy Protocols, each representing a self-contained modular unit (**Fig. 6f**). Through Airalogy Protocol Pipelines (**Fig. 3**) or Airalogy Protocol Workflows (**Fig. 5**), multiple Protocols can be assembled into a cohesive, collaborative framework. On one hand, this allows complex single- or multi-disciplinary projects to be broken down into reusable, "assembly line" components. On the other hand, each Protocol can be contributed by researchers with specialized expertise, yet they can seamlessly be combined and applied together—so that users of a given Pipeline or Workflow effectively benefit from the collective intelligence of multiple experts.

2. **Using an evolutionary mechanism to move toward standardization:** As a community-driven Platform, each Airalogy Protocol can be continually refined by domain experts and end-users, ultimately converging into a reliable, standardized operating procedure (SOP) (**Supplementary Case Study S5** illustrates how repeated use and iterative improvements transform Airalogy Protocols into SOPs). Traditionally, researchers rely on an apprenticeship model to pass down "know-how" or hidden knowledge—hard to articulate or kept private. However, because Airalogy Protocols evolve through repeated application, they progressively incorporate these elusive insights. More importantly, once such knowledge is embedded in a Protocol, it becomes easily shareable and reusable across the global research community. In this way, Airalogy helps convert formerly tacit knowledge[86] into explicit and reusable best practices. From the researcher's perspective, adopting an SOP drastically reduces the learning curve for experimental skills.

3. **Research encapsulation:** Another key concept of Airalogy Protocols is that the Protocol's designer and the end user are effectively decoupled (**Fig. 2**). This means that users only need to select and assemble the research units relevant to their goals without fully understanding each Protocol's internal principles or mechanisms—an approach analogous to the "encapsulation" concept in the Python programming language. For instance, when users call `numpy.add(a, b)`, they rely on internal implementation of Numpy[87] (a Python package) without needing to know how the addition is computed at the lowest level. In a similar way, Airalogy Protocols can be treated as encapsulated modules. Researchers focus on the functional outcome rather than the implementation details, lowering the learning curve. In addition, SOP-oriented Protocols strengthen users' confidence that each Protocol will perform as intended, allowing them to concentrate on higher-level research problems solvable by combining multiple Protocols. From the perspective of other participants in the research ecosystem, standardized, encapsulated Protocols reduce the barriers to entry in scientific work, opening opportunities for individuals, organizations, and even AIs worldwide to participate and contribute—an emerging trend referred to as "Research as a Service" (discuss below). This shift will likely spur the development of specialized automation equipment, professional service providers, and commercial outsourcing options, creating broader employment prospects in research-related fields.

4. **Deep integration of AI and automation:** Within each Airalogy Protocol, researchers can readily embed AI-driven or automated methods, transforming the Protocol into an application designed to solve a specific research task automatically (**Supplementary Case Study S4** illustrates several examples). Furthermore, any set of Protocols can be assembled into an Airalogy Protocol Workflow, which can then be executed under AI control, achieving fully automated research (**Supplementary Case Study S1** demonstrates the specifics). Under this paradigm, each Protocol or Workflow not only supports human-driven operations but also enables an ever-increasing share of scientific tasks to be performed or assisted by AI—freeing human researchers to focus on higher-level challenges. This synergy of AI-powered automation and human expertise aligns with today's broader wave of AI-driven research and holds great promise for disruptive scientific innovation driven by AI–human collaboration.

In summary, by redefining and modularizing standardized research tasks, the Airalogy Platform fundamentally improves how research personnel and resources are allocated, thereby boosting the overall efficiency of scientific endeavors.

### S1.7.2 Envisioning how Airalogy's new paradigm of research collaboration and division of labor will shape the future

### S1.7.2.1 Research as a Service (RaaS)

Under the ongoing wave of globalization, enabling researchers from diverse regions to collaborate on shared research goals promises to accelerate innovation on a worldwide scale[88]. By encapsulating specific scientific activities within Airalogy Protocols, the Airalogy Platform is poised to reshape global research collaboration around the concept of "Research as a Service" (RaaS). This model places Airalogy Protocols at the core of an entirely new research value chain, in which complex processes are decomposed into standardized, modular Protocol-based units that can be executed by individuals or institutions with specialized expertise.

For instance, tasks such as experimental execution, data collection, and data analysis may each be handled by distinct service providers on the Platform, using Airalogy Protocols as standardized solutions. Researchers can simply pick and combine these "research micro services" in a building-block fashion, both lowering the barriers to entry and freeing them to focus on innovation rather than repetitive tasks.

Additionally, every service outcome based on any Airalogy Protocol is accompanied by a corresponding Airalogy Record, which can then be distributed or shared globally via the Airalogy Platform. Serving as the hub for these Protocol-derived datasets, the Platform can integrate efforts across different disciplines, institutions, laboratories, and researchers worldwide. In this way, the Airalogy Platform may evolve into a global data nexus and a marketplace for research outputs.

Looking ahead, the Airalogy Platform will further expand its openness, attracting universities, corporations, independent scientists, and even AI agents to form a multi-stakeholder collaboration network. Equipment manufacturers might offer remote experimental interfaces, while data analysis experts provide data-based insights. Standardized workflows, resource aggregation, and easy data sharing will significantly reduce transaction costs for specialized services. Moreover, to maintain a vibrant ecosystem, the Airalogy Platform will explore incentive mechanisms tailored to the scientific community. These may include Protocol-usage-based automated trading systems, reputation systems to reward and encourage Protocol optimization and knowledge sharing, and cross-team collaboration agreements that balance individual innovation with collective wisdom. AI agents could match supply and demand in real time, dynamically adjusting resource allocations and making the Platform a central "research productivity network".

In this ecosystem, research will shift from isolated competition toward open co-creation, dramatically enhancing the flow, reusability, and efficient allocation of global intellectual resources and innovation drivers—ultimately providing sustainable momentum for expanding the boundaries of human knowledge.

## S1.7.2.2 Data openness as publication (DOAP) and "data social media"

We are witnessing an explosion in scientific literature. For example, the top AI conference NeurIPS received nearly 30,000 submissions in 2025—about ten times more than a decade ago. On one hand, this overwhelming volume burdens peer reviewers, making it challenging to identify truly high-quality papers among a flood of submissions. On the other hand, readers face an insurmountable task: no single individual can absorb such vast amounts of literature, creating a situation akin to "so much to read, so little time". Consequently, we should reconsider the essence of academic publication—namely, the sharing and dissemination of knowledge.

The traditional journal-based, peer-reviewed publishing model is analogous to classic print media. Anyone can submit a manuscript, but an editorial board decides which work gets published, and only after publication can it be evaluated by the broader community. However, the rise of preprint servers such as arXiv and bioRxiv has introduced a new model akin to "research social media". Many top-tier papers (e.g., GPT-4[29] and DeepSeek-R1[31] technical reports) are simply posted on arXiv. Despite lacking formal peer review, these works have gained widespread attention and numerous citations because of their inherent quality. This dynamic resembles modern social media: content creators strive for engagement in the form of fans, likes, and shares (which in turn affect their visibility and recognition), incentivizing them to produce high-quality content. The platform itself plays a minimal gatekeeping role; instead, it leaves quality judgments to the community.

Despite these advantages, the current preprint-driven paradigm still falls short in two aspects. First, authors typically invest significant effort in structuring experimental data and summarizing conclusions before posting on a preprint server—making it likely that they emphasize only "positive" results. Even if unintentional, this practice skews reported findings toward success stories and may overlook negative or neutral results. Second, the time spent processing data delays the public release of first-hand research and conclusions, hampering real-time knowledge sharing at a global scale.

Here, we envision a new approach to knowledge dissemination, built on the Airalogy Platform, which addresses both of these limitations:

Leveraging the Airalogy Platform, every research activity can be encapsulated by an Airalogy Protocol, while each application of a Protocol yields an Airalogy Record that captures the resulting data. Crucially, each Record is assigned a globally unique ID and timestamp, ensuring traceability of every data point. In the future, if users conduct research on the Airalogy Platform and obtain critical results, they can immediately share the corresponding Protocols and Records. Thanks to unique IDs and timestamps, any claims regarding new discoveries can be substantiated with original, verifiable data.

Moreover, as AI and LLMs continue to advance—already capable of automatically generating research reports (e.g., ChatGPT, Gemini, Grok's "Deep Research" or "DeepSearch" features) or papers[16,17]—we envision a scenario where simply publishing Records on the Airalogy Platform can trigger AI-based analysis, producing real-time summaries or paper manuscripts. This accelerates the dissemination of both raw data and the insights they contain.

Unlike traditional articles where a batch of data typically supports only the single published study, data uploaded to Airalogy Platform can be accessed by a global community of researchers and AI agents, facilitating reanalysis from diverse perspectives. This community- and AI-driven approach may unlock all the underlying insights within a given dataset, whether the results are "positive" or "negative". The Platform also supports the free combination and analysis of data from multiple sources, laboratories, researchers, and disciplines—maximizing the broader value of such datasets.

Beyond sharing Protocols and Records once a study concludes, the Airalogy Platform could also foster a real-time, globally collaborative research model. A researcher can open a public project where all Protocols and Records are openly shared, thereby "live-streaming" each experimental step and result. As a study progresses, the latest methods and findings are publicly available, inviting community participation and real-time feedback from a worldwide audience.

### S1.7.2.3 Delegating all research to AI?

If, in the future, various research endeavors could be fully delegated to AI, then AI would effectively be capable of conducting science entirely on its own; given that AI never tires or stops, this scenario could trigger an explosive growth in knowledge[13].

Although some attempts have been made to automate research end-to-end—from generating ideas and designing experiments to analyzing data and writing papers[16,17]—these efforts have mostly remained within the realm of computer science. We are still far from achieving full-scale, AI-driven automation in natural sciences, especially in domains requiring real-world experimentation. Why is that?

In many ways, today's AI (particularly LLMs) exists in a purely digital domain and lacks the ability to interact physically with the real world. For the natural sciences, while an AI might propose novel research ideas and experimental protocols, it often lacks the means to actually implement them[51]. This shortfall manifests in two ways:

1. **Embodied intelligence lags behind language intelligence:** For instance, although AI could propose hundreds of imaginative methods for genetic editing in mice, it generally cannot carry them out. Currently, no available robot or embodied intelligence system allows AI to skillfully handle a live mouse and perform genetic manipulations.
2. **Practical limitations of resources and experimental conditions:** Whether a study is feasible depends heavily on access to the facilities and resources needed to execute the proposed experiments. AI systems like ChatGPT, which primarily operate via dialogue, lack an understanding of real-world constraints. Building a well-equipped lab, for instance, can require millions of dollars, and even then, its capabilities remain finite. For instance, if a lab does not have access to a scanning electron microscope—nor the means to borrow or use one from elsewhere—it would be unable to conduct the carbon nanotube dispersion experiments showcased in **Supplementary Case Study S1**.

In contrast, for computer science, "computational experiments" can be performed simply by writing code and running it in a virtual computing environment, which significantly increases the feasibility of AI-generated ideas. Hence, while AI can cheaply and rapidly propose a vast number of research concepts, the gap between proposing an idea and possessing the physical resources and expertise to execute it remains a major obstacle for AI-driven automation—particularly outside of purely computational fields.

In this context, Airalogy Platform offers new opportunities for achieving fully AI-automated research in every scientific field, for two key perspectives:

## 1. Airalogy Protocols as digital models of physical lab resources and constraints

As a natural science lab increasingly adopts the Airalogy Platform, the experiments it can perform will be encoded as a set of Airalogy Protocols. These Protocols effectively represent the full range of that lab's experimental capabilities. When an AI generates research ideas without constraints, it often proposes experiments that cannot be performed with available resources. By contrast, if we limit the AI to only those Protocols that a lab (or institution) can execute, the resulting ideas become inherently feasible. Likewise, if the AI needs to design entirely new Protocols for innovative research, it can do so while respecting the lab's known resource boundaries.

## 2. Research as a Service (RaaS) enables AI to complete experiments without "embodiment"

As described earlier, the Airalogy Platform could evolve into a hub for "research micro services," each represented by an Airalogy Protocol offered by individuals or institutions. In this scenario, an AI seeking to achieve a research goal would not physically perform experiments; rather, it would procure the appropriate services from the Airalogy Platform. These services might be carried out by human operators, but from the AI's perspective, they are encapsulated offerings that yield experimental data (or other materials).

Even if the resulting outputs must be physically transferred from one service provider to another, mature logistics systems can handle those shipments. The AI would simply coordinate the process—analyzing data from each service provider and overseeing the transportation of any necessary materials. Ultimately, the AI could orchestrate a fully realized research project without possessing any physical lab infrastructure itself.

Notably, this "labless" research paradigm applies equally to human scientists, who may one day conduct comprehensive natural science experiments using only a computer, much like today's computational scientists. In this model, researchers merely specify experimental requirements and analyze resulting data, while specialized providers handle the physical execution—unlocking entirely new efficiencies and possibilities for collaboration and discovery.

# S2 Supplementary Methods

## S2.1 Airalogy Framework

The Airalogy Framework is designed to facilitate the recording of research data based on user-defined research protocols. It comprises three core components: Airalogy Syntax, Airalogy Protocol, and Airalogy Platform. Airalogy Syntax defines the legal syntax related to the Airalogy Framework. Using this Syntax, users can create an Airalogy Protocol, typically consisting of a series of files that adhere to this Syntax, organized within a folder. Each Airalogy Protocol can be parsed and executed by the Airalogy Platform, automatically generating an interactive data Recording Interface along with corresponding data storage structures and methods (**Supplementary Figs. S27, S28**). An Airalogy Protocol includes three main parts: Markdown, Model, and Assigner (**Supplementary Fig. S29**). These components allow users to customize research protocols in six key aspects: content, multimedia information, data fields, data field type constraints, validation relationships, and assignment relationships (**Supplementary Table S4**). Similar to a software framework[89], the design of the Airalogy Framework aims to assist scientists in focusing their time and energy on defining and developing the six aspects of an Airalogy Protocol, without needing to worry about the design of data recording interfaces or data storage structures and methods. This enables researchers to efficiently create high-quality, research-specific Protocols for recording research data (**Supplementary Table S5**).

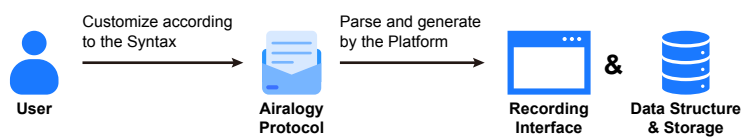

**Supplementary Fig. S27: Defining and using an Airalogy Protocol.**
A user defines an Airalogy Protocol using the Airalogy Syntax. The Airalogy Platform then parses the Protocol and automatically generates an easy-to-use, interactive interface for recording research data. It also automatically constructs the corresponding data structures and storage methods for the recorded data.

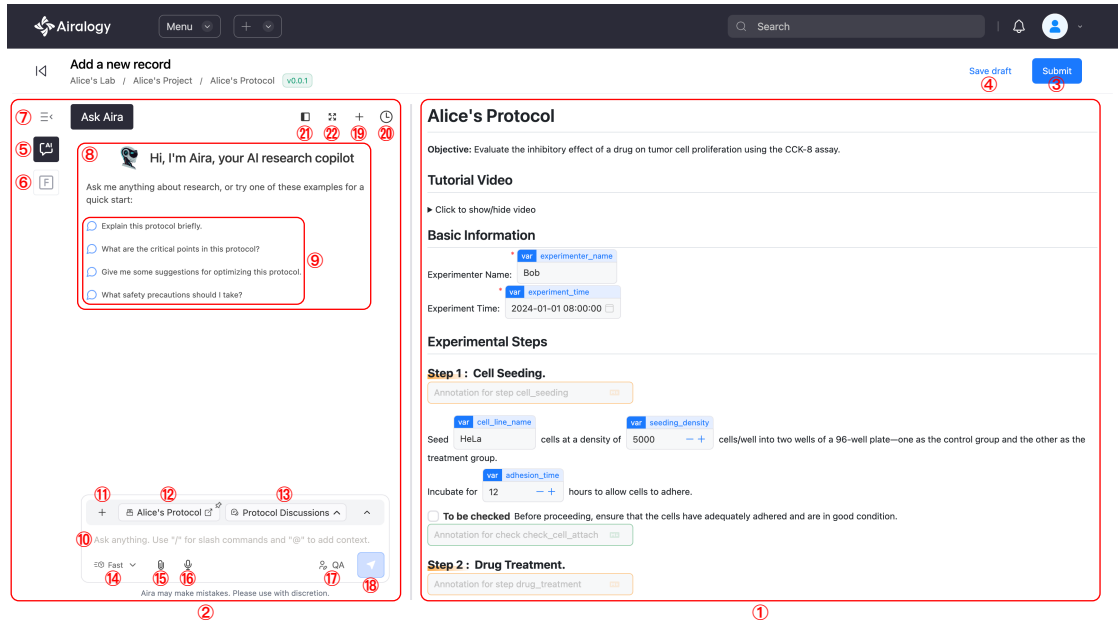

**Supplementary Fig. S28: The interface for recording data on the Airalogy Platform using an Airalogy Protocol.**

Shown with Alice's Protocol (**Fig. 2**), the page has two primary areas (①, ②). Area ① is the WYSIWYG Airalogy Protocol Recording Interface where users enter research data; after completion, they can click "Submit" (③) to generate a new Airalogy Record or Save draft (④) to store a temporary version. Area ② is an extensible sidebar that hosts special tools: for example, the AI Chat pane (⑤) opens a window for interacting with the Platform's AI copilot, Aira, whereas the Fields pane (⑥) displays detailed information for all Airalogy Fields in the current Protocol. The sidebar can be collapsed with button (⑦) to focus solely on Area ①. In AI Chat mode, an introductory panel (⑧) briefly explains Aira and offers sample prompts (⑨). Users type questions in box ⑩, and the "+" button (⑪) lets them inject context—such as relevant Protocols or Records—or limit the scope of any Discussions search (e.g., this Protocol, Project, Lab, or all public Discussions). In this example, "Alice's Protocol" is injected as context (⑫), and searches are restricted to the Protocol's own Discussions (⑬). A model selector (⑭) switches between AI models; an attachment button (⑮) uploads files (e.g., images); and a microphone button (⑯) enables voice input. The mode switch (⑰) toggles Aira's current mode—here "QA" mode for normal conversation, with alternatives such as "Field input", which extracts data from user text or multimodal files and populates the corresponding Airalogy Fields in Area ①. After composing a message, users click button ⑱ to receive Aira's response. They can start a new chat (⑲) or review past conversations (⑳). The layout of the AI Chat Interface is flexible: it can be pinned to a different area of the webpage (㉑) or expanded to full screen mode (㉒).

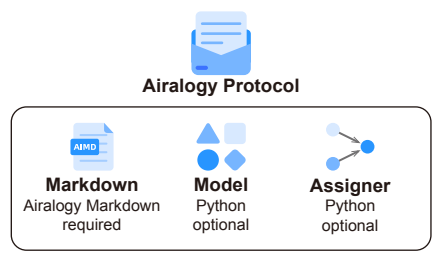

**Supplementary Fig. S29: Main components of an Airalogy Protocol.**
An Airalogy Protocol mainly contains three components: Markdown, Model, and Assigner. The Markdown is written in the file format of Airalogy Markdown, which is required for an Airalogy Protocol, while the Model and Assigner are written in Python and are optional.

**Supplementary Table S4: The three main components of an Airalogy Protocol and the six customization aspects allowed for these components.**

| Component | Customization aspect | Explanation |
|---|---|---|
| Markdown | 1. Content of a Protocol | Users can primarily use text to describe the detailed content of a Protocol related to a research activity (such as an experiment), including the title, introduction, etc. |
| | 2. Multimedia files of a Protocol | Users can insert multimedia files, including images, videos, audio, documents (such as PDFs), etc., into a Protocol to provide more intuitive and detailed descriptions of the research activity. |
| | 3. Data fields in a Protocol | Users can customize the various data fields required in a Protocol. These fields typically include variables, steps, and checkpoints associated with a research activity. For example, if users need to add a data field for the solvent volume in a Protocol to record the amount of solvent used in an experiment, they can insert a data field using the following syntax: `{{var|solvent_volume}}`, allowing them to record the solvent volume in the data field with the ID `solvent_volume`, which falls under the Variable (var) type. |
| Model | 4. Type constraints of the data fields defined in aspect 3 | Users can further define the data types of the custom data fields defined in aspect 3. For example, the `solvent_volume` data field can be defined as a floating-point number. If the user enters data that is not a floating-point number (e.g., a string of alphabet letters) in this field, the Platform will indicate a type error. |
| | 5. Validation relationships of the data fields defined in aspect 3 | Users can further define validation rules for the data fields specified in aspect 3. For example, based on the `solvent_volume` defined in aspect 3 and its type constraint from aspect 4, a validation rule can be added to ensure that the floating-point number is greater than zero. If the user enters a negative value, the system will trigger a validation error. Moreover, users can define validation relationships among multiple data fields at the same time. For instance, one might specify that the total volume of a mixture must be greater than or equal to the sum of its individual component volumes. If this condition is not met, the Platform will again report a validation error. |

| Assigner | 6. Assignment relationships of the data fields defined in aspect 3 | If users have defined multiple data fields in aspect 3 that have dependencies and assignment relationships, they can further customize these relationships. For example, if two data fields `solvent_volume` and `solvent_volume_2` are defined in aspect 3, and `solvent_volume_2` is always twice the value of `solvent_volume`, users can define this assignment relationship. Thus, whenever a new value is entered for `solvent_volume`, the Platform will automatically set `solvent_volume_2` to be twice the value of `solvent_volume`. If the value of `solvent_volume` is 5, the Platform will automatically assign the value of 10 to `solvent_volume_2`. |
|---|---|---|

**Supplementary Table S5: Responsibilities for different design aspects of a research data recording system and the corresponding reasons.**

| Design aspect | Responsibility | Reason |
|---|---|---|
| The six customizable aspects of the research Protocol described in **Supplementary Table S4** | Users | Research activities are diverse, resulting in highly variable Protocols. Therefore, it is essential to allow users to fully customize these six aspects to maximize the quality of Protocols and effectively meet the specific needs of actual research activities . |
| Layout, style, and interaction of the Interface for recording research data | Platform | Generating the research Recording Interface requires extensive computer science expertise, which poses a significant barrier for users. Therefore, the Airalogy Platform should automatically parse the user-defined Airalogy Protocol and generate the corresponding Interface. |
| Data structures and methods for storing research data | Platform | Designing data structures and methods for storing research data typically requires advanced computer science expertise, which most users do not have. Therefore, once users have customized their Airalogy Protocol, the Airalogy Platform should automatically generate the necessary data storage structures based on the Protocol. Furthermore, each time a user records new research data, the Platform should automatically store it in the database. The auto-generated data structures must be standardized to ensure global users can fully understand and utilize the data in the future. |

## S2.2 Airalogy Protocol

**S2.2.1 File structure**

An Airalogy Protocol is typically organized in a folder containing a series of files. The file structure of the Airalogy Protocol is designed to ensure that each Protocol can be effectively parsed.

As shown in **Supplementary Fig. S29**, among the three components of the Airalogy Protocol, only the Markdown component is required, while the Model and Assigner components are optional. Therefore, the simplest Airalogy Protocol may consist solely of the files related to the Markdown component (**Supplementary Code S2**). Additionally, the Airalogy Protocol may include files for both the Protocol and Model components (**Supplementary Code S3**), or it may contain files for all three components: Protocol, Model, and Assigner (**Supplementary Code S4**).

```
protocol/
└── protocol.aimd
```
**Supplementary Code S2: The simplest file structure of an Airalogy Protocol.**
By convention, Airalogy Protocol–related files are stored in a folder named `protocol`. The `.aimd` file extension signifies that `protocol.aimd` is an Airalogy Markdown file.

```
protocol/
├── protocol.aimd
└── model.py
```
**Supplementary Code S3: The file structure of an Airalogy Protocol containing only the Markdown and Model components.**
The `.py` is the filename extension of a Python file, indicating `model.py` is a Python file.

```
protocol/
├── protocol.aimd
├── model.py
└── assigner.py
```
**Supplementary Code S4: The file structure of an Airalogy Protocol containing Markdown, Model, and Assigner components.**

**S2.2.2 Airalogy Protocol Markdown**

The Airalogy Protocol Markdown is a required component of an Airalogy Protocol (**Supplementary Fig. S29**) and is defined in an Airalogy Markdown file named `protocol.aimd` (**Supplementary Code S2**). **Supplementary Code S5** shows a simple Airalogy Protocol that contains only the Markdown component:

Airalogy Protocol Markdown (`protocol.aimd`)

```
{{step|select_solvent, 1}} Select the appropriate solvent.
Solvent name: {{var|solvent_name}}
Solvent volume (L): {{var|solvent_volume}}
{{check|check_remaining_volume}} Please check if the remaining volume of the selected solvent
is sufficient.
```

**Supplementary Code S5: An example of an Airalogy Protocol containing only the Markdown component.**
In the `protocol.aimd` file, four Airalogy Fields are defined: two Variables (`var`) with the IDs `solvent_name` and `solvent_volume`, one Step (`step`) with the ID `select_solvent`, and one Checkpoint (`check`) with the ID `check_remaining_volume`.

Airalogy Markdown (AIMD) is a lightweight, custom markup language newly designed by Airalogy. It is similar to standard Markdown (CommonMark[90]) and includes well-defined syntax rules (the complete Airalogy Markdown specification is available at https://github.com/airalogy/airalogy; here we focus only on the core syntax relevant to this study). By leveraging Airalogy Markdown syntax, users can employ a simple text editor to create formatted text and customize three key aspects of an Airalogy Protocol: content, multimedia information, and data fields (**Supplementary Table S4 Customization aspects 1–3**).

Airalogy Markdown is compatible with the CommonMark specification, but also features syntax extensions tailored for the Airalogy Framework. A key feature is the "template" syntax, which allows users to insert and customize data fields (referred to as "Airalogy Fields") (**Supplementary Table S4 Customization aspect 3**). In Airalogy Markdown, these Fields are denoted by `{{...}}`. Common templates for defining Airalogy Fields are listed in **Supplementary Table S6**. Once parsed by the Airalogy Platform, these Fields generate the corresponding data-entry input boxes in the Recording Interface (detailed in **Supplementary Method S2.4.1**). Users can then enter data directly into these Fields to record relevant research data.

**Supplementary Table S6: Commonly used templates for defining Airalogy Fields in an Airalogy Protocol.**

| Template name | Syntax format | Function |
|---|---|---|
| Variable | `{{var|<variable_id>}}` | Used to mark a data field that represents a variable whose value needs to be recorded or assigned dynamically in an Airalogy Protocol. |
| Step | `{{step|<step_id>, <step_level>}}` | Used to mark a step that needs to be executed in an Airalogy Protocol. The `<step_level>` is an integer greater than or equal to 1, indicating the hierarchical level of the step. |

| Checkpoint | `{{check|<checkpoint_id>}}` | Used to mark a checkpoint that needs to be checked in an Airalogy Protocol. |
|---|---|---|

It is worth noting that within the Airalogy Framework, users can define any number of Variables, Steps, and Checkpoints using the aforementioned templates, provided they meet the following basic requirements:

- `<var_id>`, `<step_id>`, and `<checkpoint_id>` must follow the same snake_case naming convention used in Python.
- No two `<var_id>`, `<step_id>`, or `<checkpoint_id>` may share the same ID.

### S2.2.3 Airalogy Protocol Model

To obtain standardized research data that are easy to analyze, it is often necessary to specify clear data type constraints and add validation rules for the recorded research data fields. Therefore, in an Airalogy Protocol, users can include a `model.py` file upon `protocol.aimd` (**Supplementary Codes S3, S6**) to further define type constraints and add validation rules for the Airalogy Fields specified in the Markdown file (**Supplementary Table S4 Customization aspects 4, 5**). It is important to note that the Model cannot exist independently; it must be used in conjunction with the Airalogy Protocol Markdown.

Airalogy Protocol Markdown (`protocol.aimd`)

```
{{step|select_solvent, 1}} Select the appropriate solvent.
Solvent name: {{var|solvent_name}}
Solvent volume (L): {{var|solvent_volume}}
{{check|check_remaining_volume}} Please check if the remaining volume of the selected solvent
is sufficient.
```

Airalogy Protocol Model (`model.py`)

```python
from pydantic import BaseModel, Field

class VarModel(BaseModel):
    solvent_name: str
    solvent_volume: float = Field(gt=0)
```

**Supplementary Code S6: An example of an Airalogy Protocol containing the Model component.**

Building on the `protocol.aimd` file in **Supplementary Code S5**, users can add a `model.py` file to further define data types and validation constraints for the Variables declared in the Markdown. In this example, these constraints are specified in a class named `VarModel`. The `solvent_name` field is constrained to be a string (`str`), while the `solvent_volume` field is constrained to be a floating-point number (`float`) and must be greater than zero. Since Step and Checkpoint fields have fixed data types and structures (**Supplementary Code S14**), there is no need to define them in `model.py`.

The Model not only supports type constraints and numerical validation for individual Airalogy Fields, but also allows for combined validation across multiple Airalogy Fields (**Supplementary Code S7**).

Airalogy Protocol Markdown (`protocol.aimd`)

```
Integer A: {{var|a}}
Integer B: {{var|b}}
```

Airalogy Protocol Model (`model.py`)

```python
from pydantic import BaseModel, Field, model_validator

class VarModel(BaseModel):
    a: int = Field(gt=0)
    b: int = Field(gt=0)

    @model_validator(mode="after")
    def a_plus_b_lt_10(self):
        if self.a + self.b >= 10:
            raise ValueError("a + b must less than 10")
        return self
```

**Supplementary Code S7: An example code of an Airalogy Protocol with combination validation in the Model component.**

In the `model.py` file, we individually define `a` and `b` fields to be positive integers (`int` with `gt=0`). Additionally, a combination validation rule is applied to ensure that the sum of `a` and `b` is less than 10.

The Airalogy Protocol Model provides comprehensive support for various data type constraints. Moreover, the open-source Python package `airalogy` extends this functionality to multiple multimodal file types, enabling constraints on multimodal data. This allows users to record diverse types of research data through the Airalogy Platform's Recording Interface (including images, videos, audio, and files such as CSV, PDF, Markdown, Word, Excel, PowerPoint, etc). For detailed syntax instructions, please refer to: https://github.com/airalogy/airalogy.

We chose Python as the underlying programming language for the Model and Assigner primarily because it has become one of the world's most widely used languages and is a mainstream choice in AI development[91]. Python not only offers high flexibility and an active community but, most importantly, is easy to learn, making it an excellent option for researchers with limited programming experience. The Airalogy Protocol Model is mainly built on the open-source Python package

Pydantic[92], and has been further extended to meet the needs of research data type constraints and validation.

## S2.2.4 Airalogy Protocol Assigner

In real-world research data recording, many values can be derived through calculations based on other data fields. To support such logic, users can include an `assigner.py` file in an Airalogy Protocol—alongside `protocol.aimd` and `model.py`—to define the Airalogy Protocol Assigner (**Supplementary Code S4**). This file specifies assignment relationships among Airalogy Fields (**Supplementary Table S4 Customization aspect 6**), thereby enabling automatic value assignment functionality.

For instance, if a user is preparing a solution with a specific target concentration and knows both the solute's molar mass and the desired volume, the Assigner can automatically calculate the required solute mass (**Supplementary Code S8**). Consequently, within the Airalogy Protocol Recording Interface, researchers need only enter the solute's molar mass and the target volume; the Platform will compute and populate the required solute mass, greatly enhancing the efficiency and accuracy of research data recording.

Airalogy Protocol Markdown (`protocol.aimd`)

```
Solvent name: {{var|solvent_name}}
Target solution volume (L): {{var|target_solution_volume}}
Solute name: {{var|solute_name}}
Solute molar mass (g/mol): {{var|solute_molar_mass}}
Target molar concentration (mol/L): {{var|target_molar_concentration}}
Required solute mass (g): {{var|required_solute_mass}}
```

Airalogy Protocol Model (`model.py`)

```python
from pydantic import BaseModel, Field

class VarModel(BaseModel):
    solvent_name: str
    target_solution_volume: float = Field(gt=0)
    solute_name: str
    solute_molar_mass: float = Field(gt=0)
    target_molar_concentration: float = Field(gt=0)
    required_solute_mass: float = Field(gt=0)
```

Airalogy Protocol Assigner (`assigner.py`)

```python
from airalogy.assigner import AssignerBase, AssignerResult, assigner

class Assigner(AssignerBase):
```

```python
@assigner(
    assigned_fields=["required_solute_mass"],
    dependent_fields=[
        "target_solution_volume",
        "solute_molar_mass",
        "target_molar_concentration",
    ],
    mode="auto",
)
def calculate_required_solute_mass(dependent_fields: dict) -> AssignerResult:
    target_solution_volume = dependent_fields["target_solution_volume"]
    solute_molar_mass = dependent_fields["solute_molar_mass"]
    target_molar_concentration = dependent_fields["target_molar_concentration"]

    required_solute_mass = (
        target_solution_volume * target_molar_concentration * solute_molar_mass
    )

    return AssignerResult(
        assigned_fields={
            "required_solute_mass": required_solute_mass,
        },
    )
```

**Supplementary Code S8: An example of an Airalogy Protocol with automatic calculation defining by the Assigner component.**

In the `protocol.aimd` file, Variables required for preparing a solution are defined. These Variables are further constrained in the `model.py` file to ensure correct data types and validation rules. To enable automatic calculation of the `required_solute_mass`, the `assigner.py` file defines an Airalogy Protocol Assigner. The `Assigner` class uses the `assigner` decorator to specify that the `required_solute_mass` Field is dependent on `target_solution_volume`, `solute_molar_mass`, and `target_molar_concentration`. The `calculate_required_solute_mass` method computes the required solute mass based on these dependencies and assigns the result automatically. The `mode="auto"` setting indicates that this assignment is triggered automatically—the Airalogy Protocol Recording Interface will monitor the inputs for `target_solution_volume`, `solute_molar_mass`, and `target_molar_concentration`, and once valid values are entered, it will automatically trigger the assigner to calculate and populate the `required_solute_mass` Field. In addition to the automatic mode, users can also define other modes (e.g., `mode="manual"`) which require manual intervention via a button click on the Recording Interface to trigger the calculation.

The strength of the Assigner lies in its flexibility and extensibility. The relationships between dependent Airalogy Fields and assigned Airalogy Fields can theoretically be anything, as long as they can be implemented in Python code. For example, this can be achieved by writing code using Python's built-in functions, calling third-party packages, making external API requests, or utilizing AI services through API calls (**Supplementary Code S9**). The ability to define complex dependencies and calculations within the Assigner allows for sophisticated data automation.

Airalogy Protocol Markdown (`protocol.aimd`)

```
Integer A: {{var|a}}
Integer B: {{var|b}}
Integer C (equal to the sum of A and B): {{var|c}}
```

Airalogy Protocol Model (`model.py`)

```python
from pydantic import BaseModel

class VarModel(BaseModel):
    a: int
    b: int
    c: int
```

Airalogy Protocol Assigner (`assigner.py`) (using Python built-in functions)

```python
from airalogy.assigner import AssignerBase, AssignerResult, assigner

class Assigner(AssignerBase):
    @assigner(
        assigned_fields=["c"],
        dependent_fields=["a", "b"],
        mode="auto",
    )
    def calculate_sum(dependent_fields: dict) -> AssignerResult:
        a = dependent_fields["a"]
        b = dependent_fields["b"]

        c = a + b

        return AssignerResult(
            assigned_fields={"c": c},
        )
```

Airalogy Protocol Assigner (`assigner.py`) (using third-party packages)

```python
from airalogy.assigner import AssignerBase, AssignerResult, assigner
import numpy as np

class Assigner(AssignerBase):
    @assigner(
```

```python
            assigned_fields=["c"],
            dependent_fields=["a", "b"],
            mode="auto",
    )
    def calculate_sum(dependent_fields: dict) -> AssignerResult:
        a = dependent_fields["a"]
        b = dependent_fields["b"]

        c = np.add(a, b)

        return AssignerResult(
            assigned_fields={"c": c},
        )
```

Airalogy Protocol Assigner (`assigner.py`) (using API requests through the internet)

```python
import requests
from airalogy.assigner import AssignerBase, AssignerResult, assigner

class Assigner(AssignerBase):
    @assigner(
        assigned_fields=["c"],
        dependent_fields=["a", "b"],
        mode="auto",
    )
    def calculate_sum(dependent_fields: dict) -> AssignerResult:
        a = dependent_fields["a"]
        b = dependent_fields["b"]

        response = requests.get(f"https://api.example.com/sum?a={a}&b={b}")
        if response.status_code != 200:
            return AssignerResult(
                success=False,
                error_message="Failed to calculate the sum by API",
            )
        data = response.json()
        c = data["sum"]

        return AssignerResult(
            success=True,
            assigned_fields={"c": c},
        )
```

Airalogy Protocol Assigner (`assigner.py`) (using AI services)

```python
from airalogy.assigner import AssignerBase, AssignerResult, assigner
```

```python
from openai import OpenAI

class Assigner(AssignerBase):
    @assigner(
        assigned_fields=["c"],
        dependent_fields=["a", "b"],
        mode="manual",
    )
    def calculate_sum(dependent_fields: dict) -> AssignerResult:
        a = dependent_fields["a"]
        b = dependent_fields["b"]
        openai_client = OpenAI(
            base_url="https://dashscope.aliyuncs.com/compatible-mode/v1"
        )
        chat_completion = openai_client.chat.completions.create(
            messages=[
                {
                    "role": "user",
                    "content": f"{a} + {b} = ? Please only respond with the sum of the two numbers.",
                }
            ],
            model="qwen-turbo",
        )
        c = (
            int(chat_completion.choices[0].message.content)
            if chat_completion.choices[0].message.content is not None
            and chat_completion.choices[0].message.content.isnumeric()
            else None
        )
        if c is int:
            return AssignerResult(
                success=True,
                assigned_fields={"c": c},
            )
        return AssignerResult(
            success=False,
            error_message="Failed to calculate the sum by AI",
        )
```

**Supplementary Code S9: Examples of Airalogy Protocol Assigners using different methods.**
This set of examples demonstrates the flexibility and extensibility of the Assigner for data automation by showing how different methods can be used to calculate the sum of two integers, a and b, and assign the result to c. The four methods demonstrated are: 1. using Python built-in functions; 2. using third-party packages (NumPy[87] as an example); 3. using API requests through the internet; and 4. using AI services (OpenAI[93] as an example).

Another unique aspect of the Airalogy Protocol Assigner is its capability to handle complex research recording and analysis scenarios, where the assignment relationships between Airalogy Fields can be extremely intricate (e.g., multi-level or multi-dependency). To support this complexity, the Assigner internally constructs a dependency graph among the Airalogy Fields, known as the Airalogy Field Dependency Graph. This graph represents all Field assignment relationships within an Airalogy Protocol, enabling the Assigner to track dependencies and perform assignment operations accordingly. The assignment relationships between Airalogy Fields can be arbitrarily dependent and multi-tiered, provided that the corresponding Airalogy Field Dependency Graph meets the following constraints:

1. It must be a Directed Acyclic Graph (DAG).
2. Each Airalogy Field can be assigned a value by at most one Assigner function.

For example, as shown in **Supplementary Code S10** and **Supplementary Fig. S30**, there is an Airalogy Protocol containing nine different Airalogy Fields (F1 to F9). The assignment relationships among these Fields are defined by three different Assigner functions (A1 to A3), and can be represented by an Airalogy Field Dependency Graph. Suppose a user enters data in the associated Airalogy Protocol Recording Interface in the order F1 to F9. The Airalogy Platform automatically tracks the status of each Airalogy Field and completes the assignments as follows:

1. At the start, all Airalogy Fields are empty and inactive, waiting for user input.
2. The user fills in F1, which becomes active.
3. A1 depends on F1, so A1 is triggered, and F2 and F3 are assigned (thus they do not require manual input) and become active.
4. A3 depends on F2 and F3, as well as F6 and F7, but since F6 and F7 are still inactive, F8 is not assigned.
5. The user fills in F4, which becomes active.
6. A2 depends on F4 and F5; because F5 is still inactive, F6 is not assigned.
7. The user fills in F5, which becomes active.
8. Since F4 and F5 are now active, A2 is triggered, and F6 is assigned (not requiring manual input) and becomes active.
9. The user fills in F7, which becomes active.
10. A3 now detects that F2, F3, F6, and F7 are all active, so A3 is triggered and assigns F8 (again, no manual input needed), which becomes active.
11. The user fills in F9, which becomes active.
12. All Airalogy Fields are now active, indicating that this Airalogy Record is complete and ready for submission.

By utilizing this dependency-tracking mechanism, the Assigner can effectively manage complex research recording scenarios and ensure that all Fields are correctly assigned, even in multi-dependency, multi-level contexts.

**Airalogy Protocol Markdown (`protocol.aimd`)**

```
Number F1: {{var|f1}}
Number F2 (= F1 + 1): {{var|f2}}
Number F3 (= F1 × 2): {{var|f3}}
Number F4: {{var|f4}}
Number F5: {{var|f5}}
Number F6 (= F4 × F5): {{var|f6}}
Number F7: {{var|f7}}
Number F8 (= F2 + F3 + F6 + F7): {{var|f8}}
Number F9: {{var|f9}}
```

**Airalogy Protocol Model (`model.py`)**

```python
from pydantic import BaseModel

class VarModel(BaseModel):
    f1: float
    f2: float
    f3: float
    f4: float
    f5: float
    f6: float
    f7: float
    f8: float
    f9: float
```

**Airalogy Protocol Assigner (`assigner.py`)**

```python
from airalogy.assigner import AssignerBase, AssignerResult, assigner

class Assigner(AssignerBase):
    @assigner(
        assigned_fields=["f2", "f3"],
        dependent_fields=["f1"],
        mode="auto",
    )
    def a1(dependent_fields: dict) -> AssignerResult:
        f1 = dependent_fields["f1"]
        f2 = f1 + 1
        f3 = f1 * 2
        return AssignerResult(
            assigned_fields={"f2": f2, "f3": f3},
        )
```

```python
@assigner(
    assigned_fields=["f6"],
    dependent_fields=["f4", "f5"],
    mode="auto",
)
def a2(dependent_fields: dict) -> AssignerResult:
    f4 = dependent_fields["f4"]
    f5 = dependent_fields["f5"]
    f6 = f4 * f5
    return AssignerResult(
        assigned_fields={"f6": f6},
    )

@assigner(
    assigned_fields=["f8"],
    dependent_fields=["f2", "f3", "f6", "f7"],
    mode="auto",
)
def a3(dependent_fields: dict) -> AssignerResult:
    f2 = dependent_fields["f2"]
    f3 = dependent_fields["f3"]
    f6 = dependent_fields["f6"]
    f7 = dependent_fields["f7"]
    f8 = f2 + f3 + f6 + f7
    return AssignerResult(
        assigned_fields={"f8": f8},
    )
```

**Supplementary Code S10: An example of an Airalogy Protocol with multiple assignment levels.** In this Protocol, eight Airalogy Fields (F1–F8) are interdependently assigned by three Assigner functions (A1–A3). **Supplementary Fig. S30** shows the corresponding Airalogy Field Dependency Graph.

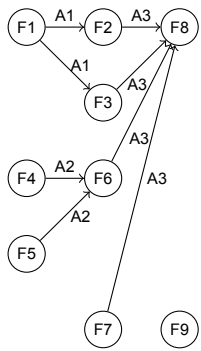

**Supplementary Fig. S30: The Airalogy Field Dependency Graph corresponding to Supplementary Code S10.**

**S2.2.5 Airalogy Protocol IDs and versioning**

A key advantage of Airalogy Protocols is their built-in support for globally unique identifiers and version control. Because each Protocol is packaged as a folder containing an Airalogy Markdown file and optional Python files (**Supplementary Fig. S29**), it can be versioned just like conventional code. On the Airalogy Platform every Protocol is assigned a unique ID, enabling effortless tracking and one-click reuse anywhere on the Platform.

To manage a Protocol's ID and version, we introduce an optional `protocol.toml` file written in the human-readable TOML format[94]. Adding this file to the example in **Supplementary Code S4** yields the structure shown below (**Supplementary Codes S11, S12**):

```
protocol/
├── protocol.aimd
├── model.py
├── assigner.py
└── protocol.toml
```

**Supplementary Code S11: Directory structure of an Airalogy Protocol that includes a TOML file.**
This example extends **Supplementary Code S4** by adding a `protocol.toml` file to manage the Protocol's metadata and version information.

```
[airalogy_protocol]
id = "alice_s_protocol"
version = "0.0.1"
name = "Alice's Protocol"
authors = [
    { name = "Alice", email = "alice@airalogy.com", airalogy_user_id =
"airalogy.id.user.alice" },
]
maintainers = [
    { name = "Alice", email = "alice@airalogy.com", airalogy_user_id =
"airalogy.id.user.alice" },
]
disciplines = ["drug discovery", "biology"]
keywords = ["cck-8", "cell viability", "drug screening", "proliferation assay"]
license = "Apache-2.0"
```

**Supplementary Code S12: Example contents of a TOML file for an Airalogy Protocol.**
Within the `[airalogy_protocol]` table, users define key metadata for the Protocol. Each Airalogy Protocol has a unique ID, and a single ID can have multiple versions, allowing the entire evolution of that Protocol to be tracked (e.g., updating from version "0.0.1" to "0.0.2"). The `name` field provides a memorable, searchable label (not guaranteed unique), while the `authors` and `maintainers` arrays record contributors; only `name` is required, whereas `email` and `airalogy_user_id` are optional. Every registered user has a distinct `airalogy_user_id`, creating a reliable link between a Protocol and its contributors. The `disciplines` and `keywords` fields can be used by the Protocol's author to specify the relevant disciplines or to add keywords that facilitate indexing and searching. Since an Airalogy Protocol is managed with a file structure similar to a standard code package, it can also include a license to inform other users of its sharing terms and conditions.

### S2.2.6 Defining and sharing an Airalogy Protocol Workflow within an Airalogy Protocol

As shown in **Supplementary Fig. S2**, an Airalogy Protocol Workflow, when applied, ultimately produces a series of Airalogy Records associated with that Workflow. Since its primary purpose is data digitization, an Airalogy Protocol Workflow can itself be encapsulated within an Airalogy Protocol. Specifically, an workflow-type Airalogy Field can be used to store the Workflow.

Within the Markdown component of an Airalogy Protocol, users can define a Workflow using a dedicated `workflow` code block or template, with the content written in YAML[95] syntax. For example, the Workflow corresponding to **Supplementary Fig. S1** can be defined as follows (**Supplementary Code S13**):

```
# Airalogy Protocol Workflow for a carbon nanotube dispersion study

```workflow
id: cnt_dispersion
title: Workflow for a carbon nanotube dispersion study
protocols:
  - protocol_index: 1
    protocol_name: Preparation of dispersion solution from carbon nanotube powder
    airalogy_protocol_id: airalogy.id.lab.airalogy.project.paper.protocol.cnt_powder.v.0
.0.1
  - protocol_index: 2
    protocol_name: Ultrasonic dispersion of carbon nanotube solution
    airalogy_protocol_id: airalogy.id.lab.airalogy.project.paper.protocol.cnt_ultrasound
.v.0.0.1
  - protocol_index: 3
    protocol_name: Preparation of low-concentration carbon nanotube dispersion solution
from high-concentration solution
    airalogy_protocol_id: airalogy.id.lab.airalogy.project.paper.protocol.cnt_dilution.v
.0.0.1
  - protocol_index: 4
```





```
    protocol_name: Characterization of carbon nanotube dispersion
    airalogy_protocol_id: airalogy.id.lab.airalogy.project.paper.protocol.cnt_characteri
zation.v.0.0.1
edges:
  - 1 -> 2
  - 2 <-> 4
  - 4 -> 3
  - 3 -> 2
logic: |
  1. The entire dispersion process must occur within a solution system, and the preparat
ion of the dispersion from solid powder can only be the first step of the experiment: Pr
otocol 1 must be the starting point of the Workflow.
  2. Every dispersion system must go through the stages of preparation, ultrasonication,
 and characterization: a Protocol Path must include at least one instance of (Protocol 1
 -> Protocol 2 -> Protocol 4), and this sequence is irreversible.
  3. Based on the characterization results, it is determined whether: 1. The sample need
s to be re-sonicated (Protocol 4 -> Protocol 2), or 2. The dispersion solution needs to
be further diluted before sonication (Protocol 4 -> Protocol 3 -> Protocol 2). After rep
eating either of these two paths, characterization (Protocol 4) must be performed again
to confirm the subsequent results. These two paths can be iteratively followed based on
the outcomes of Protocol 4.
  4. The characterization process (Protocol 4), as the only quality control step in the
experiment, can appear in the middle of the steps but must always be the final step in a
 Protocol Path.
  5. When the characterization results (Protocol 4) meet the research goal, the Workflow
 can be terminated.
```
```

**Supplementary Code S13: Syntax example for defining an Airalogy Protocol Workflow within an Airalogy Protocol.**

The Workflow corresponds to **Supplementary Fig. S1** and is defined within a `workflow` code block. The `protocols` field lists the basic information for each protocol involved, including its index, ID, and name. The `edges` field defines the directed relationships between protocols, using the `protocol_index` to refer to each node in the Workflow Graph. The `logic` field, written in Markdown format, specifies the Graph Logic of the Workflow, which is consistent with the description in **Supplementary Code S1**.

Moreover, because the Workflow is defined within an Airalogy Protocol, it can be managed and shared in the same way as the Protocol itself. As a result, a Workflow designed by one user can be easily shared with users around the world via the Airalogy Platform.

## S2.3 Airalogy Record

When researchers use a specific Airalogy Protocol to record data, each resulting record is referred to as an Airalogy Record.

### S2.3.1 Data structure of Airalogy Fields

Within the Airalogy Framework, research data recording centers on capturing values for each Airalogy Field defined in the Airalogy Protocol. Because these Fields' specifications depend on user-defined content in the Markdown and Model, the structure of any given Airalogy Record is inherently linked to those definitions. The Airalogy Platform automatically parses the relevant user-defined content and generates the corresponding data structure for each Airalogy Record (**Supplementary Fig. S27**). Note that the Assigner does not alter the data structure of the Airalogy Protocol.

In Airalogy Framework, we use JSON to manage the research data generated from an Airalogy Protocol. The data structure of each Airalogy Field may vary according to its template type. In general, the structure of Step and Checkpoint Fields is fixed (**Supplementary Code S14**), while the structure of Variable Fields is dictated by the type constraints defined in the Model's `VarModel` class.

```
{
    "annotation": "string",
    "checked": true
}
```

**Supplementary Code S14: Data structure for a Step or a Checkpoint.**
The data structure for a Step or Checkpoint is stored in a JSON object containing two fields: `annotation` and `checked`. The `annotation` field is a string for recording comments or notes. For example, if a user makes an error while performing a Step, they can describe the mistake in text and save it in this field. Additionally, within the Airalogy Platform, `annotation` can embed IDs or URLs for images, videos, or audio files, enabling multimodal data recording beyond simple text. The `checked` field is a Boolean or null value. When it is `true` or `false`, it indicates whether the Step or Checkpoint has been reviewed and confirmed by the user. A `null` value applies only to Steps and signifies that the Step has not been set to check mode, and therefore was not reviewed.

By default, if an Airalogy Protocol is defined only by a single Markdown file, any values recorded in its Variables are stored as JSON strings. However, when a user includes a Model in addition to the Markdown (**Supplementary Code S3**), the JSON data types of those Variable values reflect the definitions specified in the Model.

This distinction can be demonstrated with a simple example. Suppose an Airalogy Protocol includes only the `protocol.aimd` file from **Supplementary Code S6**, and does not include a `model.py` file. If a user records "H2O" for `solvent_name` and "1" for `solvent_volume`, the data structures for both fields are as shown in **Supplementary Code S15**, with the `solvent_volume` value stored as the JSON string `"1"`. In contrast, if the user records the same values in an Airalogy Protocol that also includes the `model.py` file from **Supplementary Code S6**, the Variables' data structures appear as shown in **Supplementary Code S16**. Because `solvent_volume` is defined as a float in the Model, its recorded value of `1.0` is stored as a JSON number.

```
{
    "solvent_name": "H2O",
    "solvent_volume": "1"
}
```

**Supplementary Code S15: Data structure of Variables with only the Airalogy Protocol Markdown defined.**
This JSON object illustrates the data structure when the `model.py` file is omitted from the Airalogy Protocol (**Supplementary Code S6**). In this scenario, all Variable values are stored as JSON strings. For example, the `solvent_name` field stores `"H2O"` as a string, and the `solvent_volume` field stores `"1"` as a string.

```
{
    "solvent_name": "H2O",
    "solvent_volume": 1.0
}
```

**Supplementary Code S16: Data structure of Variables with both the Airalogy Protocol Markdown and Model defined.**
This JSON object shows the data structure when both `protocol.aimd` and `model.py` are present in the Airalogy Protocol (**Supplementary Code S6**). In this case, the recorded values for Variables adhere to the type constraints specified in the Model. For instance, the `solvent_name` field stores `"H2O"` as a string, whereas the `solvent_volume` field stores `1.0` as a numeric value, as defined in the `model.py` file.

### S2.3.2 Complete data structure of Airalogy Records

In an Airalogy Record, data from different templates is stored under fields that share the same name as the template. For example, all Variable data is stored in a field named `var`, Step data in `step`, and Checkpoint data in `check`. In addition to these fields, each Airalogy Record includes a unique record ID and essential metadata—such as the ID of the underlying Airalogy Protocol (to track the source Protocol), the recorder's name, and the recording time—to facilitate effective management and tracking. Together, these elements form the complete data structure of an Airalogy Record. For instance, **Supplementary Code S17** illustrates a possible data structure for an Airalogy Record generated by the Airalogy Protocol shown in **Supplementary Code S6**.

```json
{
    "airalogy_record_id": "airalogy.id.record.01234567-0123-0123-0123-0123456789ab.v.2",
    "record_id": "01234567-0123-0123-0123-0123456789ab",
    "record_version": 2,
    "metadata": {
        "airalogy_protocol_id": "airalogy.id.lab.lab_demo.project.project_demo.protocol.protocol_demo.v.0.0.1",
        "lab_id": "lab_demo",
        "project_id": "project_demo",
        "protocol_id": "protocol_demo",
        "protocol_version": "0.0.1",
        "record_num": 1,
        "record_current_version_submission_time": "2024-01-02T00:00:00+08:00",
        "record_current_version_submission_user_id": "user_demo_2",
        "record_initial_version_submission_time": "2024-01-01T00:00:00+08:00",
        "record_initial_version_submission_user_id": "user_demo_1",
        "sha1": "c486349125db2a468172a4449b9e309b0c756c59"
    },
    "data": {
        "var": {
            "solvent_name": "H2O",
            "solvent_volume": 1.0
        },
        "step": {
            "select_solvent": {
                "annotation": "",
                "checked": null
            }
        },
        "check": {
            "check_remaining_volume": {
                "annotation": "",
                "checked": true
            }
        }
    }
}
```

**Supplementary Code S17: Example of the data structure of an Airalogy Record.**
This JSON object represents a complete data structure for an Airalogy Record generated by the Airalogy Protocol shown in **Supplementary Code S6**. It includes a unique Record identifier (`airalogy_record_id`), metadata containing essential background information (such as the ID of the Airalogy Protocol on which the Record is based, version submission times, user IDs, and other tracking details), and the actual Record data. The `data` field organizes the Airalogy Fields by template: Variables are stored under `var`, Steps under `step`, and Checkpoints under `check`. The `sha1` field contains a hash of the contents of the `data` field, which can be used to verify the Record's integrity and confirm that it has not been tampered with.

It is important to note that (1) each Airalogy Record has a unique ID, (2) it is stored in structured JSON format, and (3) multiple Airalogy Records generated from the same Airalogy Protocol share a uniform data structure. These three features ensure that each Airalogy Record's data is well-organized, making it easy to transfer Records from one Airalogy Protocol to the next. Because the data structure is clearly defined, a subsequent Protocol's designer can readily create methods to analyze the Records produced by a previous Protocol. This is the fundamental reason why, in the example shown in **Fig. 3**, Frank's Protocol is able to automatically process the Records generated by Alice's Protocol.

## S2.4 Airalogy Protocol Recording Interface

### S2.4.1 Airalogy Platform automatically generates the corresponding Recording Interface based on the user-defined Airalogy Protocol

he Airalogy Protocol Recording Interface primarily consists of two elements. The first is the stylized display of the parsed Airalogy Markdown (for instance, headings rendered in different font sizes to indicate various levels, known as Styled Airalogy Markdown). The second element comprises the input boxes corresponding to each Airalogy Field, referred to as Airalogy Field Input Boxes, which serve as the main interactive components for recording research data.

To enable rapid and efficient data entry while enhancing the overall recording experience, users can specify data type constraints for the Airalogy Fields defined in the Markdown by using an Airalogy Protocol Model. Consequently, Airalogy Fields with different type constraints typically appear as Input Boxes with distinct styles or interaction methods (**Supplementary Figs. S31–S33**), effectively improving the ease, efficiency, and accuracy with which researchers record various types of data.

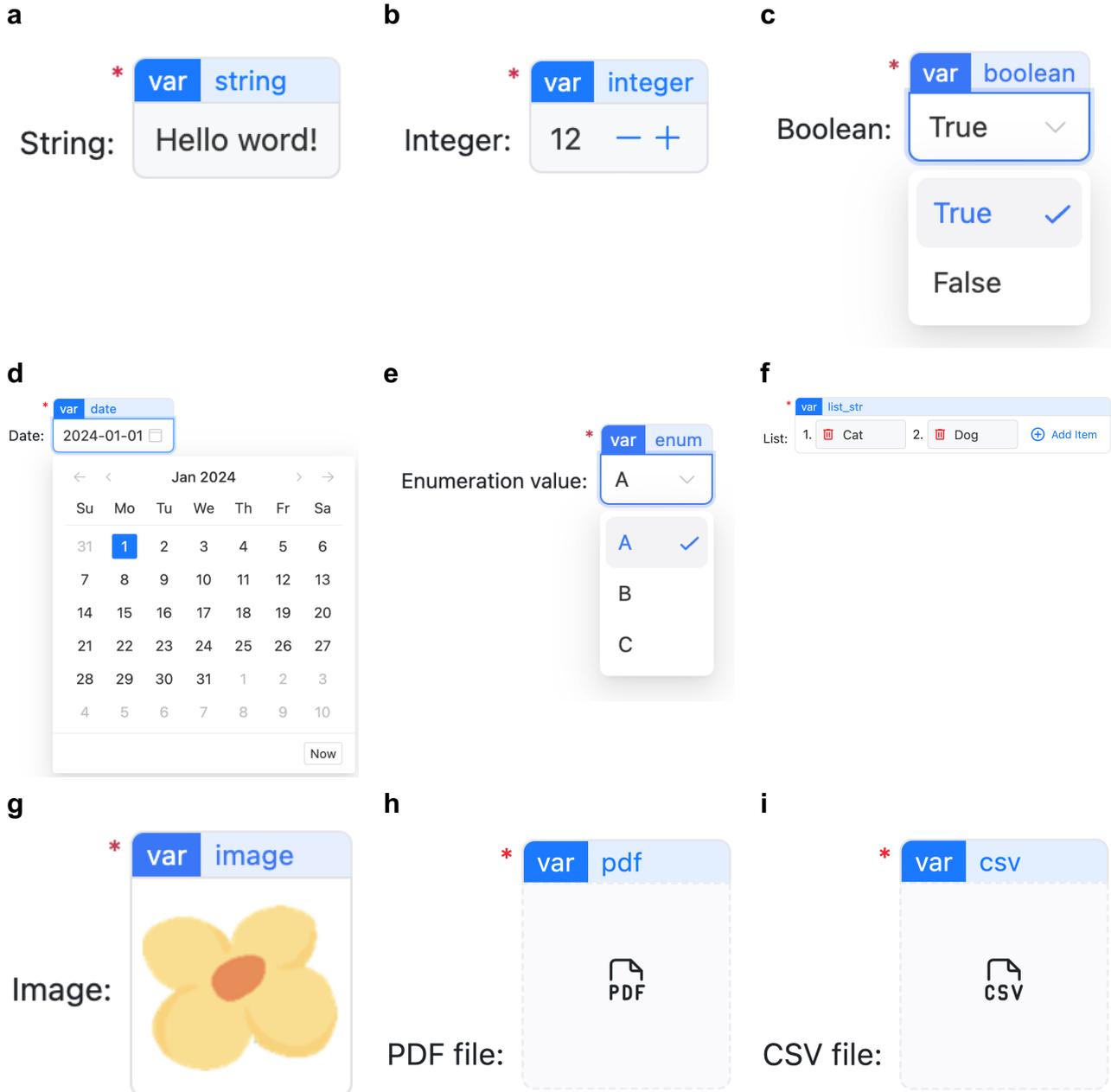

**Supplementary Fig. S31: A schematic representation of various Variable-type Airalogy Field Input Boxes.**

**a**, String type. **b**, Integer type. **c**, Boolean type. **d**, Date type. **e**, Enumeration type. Users can select from all available options via a dropdown menu. **f**, List type. Users can add an arbitrary number of list elements using the "Add item" button or remove elements using a trash-can–style button. **g–i**, Users can record image data (**g**) within an Airalogy Field, and additional modalities such as video and audio are also supported (not shown); users can also upload and preview PDFs (**h**), CSV files (**i**), or other file formats, including TXT, Markdown, Microsoft Word, Excel, PowerPoint, etc. (not shown).

**Supplementary Fig. S32: Step-type Airalogy Fields' Input Boxes and their interactive styles.**
Panels **a–b** define a Step-type Field with the ID `step_without_check`, which automatically generates a step annotation box in the Airalogy Protocol Recording Interface (used for noting unexpected operations or phenomena, though not required). Panels **a** and **b** respectively illustrate the cases where no annotation is provided versus when annotation text is entered. Panels **c–h** define a Step-type Field with the ID `step_with_check`, in which the check mode is enabled; by default (**c–d**), a "To be checked" button appears before the step, allowing the user to indicate whether the step has passed ("Check passed") or failed ("Check failed"). Panels **e–f** show the appearance after passing the check; panels **g–h** show when the step is marked as failed. Panels **c, e, and g** illustrate various check states without annotation, and panels **d, f, and h** correspond to cases where annotation text is provided.

**Supplementary Fig. S33: Checkpoint-type Airalogy Field's Input Box and its interactive styles.**
In this example, a Checkpoint-type Field with the ID `a` is defined. Within the Airalogy Protocol Recording Interface, it is rendered with a "To be checked" button (**a**), allowing the user to indicate whether the checkpoint is passed ("Check passed", **c**) or failed ("Check failed", **e**). Each Checkpoint also has an annotation box for recording any additional notes, which are optional. Panels **b, d, and f** illustrate cases where an annotation has been entered under different check conditions.

However, the Airalogy Protocol Model is primarily implemented in Python, while the Airalogy Protocol Recording Interface is essentially a web page developed using front-end languages such as JavaScript/TypeScript. To ensure that the Recording Interface correctly interprets the types and validation constraints of the various Airalogy Fields defined in the Airalogy Protocol Model and automatically generates the appropriate Input Boxes, an intermediate format is required to bridge the gap between these programming languages. In the Airalogy Framework, we employ JSON Schema[96] for this purpose.

For example, the JSON Schema corresponding to the Airalogy Protocol Model shown in **Supplementary Code S6** is provided in **Supplementary Code S18**. Because JSON Schema is expressed in JSON, it serves as a convenient medium for data exchange across different programming languages. This approach enables the Airalogy Platform to seamlessly transmit information between the Python-based Airalogy Protocol Model and the JavaScript/TypeScript-based Airalogy Protocol Recording Interface, thereby facilitating the accurate parsing of data types and validation constraints and the automatic generation of the Recording Interface.

```json
{
    "properties": {
        "solvent_name": {
            "title": "Solvent Name",
            "type": "string"
        },
        "solvent_volume": {
            "exclusiveMinimum": 0.0,
            "title": "Solvent Volume",
            "type": "number"
        }
    },
    "required": [
        "solvent_name",
        "solvent_volume"
    ],
    "title": "VarModel",
    "type": "object"
}
```

**Supplementary Code S18: Example of an Airalogy Protocol Field JSON Schema automatically generated from an Airalogy Protocol Model.**

This JSON Schema reflects the data types and numerical constraints defined in the Airalogy Protocol Model (**Supplementary Code S6**) for the two Variable-type Airalogy Fields (`solvent_name` and `solvent_volume`) thereby enabling the Airalogy Platform to automatically construct Input Boxes with appropriate styles and interactions within the Airalogy Protocol Recording Interface.

---

Specifically, in order to automatically parse a user-defined Airalogy Protocol and generate its corresponding Airalogy Protocol Recording Interface and Data Storage, the Airalogy Platform follows the process outlined below (**Supplementary Fig. S34**):

**Scenario 1: Airalogy Protocol contains only Markdown (Supplementary Code S2)**

- **Step 1:** The Airalogy Markdown file (`protocol.aimd`) corresponding to the Airalogy Protocol is first parsed, automatically extracting all user-defined Airalogy Fields, including:
  - Based on the Airalogy Markdown content, a preliminary Airalogy Protocol Recording Interface is generated, which includes Styled Airalogy Markdown and Airalogy Field Input Boxes. At this stage, all Airalogy Field Input Boxes use default styles and interactions corresponding to their data types, especially for Variable-type Airalogy Fields, where the Input Boxes are defaulted to string types.
  - The Airalogy Record data structure and database storage methods are automatically generated based on the Airalogy Fields. At this point, all Airalogy Fields are stored in JSON string format.

**Scenario 2: Airalogy Protocol contains both Markdown and Model (Supplementary Code S3)**

- **Step 1:** Same as Step 1 in Scenario 1.
- **Step 2:** The Model (`model.py`) in the Airalogy Protocol is automatically parsed, and the corresponding Airalogy Protocol Field JSON Schema is generated.
- **Step 3:** Based on the data types and value constraints defined in the Airalogy Protocol Field JSON Schema, the Airalogy Field Input Boxes generated in Step 1 are further decorated and refined to make them more user-friendly in terms of style and interaction.
- **Step 4:** The Airalogy Record data structure and database storage methods are further refined according to the rules specified in the Airalogy Protocol Field JSON Schema, ensuring they align with the defined data types and constraints. For example, in the case of **Supplementary Code S16**, since the user defined `solvent_volume` as a float in the Model, this Airalogy Field will be refined as a floating-point number, rather than the default string format.

**Scenario 3: Airalogy Protocol contains Markdown, Model, and Assigner (Supplementary Code S4)**

- **Step 1-4:** Same as Steps 1-4 in Scenario 2.
- **Step 5:** The assignment relationships defined in the Assigner are bound to the corresponding Airalogy Field Input Boxes, enabling these Fields to automatically calculate or assign values based on the defined rules.

Through these steps, the Airalogy Platform can flexibly handle user-defined Airalogy Protocols of varying complexity, automatically generating the corresponding Recording Interfaces and data storages, while supporting different types of user interactions and data processing.

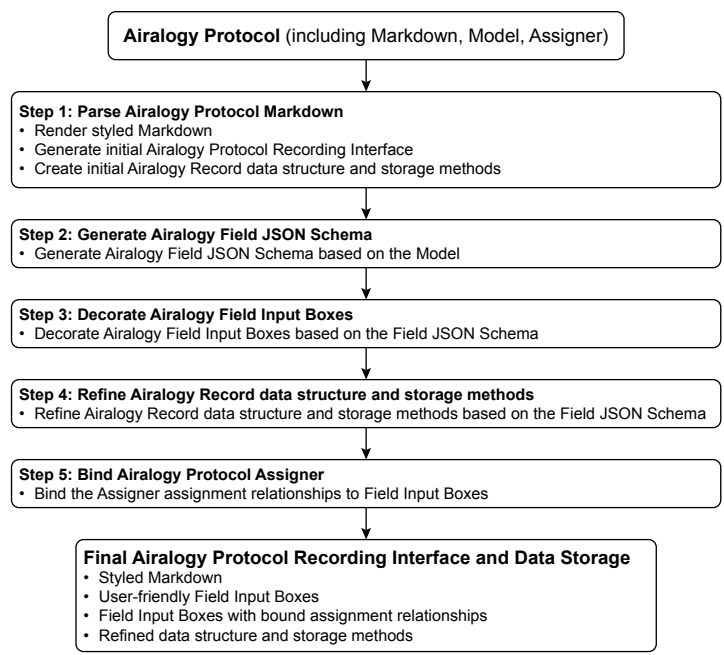

**Supplementary Fig. S34: Steps for the Airalogy Platform to automatically generate the Airalogy Recording Interface and Data Storage.**

## S2.4.2 Users record data on the Airalogy Protocol Recording Interface

When users record research data on the Airalogy Protocol Recording Interface, the Airalogy Platform leverages both the user-defined Airalogy Protocol Model and the Airalogy Protocol Assigner to ensure data validity and to enable automatic value assignment. The process is as follows (**Supplementary Fig. S35**):

1. **User input:** The user enters a value in an Airalogy Field Input Box.
2. **Real-rime validation:** The Airalogy Platform continuously monitors the input in the active Field Input Box. Once the user completes their input, the Platform validates the new value against the pre-defined data type and numerical constraints specified in the Airalogy Protocol Field JSON Schema.
3. **Error handling:** If the entered value does not meet the requirements, an error message is displayed below the Input Box, explaining the reason for the failure (e.g., data type error or violation of value constraints). The user can then correct the input by repeating steps 1 and 2 until the value meets the criteria.
4. **Activation:** Once a value satisfies the requirements, it is marked as valid. The corresponding Input Box is then flagged as "active".

5. **Automatic assignment:** For each active Input Box, the Platform checks whether the corresponding Field is a dependency for any Assigner function. If it is, the Platform further verifies whether all the dependencies required by that Assigner function are active. If all dependencies are met, the Assigner function is triggered to automatically assign values to the related Airalogy Fields, updating their Input Boxes accordingly. If not all dependencies are active, the Assigner remains in a waiting state until they are.

6. **Record completion:** The user continues to fill in the remaining Input Boxes, repeating steps 1–5, until all required Fields contain valid values. For Fields that are auto-assigned, no manual input is needed. At this point, the user can choose to submit the complete Airalogy Record.

7. **Final validation and submission:** When the user clicks the submit button, the Airalogy Platform invokes the Airalogy Protocol Model to perform a comprehensive validation of the entire Airalogy Record (e.g., verifying that all Fields contain valid values and checking complex, interdependent constraints). If the validation passes, the Record is accepted as valid and stored in the database; if it fails, the Interface displays detailed error messages. The user can then correct the issues by revisiting steps 1–6 until the Record is valid.

This process ensures that all user-entered data conforms to the specified requirements and that the automatic assignment mechanism enhances the efficiency and accuracy of research data recording.

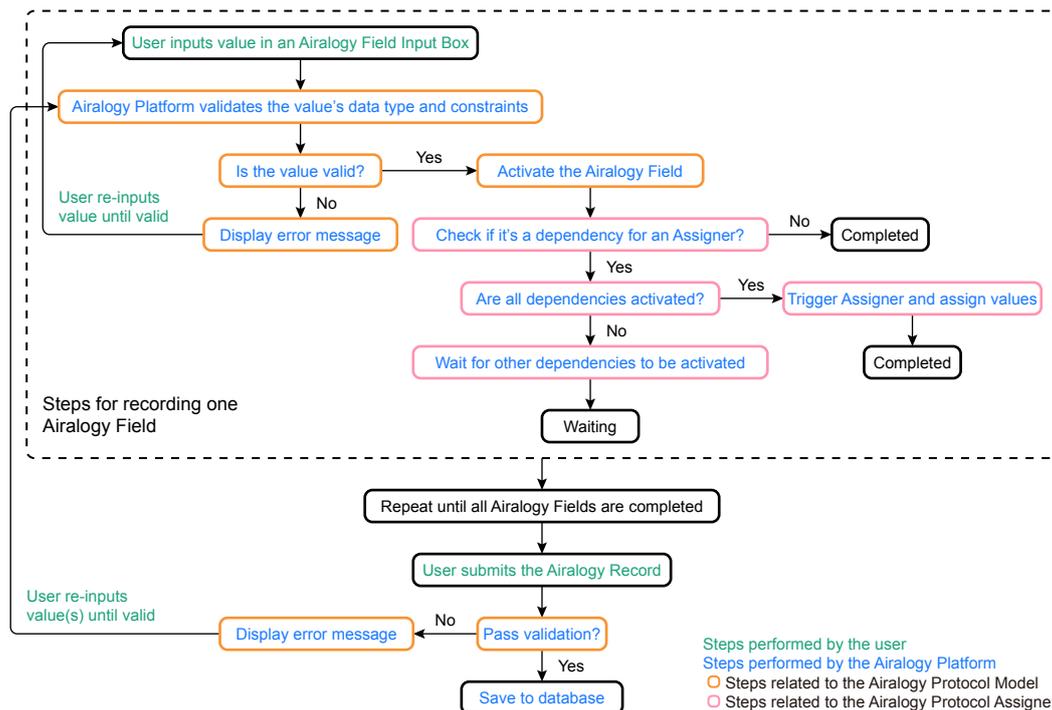

**Supplementary Fig. S35: Steps for recording research data on the Airalogy Protocol Recording Interface with automated support from the Airalogy Platform.**

### S2.4.3 Generating Airalogy Record Reports based on historical Airalogy Records

One significant advantage of Airalogy Records based on the Airalogy Protocol is that any historical Airalogy Record can automatically generate a report (referred to as the Airalogy Record Report). This allows users to intuitively browse historical data or export it as a PDF document, which can then be integrated into local documents or printed. To correctly render the Airalogy Record Report, we only need the static text files associated with the Airalogy Protocol, which include the following:

- Airalogy Protocol Markdown (Airalogy Markdown file)
- Airalogy Protocol Field JSON Schema (JSON file)
- Airalogy Record (JSON file)

The Airalogy Protocol Field JSON Schema is automatically generated and saved when the Airalogy Protocol is first loaded onto the Airalogy Platform. Therefore, users do not need to run the Python files of the Airalogy Protocol Model again when generating the Airalogy Record Report; they can simply reference the already generated Airalogy Protocol Field JSON Schema file.

The process of generating an Airalogy Record Report is outlined below (**Supplementary Fig. S36**), and it shares some similarities with the process of generating the Airalogy Recording Interface (**Supplementary Fig. S35**), so they can be understood in relation to each other:

1. **Step 1:** Parse the Airalogy Protocol Markdown file to automatically extract all user-defined Airalogy Fields, rendering the Styled Airalogy Markdown and Airalogy Field Input Boxes.
2. **Step 2:** Based on the data types defined in the Airalogy Protocol Field JSON Schema, decorate the Airalogy Field Input Boxes to match the appropriate styles.
3. **Step 3:** Parse the Airalogy Record and inject the values of all Airalogy Fields into the corresponding Airalogy Field Input Boxes.

Through this process, the Airalogy Record Report presents the detailed content of the research record in an intuitive and structured manner, offering a user-friendly style that makes Report generation and viewing more efficient and convenient. Notably, to generate an Airalogy Record Report, only the static text files associated with the Airalogy Protocol (the Airalogy Markdown file and JSON files) are required, eliminating the need to run Python files or execute Python code. This removes security risks when generating Airalogy Record Reports. As a result, the Airalogy Protocol and Record have the potential to become the standard format for research data exchange in the future.

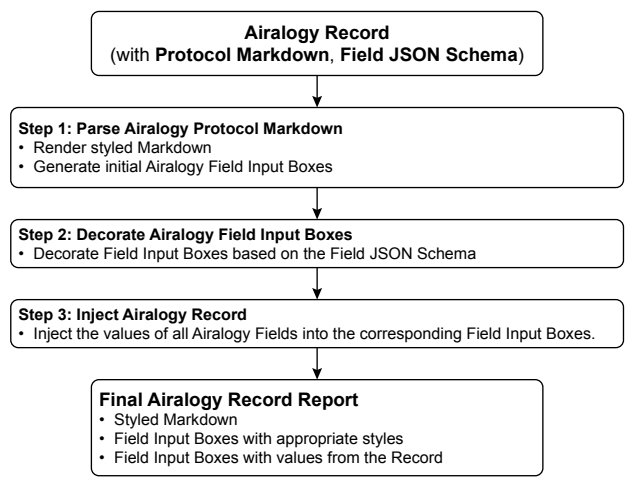

**Supplementary Fig. S36: Steps for automatically generating the Airalogy Record Report.**

## S2.5 Airalogy Platform

The Airalogy Platform (https://airalogy.com) is an AI-driven, community-driven cloud platform developed based on the Airalogy Framework. The Platform provides deep support for the cloud-based application of the Airalogy Protocol, eliminating the need for users to perform any local installations. Users simply upload their custom Airalogy Protocol to the Platform, which automatically parses and executes the Protocol according to the methods outlined above. This enables features such as online Airalogy Protocol previews, research data recording and Airalogy Record generation, automated data analysis, cloud data storage, and the online generation of historical Airalogy Record Reports.

Additionally, we have developed an AI-powered masterbrain[13] for Airalogy, called Airalogy Masterbrain (personified as "Aira"), which is based on LLMs (detailed in **Supplementary Method S2.6**). Fully integrated into the Airalogy Platform, this Masterbrain serves as the Platform's central intelligence and delivers end-to-end support for research-data digitization and automation, including AI-assisted protocol design, intelligent protocol search and recommendation, protocol Q&A, automatic field entry, and record analysis.

### S2.5.1 Organizational structure and access control of the Airalogy Platform

In the Airalogy Platform, the basic unit of user activity is the **Protocol**. The Platform's structure consists of three primary levels: **Lab → Project → Protocol** (**Supplementary Fig. S37a, b**). To promote potential cross-laboratory and cross-project collaborations, access control in the Platform primarily focuses on the **Project** level, as detailed below:

- **Users can verify whether a Lab ID is already registered on the Airalogy Platform.** Each Lab is assigned a globally unique and publicly visible ID, ensuring that researchers can reliably locate or reference the Lab. If Lab IDs were kept hidden, it would hinder global collaboration by making it impossible to confirm or identify a particular Lab.

- **Whether a user can access a Project within a specific Lab depends on the Project's access permissions.** Projects are categorized as Public (accessible to all users) or Private (accessible only to authorized users or groups). This approach allows global users to collaborate on public Projects, while private Projects remain inaccessible to external users, safeguarding data confidentiality.
- **If a user can access a Project, by default, they can view all Protocols and related information within that Project.** This policy balances confidentiality with usability: Project members need comprehensive visibility into every Protocol in order to conduct research effectively. Additional, to accommodate the finer-grained visibility requirements of real-world research management, each Protocol can also be configured with its own access controls—a topic beyond the scope of this article.

Within a Lab, there may be many members, but in practice, certain Projects may need to be restricted to a subset of lab members. Therefore, for private Projects, the Platform provides fine-grained access control (**Supplementary Fig. S37c**):

- Project administrators can directly grant certain users access to a Project within the Lab.
- Lab administrators can divide users within the Lab into different groups, and then Project administrators can grant access to specific groups for a Project.

Through this multi-level organizational structure and flexible access control mechanism, the Platform effectively promotes both intra-laboratory and cross-laboratory collaboration while ensuring the security and privacy management of research data.

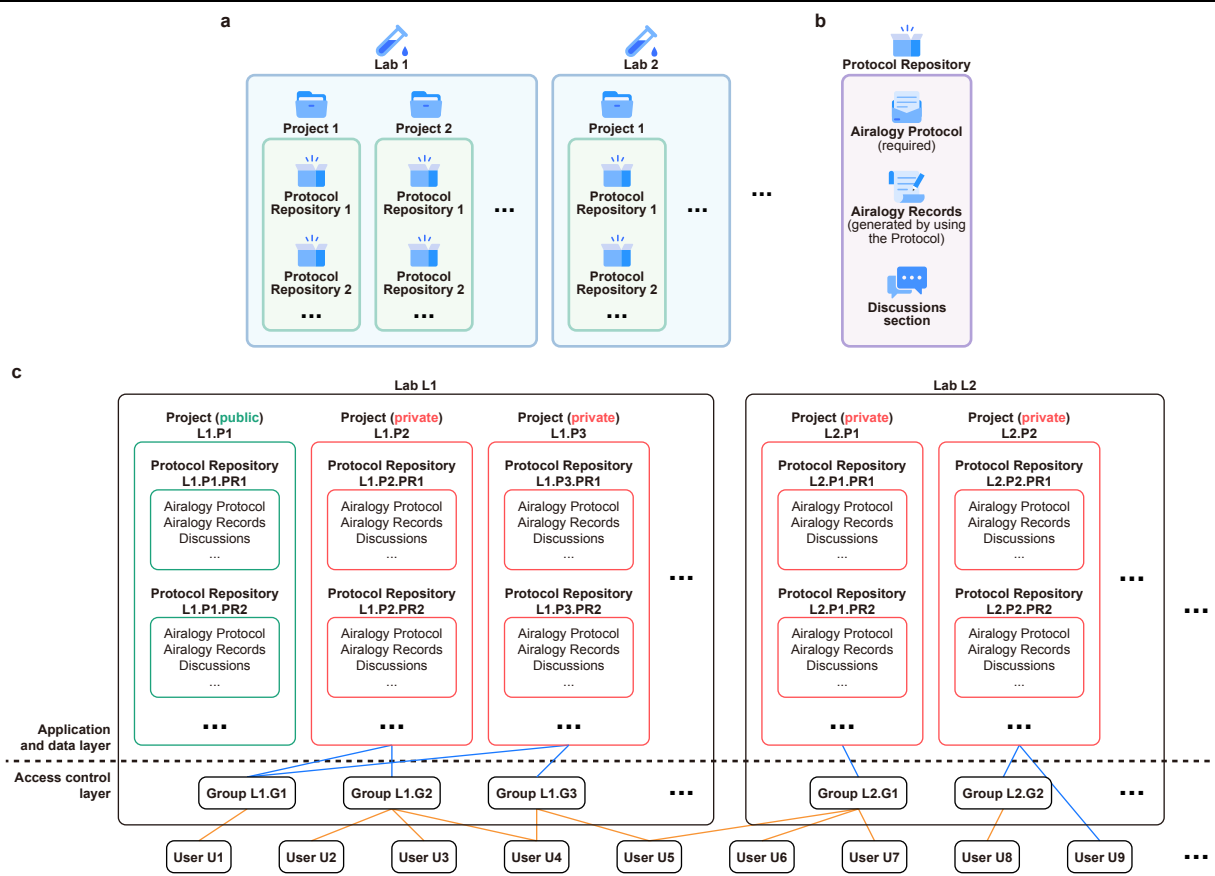

**Supplementary Fig. S37: Hierarchical structure and access control in the Airalogy Platform.**
**a,** The Airalogy Platform employs a three-tier organizational structure—Lab → Project → Protocol—where each Lab corresponds to a real or virtual laboratory. Within a Lab, users can create multiple Projects to manage different ongoing research endeavors. Each Project may contain one or more Protocol Repositories. **b,** A Protocol Repository serves as both a data repository and a workspace for running an Airalogy Protocol. Because an Airalogy Protocol can be viewed as a code package, it requires a suitable runtime environment. This environment is provided by the Protocol Repository, which allows the Protocol to be executed and generates Airalogy Records based on the Protocol's specifications. A single Protocol can be reused across multiple experimental iterations, producing multiple Airalogy Records that are stored and managed within this Repository. Each Protocol Repository also includes a Discussions section, enabling users to discuss the Protocol and share experiences, thus promoting community-driven knowledge exchange. **c,** Access control in the Airalogy Platform primarily involves regulating the accessibility of Projects. A single Lab can host any number of public or private Projects. For public Projects, all users on the Platform can access the Project and all associated Protocol Repositories (including their Protocols, Records, Discussions, and other content). For instance, L1.P1 is a public Project, so all users can access it and its Protocol Repositories (e.g., L1.P1.PR1, L1.P1.PR2). In contrast, private Projects restrict access to specified users or groups with explicit permissions. The Platform offers fine-grained access control for private Projects, including: 1. Granting access directly to certain users (e.g., L2.P2 directly grants user U9 access). 2. Granting access to specific groups (e.g., L1.P2 grants access to groups L1.G1 and L1.G2), allowing all users in these groups to access the Project. For example, since L1.G1 includes U1, and L1.G2 includes U2–4, users U1–4 can access L1.P2. Groups are created within each Lab, and Lab administrators can freely create any number of groups. Each group can contain any number of users, and a user may be included in multiple groups, whether in the same Lab (e.g., U4 belongs to L1.G2 and L1.G3) or different Labs (e.g., U5 belongs to L1.G3 and L2.G1).

## S2.5.2 Sharing expert experience through Discussions section on the Airalogy Platform

When different users record data using the same Airalogy Protocol, they may encounter unique issues not addressed by the Protocol and seek expert advice or discussion with other users; additionally, certain data within the recorded Airalogy Records may prompt further discussion. At the same time, experienced users might accumulate unique insights while using a Protocol and wish to share their valuable expertise with others. To support this, the Platform provides a Discussions section for each Protocol Repository (**Supplementary Figs. 37b, 38, 39**), where users can ask questions, seek solutions, or share their experiences in a self-Q&A format. This design not only fosters the sharing and dissemination of knowledge but also provides a platform for discussion and learning, thereby enhancing the practicality of the Platform and promoting community engagement.

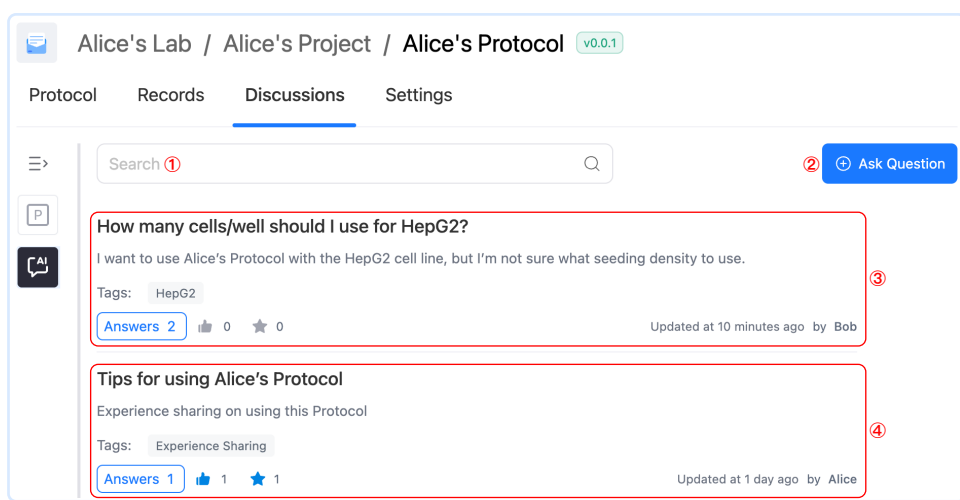

**Supplementary Fig. S38: Discussions section for an Airalogy Protocol.**
When a user encounters an issue that is not addressed in the Airalogy Protocol itself, they can turn to the Discussions section. The search bar (①) allows users to check quickly whether a similar question has already been asked and answered; if no relevant thread exists, they can post a new question at any time (②). Existing posts appear in a list. Questions may follow a standard Q&A format—e.g., ③ shows a question posted by Bob that has received two answers from other users (**Supplementary Fig. S39**)—or a self-Q&A format for sharing expertise, as in ④, where Alice, the Protocol's designer, posted and answered her own tip to share best practices.

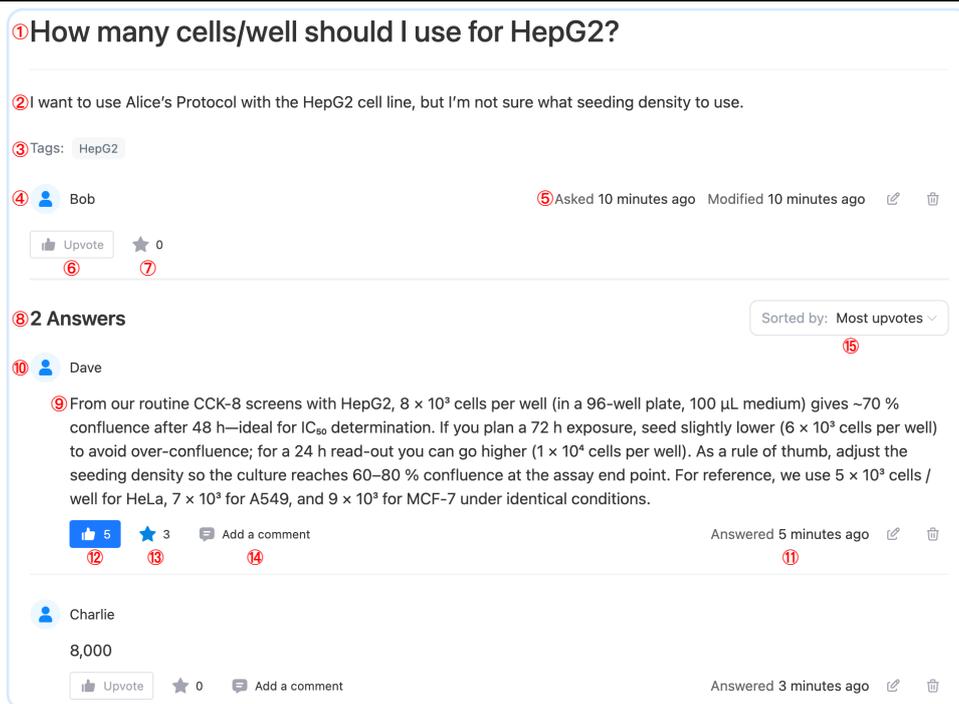

**Supplementary Fig. S39: Example Q&A thread in the Discussions section.**

A question is displayed with a title (①) and a description (②) that further clarifies the issue; tags (③) can be added for easier indexing. The interface also shows the asker (④) and the posting time (⑤). Community members can upvote (⑥) or bookmark (⑦) the question, promoting high-quality contributions and easy reference. Answers (⑧) appear sequentially, each showing its content (⑨) along with the responder's name (⑩) and the answer time (⑪). Users can upvote (⑫), bookmark (⑬), or comment (⑭) on each answer. Answers can be sorted, for example by "Most upvotes" (⑮); in this illustration, Charlie's answer was posted later than Dave's, yet Dave's appears first because it received more upvotes. Users may post new answers at any time (not shown).

In fact, both the Airalogy Protocol and the Discussions capture expert knowledge from the community; however, they differ mainly in how they are updated. Updating an Airalogy Protocol requires replacing the entire Protocol package, whereas Discussions can be updated dynamically in real time on the Platform without any modifications to the Protocol code. Thus, the Discussions section can be used to capture emerging insights and best practices, which maintainers can periodically fold into new Protocol releases. As usage grows, firsthand experience deepens, leading to ever-refined versions that may ultimately mature into widely adopted SOPs.

### S2.5.3 Sharing and reusing Protocol Repositories across Labs and Projects on the Airalogy Platform

To realize Airalogy Protocol's "design once, reuse globally" objective, the Platform incorporates an Airalogy Protocol Repository Reuse feature that enables one-click sharing and deployment of Protocol Repositories across Labs and Projects. This Reuse mechanism is divided into two main scenarios:

1.  **Reuse of Protocol Repositories from public Projects:** On the Airalogy Platform, any Protocol Repository within a public Project can be reused in other Projects, irrespective of whether the target Project belongs to the same Lab or is public or private. This design greatly facilitates global sharing and accelerates the dissemination and application of research outputs. In this context, the visibility of data and information post-reuse depends on the direction of reuse: when reusing from a public Project to another public Project, the downstream Project's referenced Protocol Repository data remains visible to the upstream Project; conversely, when reusing from a public Project to a private Project, the downstream Project's Protocol Repository data remain hidden from the upstream Project.

2.  **Reuse of Protocol Repositories from private Projects:** By default, Protocol Repositories from private Projects are not reusable. However, within a single Lab, there is often a need to share some Airalogy Protocols internally without exposing them externally. To address this, the Platform offers an inter-Project access authorization mechanism, whereby Lab administrators can grant permission for private Projects to access Protocol Repositories from other private Projects. Under this mechanism, a private Project can reuse Protocol Repositories from either public Projects or other authorized private Projects, while ensuring that the reused Protocol Repository data remain concealed from the originating (upstream) Projects. To streamline internal sharing, each lab is automatically provisioned with a dedicated private Project—"Lab Protocols" (Project ID: `lab_protocols`)—which serves as an internal hub for Protocol sharing. Lab members can create Protocol Repositories within this Project and share them with other private Projects within the Lab, thereby facilitating internal resource reuse without compromising data confidentiality.

Below, we outline how different components of a Protocol Repository (including the Airalogy Protocol, Records, and Discussions) are handled when a Protocol Repository is reused, taking into account the distinct characteristics of each component (**Supplementary Fig. S40a**):

*   **Airalogy Protocol:** Since an Airalogy Protocol essentially organizes like a code package, we use a fork mechanism, commonly employed in open-source collaborations. This mechanism creates a copy of the upstream Airalogy Protocol in the downstream Protocol Repository, ensuring that—even if the upstream Protocol Repository is deleted—the corresponding Protocol in the downstream Protocol Repository remains functional.

*   **Airalogy Records:** Airalogy Records are research data generated through the Airalogy Protocol, representing the outcome of applying the Protocol. On the Airalogy Platform, data is regarded as an asset of the Lab/Project in which it was recorded. When a Protocol Repository is reused from an upstream public Project to a downstream public or private Project, the upstream Records are not duplicated; however, each Airalogy Record on the Platform has a unique ID. Thus, the downstream user can still access those public Airalogy Records from the upstream Protocol Repository by referencing their IDs.

*   **Discussions:** On the Airalogy Platform, each Protocol Repository has its own, independent Discussions section. However, the Platform encourages broad dissemination of scientific knowledge across Labs and Projects. Therefore, if a user in the downstream Protocol Repository encounters a problem when reusing the upstream Protocol Repository and seeks an answer from

upstream maintainers, they can directly post a question in the upstream Protocol Repository's Discussions. Because the question is posed in the upstream Protocol Repository's Discussions, the upstream maintainers can easily notice and respond. Moreover, if the upstream Protocol Repository is public, these Q&As are viewable by other potential downstream users who might want to reuse the same Protocol Repository. Conversely, if a user in the downstream Protocol Repository prefers to discuss only with collaborators who have access to that private downstream Protocol Repository, they can post in the downstream Discussions instead; in that case, only members who can access the private Protocol Repository will be able to view and respond.

Since an upstream Protocol Repository may be applied to multiple downstream Projects, a radial Protocol network can form around a central upstream Protocol Repository (**Supplementary Fig. S40b**). The Airalogy Platform automatically tracks these upstream-to-downstream reuse relationships and provides built-in search capabilities to allow users to explore all public Discussions within the reuse network of any given Protocol Repository. As upstream and downstream Protocol Repositories continue to be applied, community members collectively identify and address questions about the Protocol within the Protocol Repository. Over time, these Discussions accumulate and contribute to continuous optimization of the Protocol, enhancing its overall quality. Through this search feature, valuable insights from multiple sources can be easily discovered and shared across the research community, enabling both broad dissemination and collaborative refinement.

In summary, this comprehensive Reuse framework enables efficient, global sharing of research resources while maintaining robust data privacy and security across diverse collaborative settings.

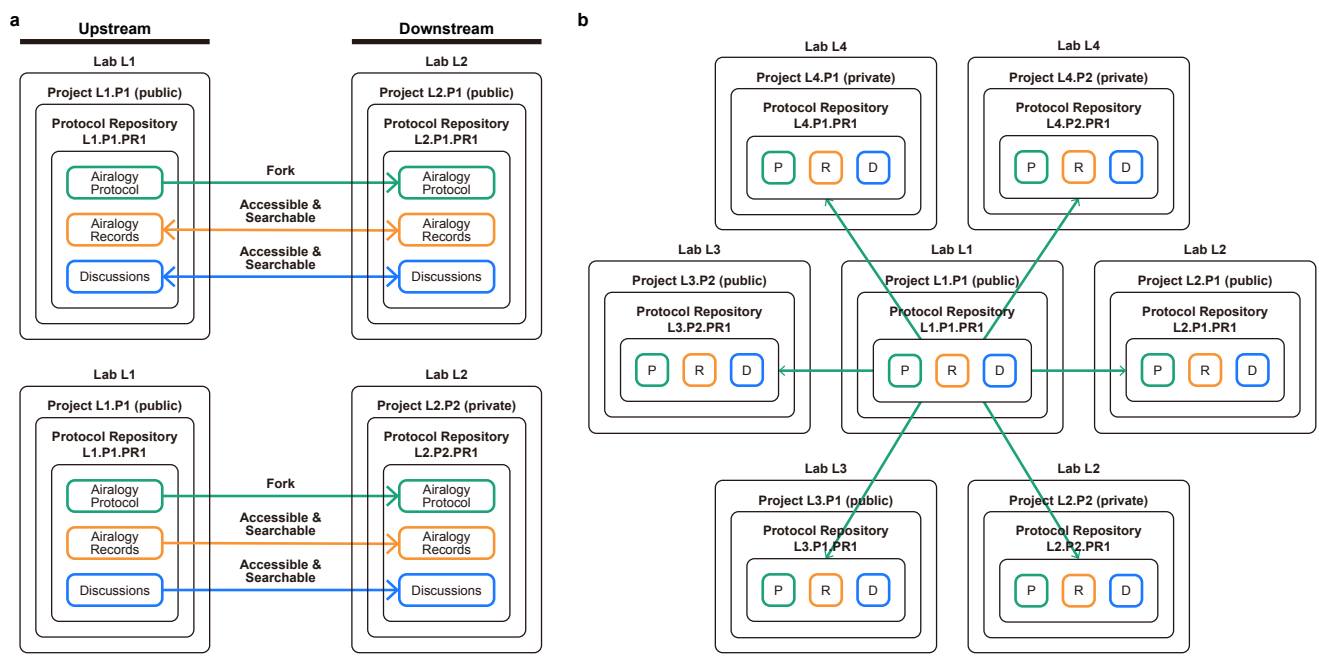

**Supplementary Fig. S40: Airalogy Platform enables Protocol Repository sharing and reuse across Labs and Projects.**

**a,** An upstream Protocol Repository can be reused in Projects of other Labs with the same level of accessibility (top, e.g., public Project → public Project) or a lower level of accessibility (bottom, e.g., public Project → private Project). Different mechanisms are employed for reusing the Airalogy Protocol (fork), Records (access and search permissions), and Discussions (access and search permissions). Bidirectional arrows indicate mutual accessibility and search capability, while unidirectional arrows indicate one-way access and search. **b,** A single Protocol Repository can be reused by multiple downstream Projects across different Labs, forming a Protocol network.

### S2.5.4 Airalogy Hub

The Airalogy Hub is the central data resource hub of the Airalogy Platform, allowing users to easily search for publicly accessible Airalogy Protocols, as well as to conveniently reuse them (**Supplementary Fig. S41**).

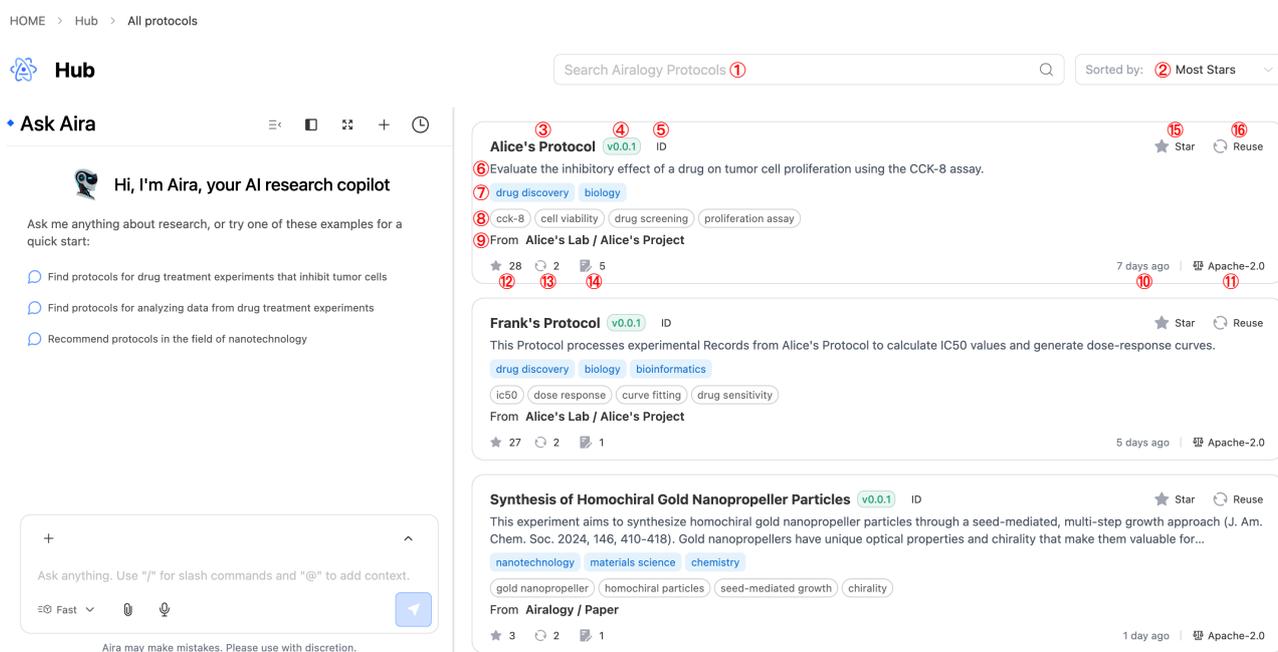

**Supplementary Fig. S41: The interface of the Airalogy Hub.**
Within the Airalogy Hub, users can enter search terms (①) to quickly locate specific Airalogy Protocols and reorder the results with the sorting menu (②) (e.g., by most stars). Protocols are displayed as list cards, each showing the Protocol name (③), version (④), ID (⑤), description (⑥), disciplines (⑦), keywords (⑧), source Project (⑨), last-updated time (⑩), and license (⑪). Clicking a card opens the Protocol's detail page. Additional metrics—stars (⑫), reuse count (⑬), and the number of Records generated with the Protocol (⑭)—help users gauge popularity and community adoption (more stars or reuses generally indicate higher practical value). With one click, users can "Star" (⑮) a Protocol for their personal collection or "Reuse" (⑯) it in their own Project. The Hub's left-hand panel hosts Aira, the Platform's AI research copilot, enabling conversational search and Protocol recommendations. Aira's responses embed summary cards for the suggested Protocols so that users can quickly evaluate, reuse, or navigate to the corresponding repository (not shown).

## S2.5.5 Airalogy Protocol Editor

In addition to uploading a locally edited Airalogy Protocol directly to the Airalogy Platform, users can conveniently create, edit, and modify Protocols online via the built-in Airalogy Protocol Editor (**Supplementary Fig. S42**). This Editor supports source-code editing, real-time preview of the Protocol's contents, and deep integration with the Platform's AI research copilot, Aira, enabling chat-based explanations of the Protocol's syntax, guided revisions, and automatic generation of Protocol files.

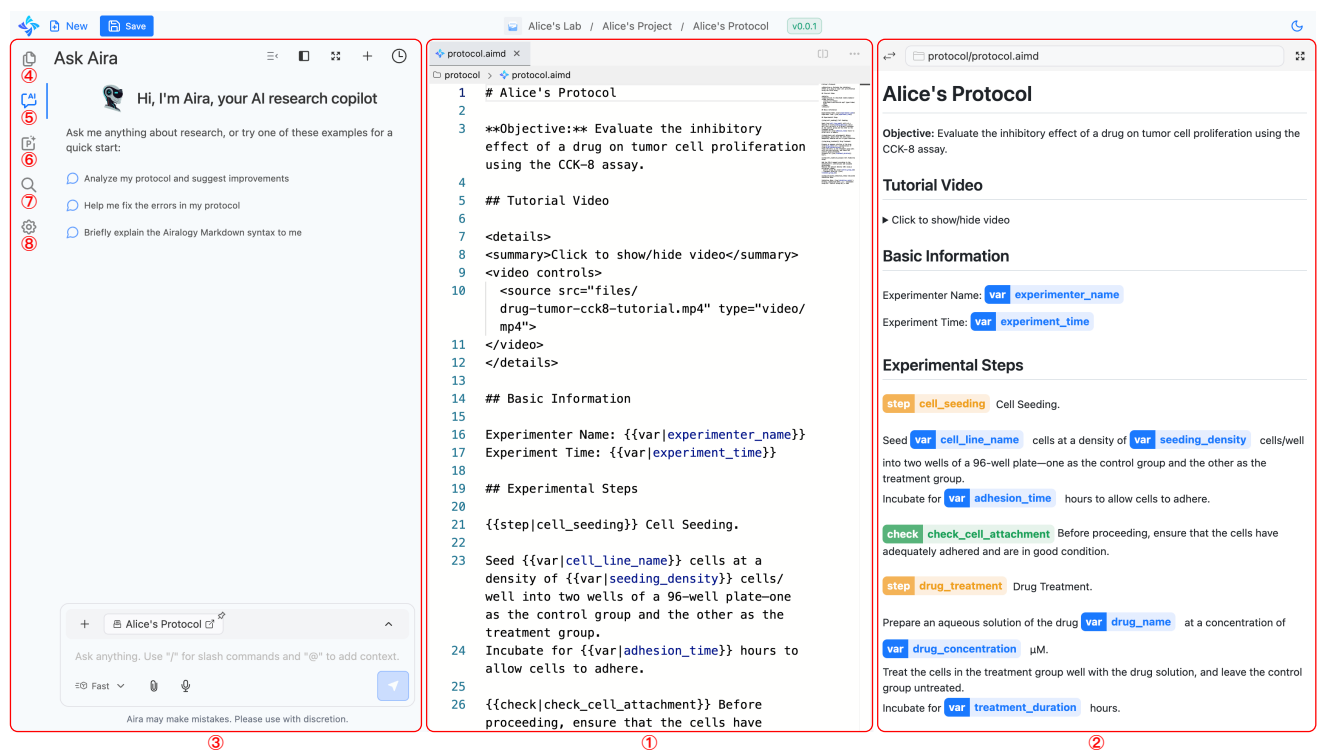

**Supplementary Fig. S42: The interface of the Airalogy Protocol Editor.**
The Editor features a source-code pane (①) where users can modify every file that makes up an Airalogy Protocol—such as `protocol.aimd`, `model.py`, and `assigner.py`. A built-in Airalogy Markdown parser and renderer instantly transform the `protocol.aimd` source into stylized Markdown (②): for example, headings appear in appropriately scaled fonts, and each Airalogy Field (e.g., var, step, check) is highlighted in a distinct color scheme. The tool sidebar (③) offers several practical panels. A "Files" panel (④) shows the Protocol's directory tree, much like an integrated development environment (IDE). An "AI Chat" panel (⑤) opens a conversation window with Aira, allowing users to ask about Protocol content, syntax, or request edits; because Aira understands the latest Airalogy Protocol syntax and can read the files open in pane (①), it delivers precise guidance and revision suggestions. Users can also employ the "AI-driven Protocol Auto-generation" panel (⑥): by supplying reference protocols or generation prompts, Aira automatically creates `protocol.aimd`, `model.py`, and `assigner.py`. A "Search" panel (⑦) enables rapid text queries across all Protocol files, while a "Settings" panel (⑧) lets users view or edit key metadata (mirroring the contents of `protocol.toml`), such as protocol ID, version, name, and description.

## S2.6 Airalogy Masterbrain

Airalogy Masterbrain (personified as "Aira") serves as the AI Masterbrain[13] for the Airalogy Platform. Because the Platform is community-driven, its content and data are updated in real time. Consequently, continually retraining the LLM to ensure the AI has access to the latest Platform knowledge is not feasible. Moreover, a unique aspect of deploying AI within Airalogy is that each AI application is closely tied to a specific research context. Therefore, Aira must automatically tailor its functionality by conditionally injecting relevant contextual information based on the scenario.

To address these challenges, we have designed an AI architecture based on context injection that flexibly and scalably integrates various external tools and services (**Supplementary Fig. S43**). This architecture not only accommodates the frequent updates in data and content within the Platform but also provides targeted intelligent support tailored to different research scenarios, thereby enhancing the efficiency and accuracy of research activities.

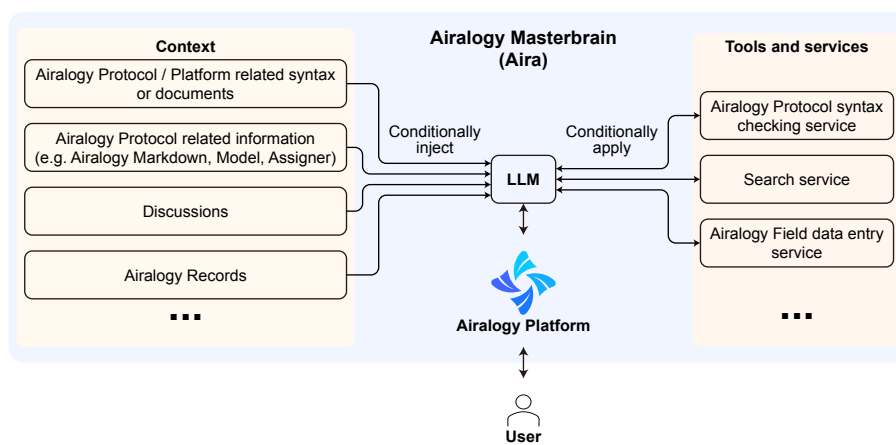

**Supplementary Fig. S43: Architecture of the Airalogy Masterbrain.**
Airalogy Masterbrain (Aira) is built around a Large Language Lodel (LLM). In different application scenarios, Aira conditionally injects relevant context into the LLM and can conditionally utilize external tools or services required to accomplish tasks specific to the given research context.

### S2.6.1 AI-automated Airalogy Protocol generation and AI-assisted authoring

To help users quickly customize an Airalogy Protocol and lower the learning curve, Aira provides an AI-automated Airalogy Protocol generation feature.

As previously mentioned, an Airalogy Protocol comprises three main components—Markdown, Model, and Assigner—which correspond to three files: `protocol.aimd`, `model.py`, and `assigner.py`. The goal of the AI-automated Airalogy Package design is to automatically generate the content for these three files.

Since Airalogy is an actively developed framework and its Protocol syntax is continuously evolving, we do not adopt a fine-tuning approach. Instead, we employ the strategy illustrated in **Supplementary Fig. S44**. Under the Aira architecture, when a user wishes to convert an existing protocol document (for example, a protocol description in the methods section of a paper) into an Airalogy Protocol, we leverage the significantly enhanced context length of LLMs by first injecting the Airalogy Protocol syntax as contextual information into the conversation, ensuring that the LLM understands the syntax.

Typically, when writing an Airalogy Protocol manually, one first writes `protocol.aimd`, then `model.py`, and finally `assigner.py`; our AI-automated generation method follows the same sequence. By using a few-shot learning strategy, we provide the LLM with one or more examples of converting a protocol into an Airalogy Protocol. The user then supplies the protocol document to be converted, and the LLM first automatically generates the `protocol.aimd` file. Next, `model.py` is generated based on `protocol.aimd`, and finally, `assigner.py` is produced based on both `protocol.aimd` and `model.py`. Notably, since `model.py` and `assigner.py` are optional components in an Airalogy Protocol, the generated files may be empty if the content of the Protocol does not require these sections.

Moreover, thanks to the explicitness of the Airalogy Protocol syntax, Aira can employ an Airalogy Protocol Syntax Checker during the generation process. This checker automatically verifies the correctness of the syntax produced by the LLM. If errors are detected, the checker provides feedback, prompting the LLM to automatically correct the generated content.

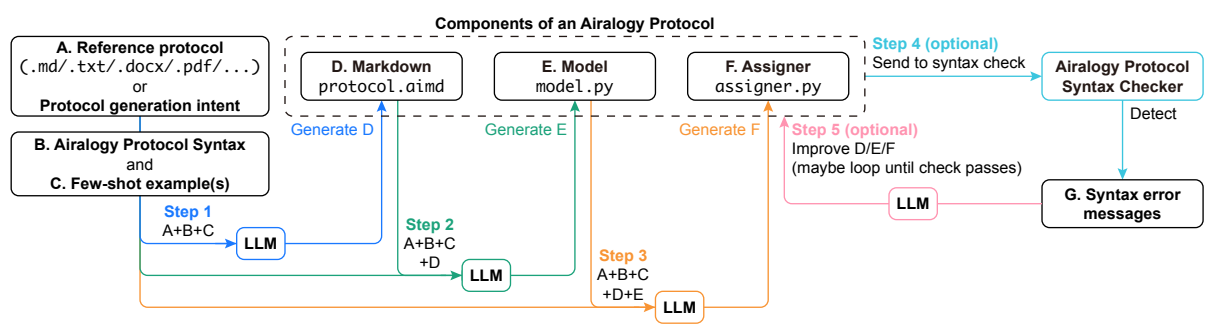

**Supplementary Fig. S44: Method for Aira's AI-automated generation of Airalogy Protocols.**
The AI-automated Airalogy Protocol generation process leverages a reference protocol—or aligns with the user's Protocol generation intent—to create the initial files that form an Airalogy Protocol (i.e., `protocol.aimd`, `model.py`, and `assigner.py`). The process is divided into sequential steps: the LLM first generates the `protocol.aimd` file, then the `model.py` file, and finally the `assigner.py` file. The generated files are subsequently verified for syntax errors using the built-in Airalogy Protocol Syntax Checker (optional), with any detected errors fed back to the LLM for automatic refinement.

After Aira generates an initial, AI-designed Airalogy Protocol, it also provides an AI-assisted, chat-based explanation and revision feature (**Supplementary Fig. S42** ⑤) to help users quickly understand and modify the generated content. When a user requests an explanation or suggests a change to a part of the Protocol, Aira automatically generates the corresponding explanation or revision proposal based on the Airalogy Protocol syntax. This method supports multi-turn dialogue, allowing interactive communication with the AI. Such conversational explanation and editing further lower the barrier to using Airalogy Protocol.

For example, regarding Alice's Protocol shown in **Fig. 2b**, a user might ask:

> What does {{var|cell_line_name}} mean?

Aira would respond:

> {{var|cell_line_name}} is part of the Airalogy Markdown syntax, representing an Airalogy Field with the ID cell_line_name, categorized as a Variable (indicated by var).

Similarly, a user may request modifications to an existing Airalogy Protocol. For instance, for Alice's Protocol in **Fig. 2b**, a user might state:

> I would like to add a final step in the protocol reminding the user to clean up all experimental materials after the experiment.

Aira would then reply:

> Sure, you can add the following code at the end of the protocol.aimd file:
>
> {{step|cleanup, 1}} Clean up all experimental materials after the experiment.

The Airalogy Platform offers an online visual editor for Airalogy Protocols (**Supplementary Fig. S 42**), enabling users to conveniently leverage these AI-generated Protocol and dialogue functionalities.

### S2.6.2 Chat-based Airalogy Protocol search

To enable users to search for Airalogy Protocols more easily and efficiently, Aira provides a chat-based search feature powered by an LLM (**Supplementary Fig. S45**). Users simply express their search intent via the chat interface, and Aira automatically conducts a matching search within the Airalogy Hub (**Supplementary Fig. S41**). The search results are then relayed to the user conversationally. Notably, we have enhanced the Chat Interface for this functionality so that the retrieved Protocols are intuitively displayed alongside Aira's responses. Users can click on the summary cards of these Protocols to quickly navigate to the corresponding Protocol.

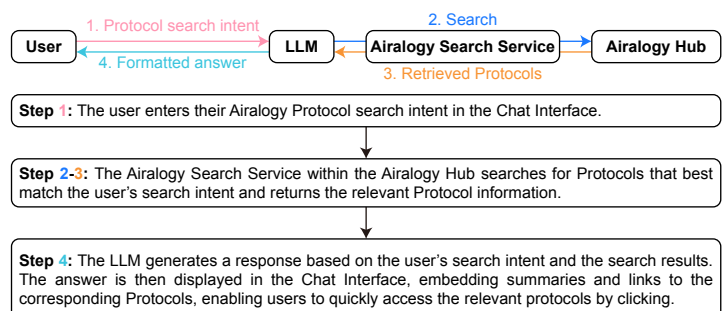

**Supplementary Fig. S45: Method for searching Airalogy Protocols in the Airalogy Hub through chat.**

### S2.6.3 AI chat for Airalogy Protocols

To help users better understand and apply an Airalogy Protocol, Aira offers an AI-driven dialogue feature specifically designed for Airalogy Protocols. Unlike general-purpose chatbots (such as ChatGPT), Aira exhibits two unique characteristics when responding to user queries (**Supplementary Fig. S46**):

● **Feature 1:** Aira fully leverages the contextual information associated with the Airalogy Protocol during the conversation.
● **Feature 2:** Aira automatically searches for and incorporates relevant knowledge from Discussions related to the Airalogy Protocol—or from the entire Airalogy Platform—to enrich the dialogue.

These features enable Aira to provide accurate responses tailored to the specific research scenarios represented by the Airalogy Protocol, thereby enhancing users' ability to resolve practical research issues and improving the overall intelligent support for research data recording.

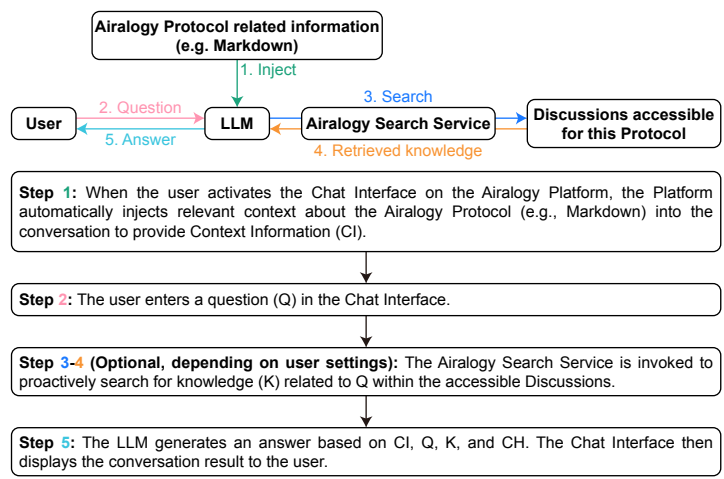

**Supplementary Fig. S46: Method for chatting based on Airalogy Protocol information.**
Users can choose how much Discussion content to incorporate—for example, they may (1) ignore Discussions altogether; (2) draw only on the Discussions in the Protocol Repository that contains the Protocol; (3) include Discussions from other Protocol Repositories within the same Lab/Project; or (4) leverage Discussions from every other public Protocol Repository on the Airalogy Platform.

### S2.6.4 AI-automated data entry based on Airalogy Protocol

Although the Airalogy Protocol Recording Interface provides an intuitive WYSIWYG method for data entry, in real experimental settings users may not always have immediate computer access to the Airalogy Platform. In such cases, data might first be recorded on alternative media (e.g., handwritten notes). Moreover, some laboratories may already maintain historical records using tools such as Microsoft Word. To enable users to efficiently and automatically convert these alternative or historical records into standardized Airalogy Records, Aira offers an AI-driven data entry function.

Specifically, when an Airalogy Protocol is specified, users can send multimodal information (e.g., plain text, images, audio, or files such as TXT, Markdown, Word, or PDF) via the Chat Interface. Aira automatically interprets the multimodal content and, based on the Airalogy Protocol, extracts the values corresponding to the Airalogy Fields defined therein. Furthermore, Aira can control the Airalogy Protocol Recording Interface to automatically populate the appropriate input boxes with these extracted values. In this way, Aira efficiently converts unstructured multimodal information into structured, standardized Airalogy Records (**Supplementary Fig. S47**).

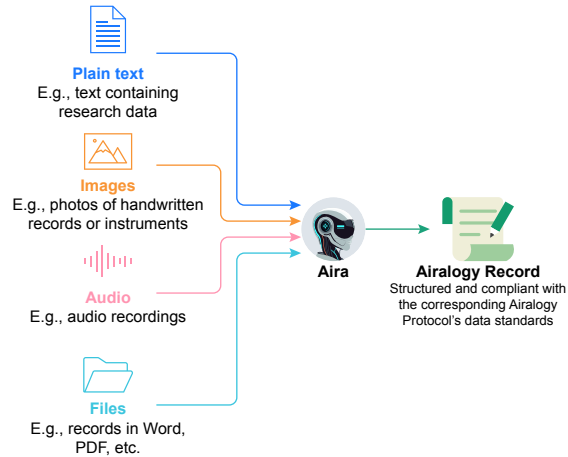

**Supplementary Fig. S47: Aira automatically extracts research data from the multimodal information provided by users and records it as a structured and standardized Airalogy Record that conforms to the corresponding Airalogy Protocol data standards.**

To enable automated data entry, we have designed a method that coordinates user input, the AI system (Aira), and the Airalogy Protocol Recording Interface (**Supplementary Fig. S48; Supplementary Code S19**):

- **Step 1:** The user inputs information in multimodal forms (such as text, voice, or images) through the Chat Interface.
- **Step 2:** The Chat Interface sends this information to Aira.
- **Step 3:** Using the current Airalogy Protocol and its corresponding Airalogy Protocol Field JSON Schema, Aira automatically extracts the relevant Airalogy Field data from the user's multimodal input and converts it into a structured list of operation instructions.
- **Step 4:** Aira sends the structured list of operation instructions to the Airalogy Protocol Recording Interface.
- **Step 5a:** The Recording Interface processes each instruction sequentially, generating an acknowledgment for every operation and ultimately creating an acknowledgment list.
- **Step 5b:** Whenever an instruction is successfully processed, the Recording Interface instantly populates the corresponding Airalogy Field Input Box with the value specified by that instruction and visually displays this filling process to the user.
- **Step 6a:** The Recording Interface sends the acknowledgment list back to Aira.
- **Step 6b:** Aira compiles a final response for the user by combining (a) the user's original input, (b) the structured list of operation instructions, and (c) the acknowledgment list from the Recording Interface.
- **Step 6c:** Aira sends this response to the Chat Interface.
- **Step 6d:** The Chat Interface displays the response to the user.

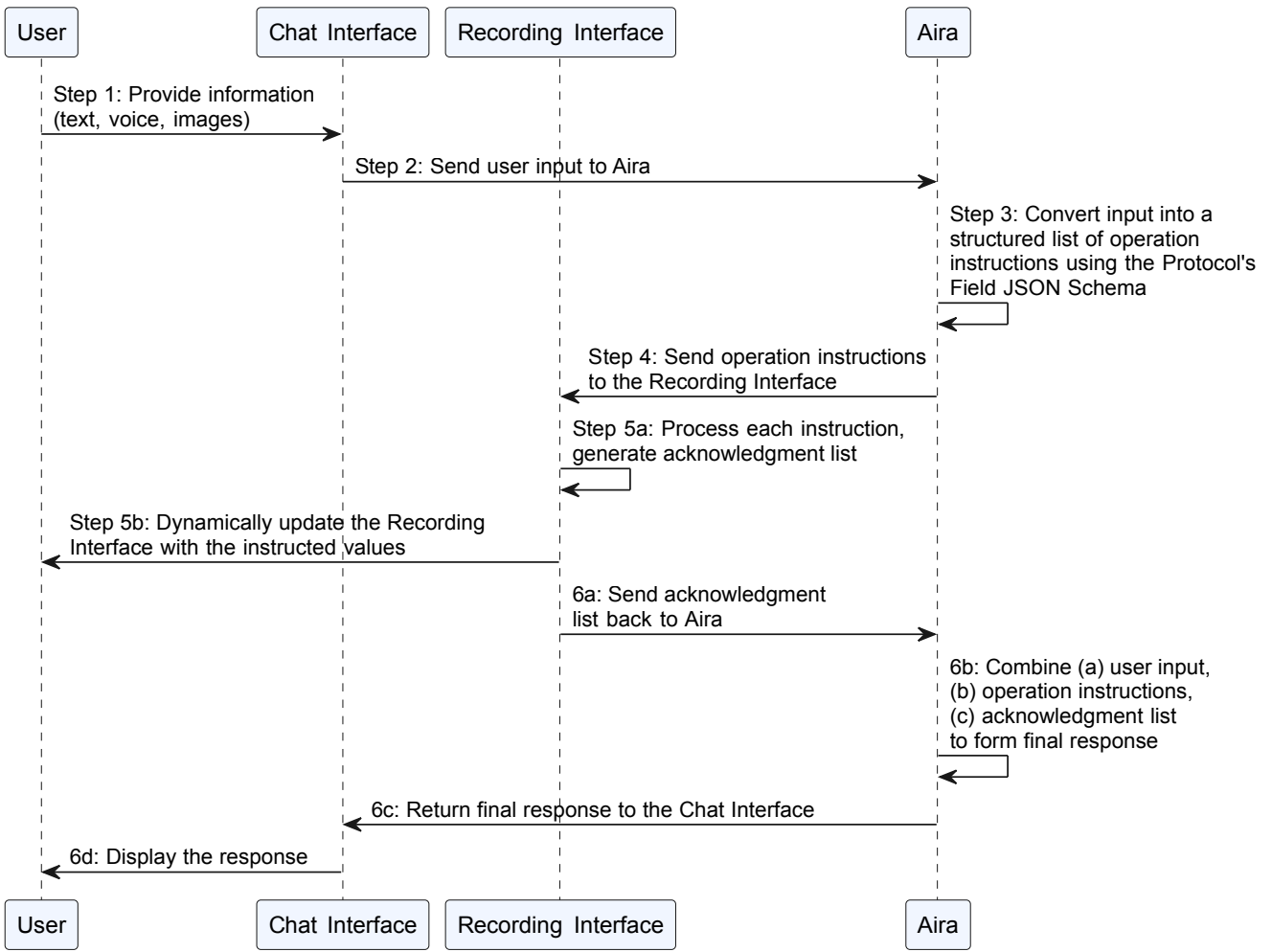

**Supplementary Fig. S48:** Aira's method for automatically extracting research data from the multimodal information provided by users and, in accordance with Airalogy Protocol data standards, recording it to the Airalogy Protocol Recording Interface to generate a structured Airalogy Record.

**Input information (from user)**

```
Using cell line HeLa, with a cell seeding density of 5000.
```

**Structured operation list (generated by Aira)**

```json
[
    {
        "operation": "update",
        "field_id": "cell_line_name",
        "field_value": "HeLa"
    },
    {
        "operation": "update",
        "field_id": "seeding_density",
        "field_value": 5000
```

```
    }
]
```

**Acknowledgement list (generated by Airalogy Protocol Recording Interface)**

```json
[
    {
        "success": true,
        "field_id": "cell_line_name",
        "field_value_updated": "HeLa",
        "message": "The value of cell_line_name has been set to HeLa."
    },
    {
        "success": true,
        "field_id": "seeding_density",
        "field_value_updated": 5000,
        "message": "The value of seeding_density has been set to 5000."
    }
]
```

**Final response (generated by Aira)**

```
The value of cell_line_name has been set to HeLa. The value of seeding_density has been set
to 5000.
```

**Supplementary Code S19: An example of user input information, along with the corresponding structured operation list generated by Aira, the acknowledgement list produced by the Airalogy Protocol Recording Interface after execution, and the final response generated by Aira.**

The final response can be produced either by Aira through a second LLM call or, more simply, by the Recording Interface itself via a rule-based method—for example, by concatenating the `message` fields in the acknowledgement list.

## S2.6.5 AI-assisted Airalogy Record analysis

Because each Airalogy Record is generated under a specific Airalogy Protocol but stored separately, this design confers a significant advantage for data analysis. Suppose we have an Airalogy Protocol $AP_x$ that has produced $n$ Airalogy Records, denoted as:

$$[AR_1^{AP_x}, AR_2^{AP_x}, \cdots, AR_n^{AP_x}] \tag{S30}$$

As described in **Supplementary Method S2.4.3**, generating a complete Airalogy Record Report for a given Airalogy Record requires only two additional files beyond the Record itself: the Airalogy Protocol Markdown and the Airalogy Protocol Field JSON Schema. Hence, once these two files are provided, the AI has the primary information needed to interpret that Record.

If we wish to have an LLM analyze $n$ Airalogy Records and produce an AI Analysis Result (AIAR), and all these Records originate from the same Airalogy Protocol with a consistent data structure and specifications, it suffices to supply the Markdown $M^{AP_x}$ and Field JSON Schema $FJS^{AP_x}$ just once. Formally, this can be expressed as follows:

$$AIAR = AI([AR_1^{AP_x}, AR_2^{AP_x}, \cdots, AR_n^{AP_x}], M^{AP_x}, FJS^{AP_x}) \tag{S31}$$

This approach avoids having to repeatedly provide $M^{AP_x}$ and $FJS^{AP_x}$ for each Record, as would be required by:

$$AIAR = AI([AR_1^{AP_x}, M^{AP_x}, FJS^{AP_x}], [AR_2^{AP_x}, M^{AP_x}, FJS^{AP_x}], \cdots, [AR_n^{AP_x}, M^{AP_x}, FJS^{AP_x}]) \tag{S32}$$

Clearly, the method in **Supplementary Eq. S31** is more efficient than that in **Supplementary Eq. S32**, eliminating $n - 1$ redundant injections of $M^{AP_x}$ and $FJS^{AP_x}$ without losing any essential information.

In scenarios where we want the AI to fully comprehend every component of the Airalogy Protocol—including the Model Model$^{AP_x}$ and Assigner Model$^{AP_x}$—we can also include these modules. In that case, the AI analysis result formula extends to:

$$AIAR = AI([AR_1^{AP_x}, AR_2^{AP_x}, \cdots, AR_n^{AP_x}], M^{AP_x}, FJS^{AP_x}, Model^{AP_x}, Assigner^{AP_x}) \tag{S33}$$

This configuration allows the AI to leverage the entire Protocol definition for more comprehensive and context-aware analyses.

Further extending the approach, suppose we have $m$ Airalogy Protocols, denoted as $[AP_{x_1}, \cdots, AP_{x_i}, \cdots, AP_{x_m}]$, and that each Protocol has generated $[n_1, \cdots, n_i, \cdots, n_m]$ Airalogy Records. The collection of these Records can be represented as:

$$\begin{aligned}
&[[AR_1^{AP_{x_1}}, AR_2^{AP_{x_1}}, \cdots, AR_n^{AP_{x_1}}], \\
&\cdots, \\
&[AR_1^{AP_{x_i}}, AR_2^{AP_{x_i}}, \cdots, AR_n^{AP_{x_i}}], \\
&\cdots, \\
&[AR_1^{AP_{x_m}}, AR_2^{AP_{x_m}}, \cdots, AR_n^{AP_{x_m}}]]
\end{aligned} \tag{S34}$$

At this point, we can supply each Protocol's corresponding Markdown $M^{AP_{x_i}}$ and Field JSON Schema $FJS^{AP_{x_i}}$ as input for AI analysis (for brevity, we omit the explicit inclusion of Model and Assigner, which can be readily extended as in **Supplementary Eq. S33**). Then, **Supplementary Eq. S31** extends to:

$$
\begin{aligned}
\text{AIAR} = \text{AI}([ \\
[[\text{AR}_1^{\text{AP}_{x_1}}, \text{AR}_2^{\text{AP}_{x_1}}, \cdots, \text{AR}_n^{\text{AP}_{x_1}}], \text{M}^{\text{AP}_{x_1}}, \text{FJS}^{\text{AP}_{x_1}}], \\
\cdots, \\
[\text{AR}_1^{\text{AP}_{x_i}}, \text{AR}_2^{\text{AP}_{x_i}}, \cdots, \text{AR}_n^{\text{AP}_{x_i}}], \text{M}^{\text{AP}_{x_i}}, \text{FJS}^{\text{AP}_{x_i}}], \\
\cdots, \\
[\text{AR}_1^{\text{AP}_{x_m}}, \text{AR}_2^{\text{AP}_{x_m}}, \cdots, \text{AR}_n^{\text{AP}_{x_m}}], \text{M}^{\text{AP}_{x_m}}, \text{FJS}^{\text{AP}_{x_m}}] \\
])
\end{aligned}
\tag{S35}
$$

This can be simplified as:

$$
S_i = [\text{AR}_1^{\text{AP}_{x_i}}, \text{AR}_2^{\text{AP}_{x_i}}, \cdots, \text{AR}_n^{\text{AP}_{x_i}}], \text{M}^{\text{AP}_{x_i}}, \text{FJS}^{\text{AP}_{x_i}}
\tag{S36}
$$

$$
S = [S_1, \cdots, S_i, \cdots, S_m]
\tag{S37}
$$

$$
\text{AIAR} = \text{AI}(S)
\tag{S38}
$$

Furthermore, we can inject a specified analysis purpose (abbreviated as APU) into the AI for more targeted analysis. In that case, **Supplementary Eq. S38** extends to:

$$
\text{AIAR} = \text{AI}(S, \text{APU})
\tag{S39}
$$

Moreover, in practical analysis—such as in multi-turn AI dialogues—a user might provide a second-round analysis intent ($\text{APU}_2$) following the initial intent ($\text{APU}_1$). In this scenario, we can feed the first-round AI analysis result ($\text{AIAR}_1$) back into the system for deeper analysis to obtain a second-round result ($\text{AIAR}_2$). Thus, **Supplementary Eq. S39** can be further extended as:

$$
\text{AIAR}_1 = \text{AI}(S, \text{APU}_1)
$$
$$
\text{AIAR}_2 = \text{AI}(S, \text{APU}_2, [\text{APU}_1, \text{AIAR}_1])
\tag{S40}
$$

Furthermore, we can extend the analysis in **Supplementary Eq. S40** to an arbitrary *t*-round process:

$$
\text{AIAR}_1 = \text{AI}(S, \text{APU}_1, [])
$$
$$
\text{AIAR}_2 = \text{AI}\big(S, \text{APU}_2, [[\text{APU}_1, \text{AIAR}_1]]\big)
$$
$$
\cdots
$$
$$
\text{AIAR}_j = \text{AI}\big(S, \text{APU}_j, [[\text{APU}_1, \text{AIAR}_1], [\text{APU}_2, \text{AIAR}_2], \cdots, [\text{APU}_{j-1}, \text{AIAR}_{j-1}]]\big)
\tag{S41}
$$
$$
\cdots
$$
$$
\text{AIAR}_t = \text{AI}\big(S, \text{APU}_t, [[\text{APU}_1, \text{AIAR}_1], [\text{APU}_2, \text{AIAR}_2], \cdots, [\text{APU}_{t-1}, \text{AIAR}_{t-1}]]\big)
$$

Based on this analysis, we have developed an LLM-based method in Aira for AI analysis of Airalogy Records that supports multi-turn dialogues, as illustrated in **Supplementary Fig. S49**. The method can analyze multiple Records originating from a single Protocol or—just as readily—Records drawn from several different Protocols. Crucially, it is not limited to brief question-and-answer exchanges. In practice, researchers often pursue a specific research goal by running a series of experiments across multiple Protocols, thereby generating a heterogeneous set of Records. All of these Records, together with their source Protocols, can be supplied to Aira, which can then draft a full research report—or even a data-driven manuscript—based on the combined dataset. In this way, Aira functions as an AI scientist, capable of automatically composing comprehensive research narratives from the underlying research data.

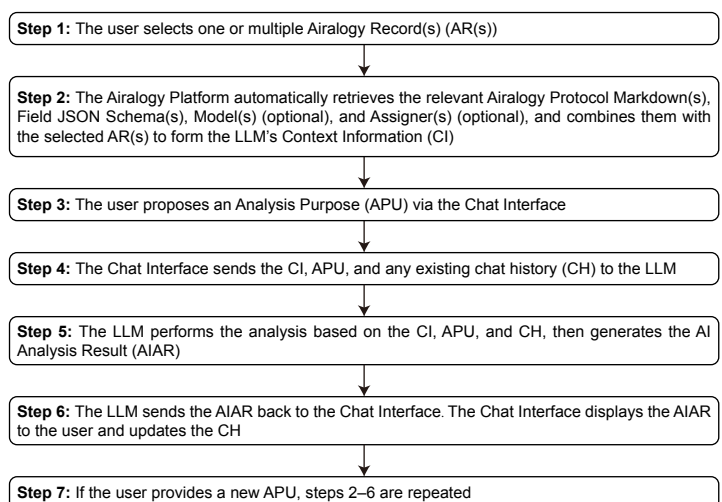

**Supplementary Fig. S49: Steps for AI-assisted Airalogy Record analysis in Aira.**

## S2.7 AI Research Automation method

The theoretical underpinnings of the AI Research Automation (AIRA) method, along with detailed application examples, are provided in **Supplementary Case Study S1**. Our Python implementation of the AIRA method is open-sourced on GitHub (https://github.com/airalogy/masterbrain) and published on PyPI (https://pypi.org/project/masterbrain).

# S3 Supplementary References

6        Boiko, D. A., MacKnight, R., Kline, B. & Gomes, G. Autonomous chemical research with large language models. *Nature* **624**, 570-578 (2023).

7        Szymanski, N. J. *et al.* An autonomous laboratory for the accelerated synthesis of novel materials. *Nature* **624**, 86-91 (2023).

13       Yang, Z., Wang, Y. & Zhang, L. AI becomes a masterbrain scientist. *bioRxiv*, 2023.2004.2019.537579 (2023).

14       Lu, C. *et al.* The ai scientist: Towards fully automated open-ended scientific discovery. *arXiv preprint arXiv:2408.06292* (2024).

16       Weng, Y. *et al.* Cycleresearcher: Improving automated research via automated review. *arXiv preprint arXiv:2411.00816* (2024).

17       Yamada, Y. *et al.* The AI Scientist-v2: Workshop-Level Automated Scientific Discovery via Agentic Tree Search. *arXiv preprint arXiv:2504.08066* (2025).

26       Zheng, Y. *et al.* Homochiral Nanopropeller via Chiral Active Surface Growth. *J. Am. Chem. Soc.* **146**, 410-418, doi:10.1021/jacs.3c09652 (2024).

29       Achiam, J. *et al.* Gpt-4 technical report. *arXiv preprint arXiv:2303.08774* (2023).

31       Guo, D. *et al.* Deepseek-r1: Incentivizing reasoning capability in llms via reinforcement learning. *arXiv preprint arXiv:2501.12948* (2025).

39       Baker, M. 1,500 scientists lift the lid on reproducibility. *Nature* **533**, 452-454, doi:10.1038/533452a (2016).

41       protocols.io. *protocols.io*, <https://www.protocols.io> (2025).

43       Zhu, M., Weng, Y., Yang, L. & Zhang, Y. in *ACL 2025.*

44       Zhang, X., Li, M., Xu, Y., Ren, J. & Zeng, A.-P. Quantitative study of H protein lipoylation of the glycine cleavage system and a strategy to increase its activity by co-expression of LplA. *Journal of biological engineering* **13**, 1-9 (2019).

45       Chen, L. & Zeng, A.-P. Rational design and metabolic analysis of Escherichia coli for effective production of L-tryptophan at high concentration. *Applied microbiology and biotechnology* **101**, 559-568 (2017).

46       Smith, A. An Inquiry into the Nature and Causes of the Wealth of Nations. *Readings in economic sociology*, 6-17 (2002).

51       Si, C., Yang, D. & Hashimoto, T. Can llms generate novel research ideas? a large-scale human study with 100+ nlp researchers. *arXiv preprint arXiv:2409.04109* (2024).

53       De Volder, M. F., Tawfick, S. H., Baughman, R. H. & Hart, A. J. Carbon nanotubes: present and future commercial applications. *science* **339**, 535-539 (2013).

54       Bergin, S. D. *et al.* Multicomponent solubility parameters for single-walled carbon nanotube−solvent mixtures. *ACS nano* **3**, 2340-2350 (2009).

55       Davis, V. A. *et al.* True solutions of single-walled carbon nanotubes for assembly into macroscopic materials. *Nature nanotechnology* **4**, 830-834 (2009).

56       Dai, T. *et al.* Autonomous mobile robots for exploratory synthetic chemistry. *Nature*, 1-8 (2024).

57       Zhang, P. *et al.* Scaling Laws of Scientific Discovery with AI and Robot Scientists. *arXiv preprint arXiv:2503.22444* (2025).

58       Bergman, R. G. & Danheiser, R. L. Reproducibility in Chemical Research. *Angewandte*

*Chemie International Edition* **55**, 12548-12549, doi:https://doi.org/10.1002/anie.201606591 (2016).

59  Ciriminna, R., Angellotti, G., Petri, G. L. & Pagliaro, M. Reproducibility in chemistry research. *Heliyon* **10**, e33658, doi:https://doi.org/10.1016/j.heliyon.2024.e33658 (2024).

60  Sciences, N. A. o. *et al. Reproducibility and replicability in science*. (National Academies Press, 2019).

61  Scarabelli, L., Sanchez-Iglesias, A., Perez-Juste, J. & Liz-Marzan, L. M. A "Tips and Tricks" Practical Guide to the Synthesis of Gold Nanorods. *J. Phys. Chem. Lett.* **6**, 4270-4279, doi:10.1021/acs.jpclett.5b02123 (2015).

62  Wang, Y., He, J., Liu, C., Chong, W. H. & Chen, H. Thermodynamics versus kinetics in nanosynthesis. *Angew. Chem. Int. Ed.* **54**, 2022-2051, doi:10.1002/anie.201402986 (2015).

63  Personick, M. L. & Mirkin, C. A. Making sense of the mayhem behind shape control in the synthesis of gold nanoparticles. *J. Am. Chem. Soc.* **135**, 18238-18247, doi:10.1021/ja408645b (2013).

64  Moosa, I. A. *Publish or perish: Perceived benefits versus unintended consequences*. (Edward Elgar Publishing, 2024).

65  Chen, M. *et al.* Evaluating large language models trained on code. *arXiv preprint arXiv:2107.03374* (2021).

66  Kulkarni, A. *Hair Health Prediction*, <https://www.kaggle.com/datasets/amitvkulkarni/hair-health> (2024).

67  AAAI. *AAAI Launches AI-Powered Peer-Review Assessment System*, <https://aaai.org/aaai-launches-ai-powered-peer-review-assessment-system/> (2025).

68  Opentrons. *Opentrons*, <https://github.com/Opentrons/opentrons> (2025).

69  Model Context Protocol. *Model Context Protocol*, <https://modelcontextprotocol.io/introduction> (2025).

70  Ten years of Nature Protocols. *Nature Protocols* **11**, 1327-1327, doi:10.1038/nprot.2016.115 (2016).

71  Ensuring accurate resource identification. *Nature Protocols* **15**, 1879-1880, doi:10.1038/s41596-020-0334-4 (2020).

72  Pallmann, P. *et al.* Adaptive designs in clinical trials: why use them, and how to run and report them. *BMC Medicine* **16**, 29, doi:10.1186/s12916-018-1017-7 (2018).

73  Schreier, A. A., Wilson, K. & Resnik, D. Academic research record-keeping: best practices for individuals, group leaders, and institutions. *Acad Med* **81**, 42-47, doi:10.1097/00001888-200601000-00010 (2006).

74  Bird, C. L., Willoughby, C. & Frey, J. G. Laboratory notebooks in the digital era: the role of ELNs in record keeping for chemistry and other sciences. *Chemical Society Reviews* **42**, 8157-8175, doi:10.1039/C3CS60122F (2013).

75  Samuel, S. & König-Ries, B. Understanding experiments and research practices for reproducibility: an exploratory study. *PeerJ* **9**, e11140, doi:10.7717/peerj.11140 (2021).

76  Marx, V. The big challenges of big data. *Nature* **498**, 255-260 (2013).

77  Hey, T., Tansley, S. & Tolle, K. M. *The fourth paradigm: data-intensive scientific discovery*. Vol. 1 (Microsoft research Redmond, WA, 2009).

78  Chen, M., Chen, L. & Zeng, A.-P. CRISPR/Cas9-facilitated engineering with growth-coupled and sensor-guided in vivo screening of enzyme variants for a more efficient chorismate pathway in E. coli. *Metabolic engineering communications* **9**, e00094 (2019).

79      Helleckes, L. M., Hemmerich, J., Wiechert, W., von Lieres, E. & Grünberger, A. Machine learning in bioprocess development: from promise to practice. *Trends in biotechnology* **41**, 817-835 (2023).

80      Tripathi, M. K., Nath, A., Singh, T. P., Ethayathulla, A. & Kaur, P. Evolving scenario of big data and Artificial Intelligence (AI) in drug discovery. *Molecular Diversity* **25**, 1439-1460 (2021).

81      Lipkova, J. *et al.* Artificial intelligence for multimodal data integration in oncology. *Cancer cell* **40**, 1095-1110 (2022).

82      Perakakis, N., Yazdani, A., Karniadakis, G. E. & Mantzoros, C.   Vol. 87    A1-A9 (Elsevier, 2018).

83      Rodrigues, J. F., Florea, L., De Oliveira, M. C., Diamond, D. & Oliveira, O. N. Big data and machine learning for materials science. *Discover Materials* **1**, 1-27 (2021).

84      Von Hayek, F. A. Economics and knowledge. *Economica* **4**, 33-54 (1937).

85      HISTORY.com    Editors.    *Ford's    assembly    line    starts    rolling*, <https://www.history.com/this-day-in-history/december-1/fords-assembly-line-starts-rolling> (2009).

86      Polanyi, M. *Personal Knowledge: Towards a Post-Critical Philosophy*.    (Routledge & Kegan Paul, 1958).

87      Harris, C. R. *et al.* Array programming with NumPy. *Nature* **585**, 357-362 (2020).

88      Bai, J. *et al.* A dynamic knowledge graph approach to distributed self-driving laboratories. *Nature Communications* **15**, 462 (2024).

89      Wikipedia contributors. *Software framework — Wikipedia, The Free Encyclopedia*, <https://en.wikipedia.org/wiki/Software_framework> (2025).

90      CommonMark. *CommonMark*, <https://commonmark.org/> (2024).

91      Stack    Overflow.    *2024    Stack    Overflow    Developer    Survey*, <https://survey.stackoverflow.co/2024/> (2024).

92      Pydantic Developers. *Pydantic*, <https://github.com/pydantic/pydantic> (2025).

93      OpenAi. *openai-python: The official Python library for the OpenAI API*, <https://github.com/openai/openai-python> (2025).

94      Preston-Werner, T. *TOML: Tom's Obvious Minimal Language*, <https://toml.io/en/> (2025).

95      Ben-Kiki, O., Evans, C. & Netzt, I. d. *YAML: YAML Ain't Markup Language*, <https://yaml.org/> (2025).

96      Json Schema Community. *JSON Schema*, <https://json-schema.org/> (2025).